\documentclass[acmsmall,screen, nonacm]{acmart}

\renewcommand\footnotetextcopyrightpermission[1]{}

\AtBeginDocument{%
  }

\newif\ifacm 
\acmfalse 

\newif\ifreview 
\reviewfalse 

\usepackage{booktabs}
\usepackage{array}
\usepackage{ragged2e}
\usepackage{etoolbox}   
\usepackage{adjustbox}  
\usepackage{makecell}
\usepackage{multirow}
\usepackage[table]{xcolor}
\usepackage{tabularx}
\usepackage{tabulary}
\usepackage{longtable}
\usepackage{xltabular} 

\usepackage[acronym,nogroupskip,nopostdot]{glossaries} 
\makenoidxglossaries
\usepackage{tikz}
\usetikzlibrary{shapes.geometric,arrows.meta,positioning,calc,fit,backgrounds,decorations.pathreplacing}

\usepackage{pdfcomment} 
\usepackage{xspace}
\usepackage[most]{tcolorbox}
\tcbuselibrary{skins}
\usepackage{microtype}
\usepackage{graphicx}

\renewcommand{\tabularxcolumn}[1]{m{#1}}

\newcounter{techrow}
\newcommand{\tech}[1]{%
  \refstepcounter{techrow}%
  \ifcsname phantomsection\endcsname\phantomsection\fi
  \thetechrow
  \label{tech:#1}
}
\newacronym{dl}{DL}{Deep Learning}
\newacronym{llm}{LLM}{Large Language Model}
\newacronym{nist}{NIST}{National Institute of Standards and Technology}
\newacronym{rmf}{RMF}{Risk Management Framework}
\newacronym{ml}{ML}{Machine Learning}
\newacronym{mlops}{MLOps}{Machine Learning Operations}
\newacronym{tl}{TL}{Transfer Learning}
\newacronym{rag}{RAG}{Retrieval-Augmented Generation}
\newacronym{icl}{ICL}{In-Context Learning}
\newacronym{cl}{CL}{Continual Learning}
\newacronym{ai}{AI}{Artificial Intelligence}
\newacronym{ct}{CT}{Continuous Training}
\newacronym{da}{DA}{Domain Adaptation}
\newacronym{mmd}{MMD}{Maximum Mean Discrepancy}
\newacronym{ft}{FT}{Fine-Tuning}
\newacronym{peft}{PEFT}{Parameter-Efficient Fine-Tuning}
\newacronym{sft}{SFT}{Supervised Fine-Tuning}
\newacronym{fsl}{FSL}{Few-Shot Learning}
\newacronym{ssl}{SSL}{Self-Supervised Learning}
\newacronym{rlhf}{RLHF}{Reinforcement Learning from Human Feedback}
\newacronym{rl}{RL}{Reinforcement Learning}
\newacronym{dpo}{DPO}{Direct Preference Optimization}
\newacronym{ppo}{PPO}{Proximal Policy Optimization}
\newacronym{tta}{TTA}{Test-Time Adaptation}
\newacronym{fl}{FL}{Federated Learning}
\newacronym{kd}{KD}{Knowledge Distillation}
\newacronym{al}{AL}{Active Learning}
\newacronym{fm}{FM}{Foundation Model}
\newacronym{genfm}{GenFM}{Generative Foundation Model}
\newacronym{repfm}{RepFM}{Representation Foundation Model}
\newacronym{mllm}{MLLM}{Multimodal Large Language Model}
\newacronym{moe}{MoE}{Mixture of Experts}
\newacronym{rlvr}{RLVR}{Reinforcement Learning from Verifiable Rewards}
\newacronym{grpo}{GRPO}{Group Relative Policy Optimization}
\newacronym{nas}{NAS}{Neural Architecture Search}
\newacronym{fti}{FTI}{Feature-Training-Inference}
\newacronym{pccp}{PCCP}{Predetermined Change Control Plan}
\newacronym{rmu}{RMU}{Representation Misdirection for Unlearning}
\newacronym{gpai}{GPAI}{General-Purpose AI}
\newacronym{mdr}{MDR}{Medical Device Regulation}
\newacronym{mlr}{MLR}{Multivocal Literature Review}
\newacronym{aegis}{AEGIS}{AI/ML Evaluation and Governance Infrastructure for Safety}
\newacronym{gdpr}{GDPR}{General Data Protection Regulation}
\newacronym{flop}{FLOP}{Floating-Point Operation}
\newacronym{memit}{MEMIT}{Mass-Editing Memory in a Transformer}
\newacronym{rome}{ROME}{Rank-One Model Editing}
\newacronym{mend}{MEND}{Model Editor Networks with Gradient Decomposition}
\newacronym{serac}{SERAC}{Semi-Parametric Editing with a Retrieval-Augmented Counterfactual Model}
\newacronym{caa}{CAA}{Contrastive Activation Addition}
\newacronym{cpt}{CPT}{Continued Pre-Training}
\newacronym{semiSL}{Semi-SL}{Semi-Supervised Learning}
\newacronym{mtl}{MTL}{Multi-Task Learning}
\newacronym{gmlp}{GMLP}{Good Machine Learning Practice}
\newacronym{imdrf}{IMDRF}{International Medical Device Regulators Forum}
\newacronym{lora}{LoRA}{Low-Rank Adaptation}
\newacronym{qlora}{QLoRA}{Quantized Low-Rank Adaptation}
\newacronym{cicd}{CI/CD}{Continuous Integration/Continuous Deployment}
\newacronym{ce}{CE}{Context Engineering}
\newacronym{mc}{MC}{Monte Carlo}
\newacronym{pe}{PE}{Prompt Engineering}
\newacronym{fda}{FDA}{Food and Drug Administration}
\newacronym{ivdr}{IVDR}{In Vitro Diagnostic Regulation}
\newacronym{mdsw}{MDSW}{Medical Device Software}
\newacronym{fria}{FRIA}{Fundamental Rights Impact Assessment}
\newacronym{pms}{PMS}{Post-Market Surveillance}
\newacronym{cdss}{CDSS}{Clinical Decision Support System}
\newacronym{rlaif}{RLAIF}{Reinforcement Learning from AI Feedback}
\newacronym{lp}{LP}{Linear Probing}
\newacronym{til}{TIL}{Task-Incremental Learning}
\newacronym{dil}{DIL}{Domain-Incremental Learning}
\newacronym{leace}{LEACE}{LEAst-squares Concept Erasure}
\newacronym{clip}{CLIP}{Contrastive Language-Image Pre-Training}
\newacronym{ews}{EWC}{Elastic Weight Consolidation}
 \newacronym{lwf}{LwF}{Learning without Forgetting}
\newacronym{kl}{KL}{Kullback-Leibler}
\newacronym{cai}{CAI}{Constitutional AI}
\newacronym{maml}{MAML}{Model-Agnostic Meta-Learning}
 \newacronym{star}{STaR}{Self-Taught Reasoner}
 \newacronym{rtn}{RTN}{Round-To-Nearest}
 \newacronym{awq}{AWQ}{Activation-aware Weight Quantization}
 \newacronym{ptq}{PTQ}{Post-Training Quantization}
\newacronym{gptq}{GPTQ}{Generative Pre-trained Transformers Quantization}
 \newacronym{qat}{QAT}{Quantization-Aware Training}
 \newacronym{cot}{CoT}{Chain-of-Thought}
 \newacronym{raft}{RAFT}{Retrieval-Augmented Fine-Tuning}
\newacronym{ece}{ECE}{Expected Calibration Error}
\newacronym{mlp}{MLP}{Multi-Layer Perceptron}
\newacronym{dapo}{DAPO}{Decoupled clip and Dynamic sAmpling Policy Optimization}
\newacronym{gan}{GAN}{Generative Adversarial Network}
\newacronym{nlp}{NLP}{Natural Language Processing}
\newacronym{swa}{SWA}{Stochastic Weight Averaging}
\newacronym{sgd}{SGD}{Stochastic Gradient Descent}
\newacronym{dpr}{DPR}{Dense Passage Retrieval}
\newacronym{retro}{RETRO}{Retrieval-Enhanced Transformer}
\newacronym{sisa}{SISA}{Sharded, Isolated, Sliced, and Aggregated}
\newacronym{tcav}{TCAV}{Testing with Concept Activation Vectors}
\newacronym{eda}{EDA}{Easy Data Augmentation}
\newacronym{slerp}{SLERP}{Spherical Linear Interpolation}
\newacronym{trl}{TRL}{Transformer Reinforcement Learning}
\newacronym{mcp}{MCP}{Model Context Protocol}
\newacronym{api}{API}{Application Programming Interface}
\newacronym{repe}{RepE}{Representation Engineering}
\newacronym{dora}{DoRA}{Weight-Decomposed Low-Rank Adaptation}
\newacronym{adalora}{AdaLoRA}{Adaptive Low-Rank Adaptation}
\newacronym{vera}{VeRA}{Vector-based Random Matrix Adaptation}
\newacronym{rslora}{rsLoRA}{rank-stabilized Low-Rank Adaptation}
\newacronym{pissa}{PiSSA}{Principal Singular values and Singular vectors Adaptation}
\newacronym{dude}{DuDe}{Dual Decomposition of Weights and Singular Value Low-Rank Adaptation}
\newacronym{fedpara}{FedPara}{Federated Parameterization}
\newacronym{dylora}{DyLoRA}{Dynamic Low-Rank Adaptation}
\newacronym{moelora}{MoELoRA}{Low-Rank Adaptation as a Mixture of Experts }
\newacronym{ia3}{IA$^3$}{Infused Adapter by Inhibiting and Amplifying Inner Activations}
\newacronym{ties}{TIES-MERGING}{TrIm, Elect Sign \& Merge}
\newacronym{dare}{DARE}{Drop And REscale}
\newacronym{della}{DELLA-Merging}{Drop and rEscaLe via sampLing with mAgnitude}
\newacronym{emr}{EMR-Merging}{Elect, Mask \& Rescale-Merging}
\newacronym{lata}{LATA}{Layer-Aware Task Arithmetic}
\newacronym{gspo}{GSPO}{Group Sequence Policy Optimization}
\newacronym{sapo}{SAPO}{Step-Aligned Policy Optimization}
\newacronym{cispo}{CISPO}{Clipped IS-weight Policy Optimization}
\newacronym{rloo}{RLOO}{REINFORCE Leave-One-Out}
\newacronym{orpo}{ORPO}{Odds Ratio Preference Optimization}
\newacronym{simpo}{SimPO}{Simple Preference Optimization}
\newacronym{kto}{KTO}{Kahneman-Tversky Optimization}
\newacronym{cpo}{CPO}{Contrastive Preference Optimization}
\newacronym{mallows}{MallowsPO}{Mallows-model Preference Optimization}
\newacronym{ipo}{IPO}{Identity Preference Optimization}
\newacronym{fomaml}{FOMAML}{First-Order Model Agnostic Meta Learning}
\newacronym{metasgd}{Meta-SGD}{Meta Stochastic Gradient Descent}
\newacronym{leo}{LEO}{Latent Embedding Optimization}
\newacronym{cavia}{CAVIA}{Context Adaptation VIA meta-learning}
\newacronym{hipporag}{HippoRAG}{Hippocampus-inspired Retrieval-Augmented Generation}
\newacronym{graphrag}{GraphRAG}{Graph Retrieval-Augmented Generation}
\newacronym{selfrag}{Self-RAG}{Self-Reflective Retrieval-Augmented Generation}
\newacronym{pmet}{PMET}{Precise Model Editing in a Transformer}
\newacronym{make}{MAKE}{Memory-Associated Knowledge Editing}
\newacronym{wmdp}{WMDP}{Weapons of Mass Destruction Proxy}
\newacronym{sae}{SAE}{Sparse AutoEncoder}
\newacronym{fila}{FILA}{Fisher-Initialization of Low-rank Adapters}
\newacronym{loku}{LoKU}{Low-rank Knowledge Unlearning}
\newacronym{loka}{LOKA}{Large language mOdel Knowledge updAtes}
\newacronym{fv}{FV}{Function Vector}
\newacronym{svd}{SVD}{Singular Value Decomposition}
\newacronym{bert}{BERT}{Bidirectional Encoder Representations from Transformers}
\newacronym{ftl}{FTL}{Federated Transfer Learning}
 \newacronym{magprune}{MAGPRUNE}{Magnitude-based
Pruning}
\newacronym{prf}{PRF}{Probabilistic Relevance Framework}
\newacronym{gpt}{GPT}{Generative Pre-trained Transformer}
\newacronym{iti}{ITI}{Inference-Time Intervention}
\newacronym{actadd}{ActAdd}{Activation Addition}
\newacronym{nmt}{NMT}{Neural Machine Translation}
\newacronym{rope}{RoPE}{Rotary Positional Embedding}
\newacronym{reft}{ReFT}{Representation Fine-Tuning}
\newacronym{apo}{APO}{Automatic Prompt Optimization}
\newacronym{dspy}{DSPy}{Declarative Self-improving Python}
\newacronym{dpft}{DP-FT}{Differentially Private Fine-Tuning}
\newacronym{dppeft}{DP-PEFT}{Differentially Private Parameter-Efficient Fine-Tuning}
\newacronym{dpsgd}{DP-SGD}{Differentially Private Stochastic Gradient Descent}
\newacronym{pi}{PI}{Position Interpolation}
\newacronym{yarn}{YaRN}{Yet another RoPE extensioN}
\newacronym{pepe}{PEPE}{Periodic Extrapolation Positional Encodings}
\newacronym{gd}{GD}{Gradient Descent}
\newacronym{ga}{GA}{Gradient Ascent}
\newacronym{orpo_prompt}{OPRO}{Optimization by PROmpting}
\newacronym{bitfit}{BitFit}{BIas-Term FIne-Tuning}
 \newacronym{ttt}{TTT}{Test-Time Training}
\newacronym{raft_rank}{RAFT}{Reward rAnked FineTuning}
\newacronym{milora_svd}{MiLoRA}{Minor singular component based Low-Rank Adaptation}
\newacronym{loso}{LOSO}{Leave One Seed Out}
\newacronym{spin}{SPIN}{Self-Play fIne-tuNing}
\newacronym{cpt_hf}{CPT}{Context-Aware Prompt Tuning}
  
\newcommand{\dimens}[2]{$\text{#1}_\text{{#2}}$} 
\newcommand{\dMechanism}{\dimens{D}{1} (Mechanism)\xspace}
\newcommand{\dGoal}{\dimens{D}{2} (Goal)\xspace}
\newcommand{\dData}{\dimens{D}{3} (Data Requirements)\xspace}
\newcommand{\dPersistence}{\dimens{D}{4} (Persistence)\xspace}
\newcommand{\dScope}{\dimens{D}{5} (Scope)\xspace}
\newcommand{\dModel}{\dimens{D}{6} (Model Type)\xspace}

\newtcbox{\taxbox}{
  on line,
  nobeforeafter,
  tcbox raise base,
  arc=4.5pt,
  colback=black!5,
  colframe=black!15,
  colupper=black!60,
  boxrule=0.5pt,
  boxsep=0pt,
  left=1.5pt,right=1.5pt,
  top=1.5pt,bottom=1.5pt,
  fontupper=\sffamily\scriptsize
}

\newcommand{\tax}[2]{\allowbreak\taxbox{{\dimens{d}{#1} ${=}$ \hyperlink{dim:#2}{\csname #2\endcsname}}}}
\newcommand{\taxNo}[2]{\allowbreak\taxbox{{\dimens{d}{#1} ${=}$ {\csname #2\endcsname}}}} 
\newcommand{\taxNot}[2]{\taxbox{{\dimens{d}{#1} ${\neq}$ \hyperlink{dim:#2}{\csname #2\endcsname}}}}
\newcommand{\ctx}{Context Injection\xspace}
\newcommand{\ctxAbbr}{\hyperlink{dim:ctx}{Context Inj.\xspace}}

\newcommand{\selAbbr}{\paramUpdAbbr}

\newcommand{\fullAbbr}{\paramUpdAbbr}
\newcommand{\paramUpd}{Parametric Update\xspace}
\newcommand{\paramUpdAbbr}{\hyperlink{dim:paramUpd}{Param. Upd.\xspace}}
\newcommand{\selFullAbbr}{\paramUpdAbbr}

\newcommand{\WeightSpaceComp}{Parameter Composition\xspace}
\newcommand{\WeightSpaceCompAbbr}{\hyperlink{dim:WeightSpaceComp}{Param. Comp.\xspace}}
\newcommand{\WeightSpaceCompConcise}{Algebraic combination of models or mathematical transformation of parameters.\xspace}
\newcommand{\CrossModelTransf}{Cross-Model Transfer\xspace}
\newcommand{\CrossModelTransfAbbr}{\hyperlink{dim:CrossModelTransf}{Cross-Model Transf.\xspace}}
\newcommand{\CrossModelTransfConcise}{Training a new model artifact to replicate a teacher model's outputs or representations.\xspace}
\newcommand{\StructReduct}{Parametric Compression\xspace}
\newcommand{\StructReductAbbr}{\hyperlink{dim:StructReduct}{Param. Compres.\xspace}}
\newcommand{\StructReductConcise}{
Reducing a model's parametric size, precision, or density to optimize for deployment, without expanding the computational graph.
\xspace}
\newcommand{\arch}{Architectural Modification\xspace}
\newcommand{\archAbbr}{\hyperlink{dim:arch}{Arch.\xspace}}
\newcommand{\manip}{Pipeline-Mediated\xspace}
\newcommand{\manipAbbr}{\hyperlink{dim:manip}{Pipe.-Med.\xspace}}

\newcommand{\paramArchAbbr}{\{\paramUpdAbbr, {\archAbbr}\} \xspace}
\newcommand{\transAct}{Activation-Space Manipulation\xspace}
\newcommand{\transActAbbr}{\hyperlink{dim:transAct}{Act.-Space Manip.\xspace}}
\newcommand{\infTimeSearch}{Inference-Time Search\xspace}
\newcommand{\infTimeSearchAbbr}{\hyperlink{dim:infTimeSearch}{Infer.-Time Search\xspace}}
\newcommand{\infTimeSearchConcise}{Allocating variable inference compute through internal generation processes without parameter modification.\xspace}

\newcommand{\selWeightSpaceAbbr}{\{\selAbbr, {\WeightSpaceCompAbbr}\}\xspace}

\newcommand{\task}{Task Specialization\xspace}
\newcommand{\taskAbbr}{\hyperlink{dim:task}{Task Spec.\xspace}}
\newcommand{\domAdapt}{Distributional Gap Bridging\xspace}
\newcommand{\domAdaptAbbr}{\hyperlink{dim:domAdapt}{Dist. Gap Bridging\xspace}}
\newcommand{\capExt}{Capability Extension\xspace}
\newcommand{\capExtAbbr}{\hyperlink{dim:capExt}{Capab. Ext.\xspace}}

\newcommand{\capExtCompEffAbbr}{\{\capExtAbbr, {\compEffAbbr}\}\xspace}
\newcommand{\capExtAlgnAbbr}{\{\capExtAbbr, {\algnAbbr}\}\xspace}
\newcommand{\domAdaptCapExtAbbr}{\{\capExtAbbr, {\domAdaptAbbr}\}\xspace}

\newcommand{\augK}{Knowledge Update\xspace}
\newcommand{\augKAbbr}{\hyperlink{dim:augK}{Knowl. Upd.\xspace}}
\newcommand{\rem}{Remove Knowledge\xspace}
\newcommand{\remKAbbr}{\hyperlink{dim:rem}{Rem. Knowl.\xspace}}

\newcommand{\remKPrivAbbr}{\{\remKAbbr, {\privAbbr}\}\xspace}
\newcommand{\algn}{Alignment\xspace} 
\newcommand{\algnAbbr}{\hyperlink{dim:algn}{Align.\xspace}} 
\newcommand{\safety}{Safety\xspace} 
\newcommand{\safetyAbbr}{\hyperlink{dim:safety}{Safety\xspace}}
\newcommand{\reas}{Reasoning\xspace}
\newcommand{\reasAbbr}{\hyperlink{dim:reas}{Reas.\xspace}}

\newcommand{\algnSafetyAbbr}{\{\algnAbbr, {\safetyAbbr}\}\xspace}
\newcommand{\DriftRemed}{Drift Remediation\xspace}
\newcommand{\DriftRemedAbbr}{\hyperlink{dim:DriftRemed}{Drift Remed.\xspace}}
\newcommand{\ContAdapt}{Continual Adaptation\xspace}
\newcommand{\ContAdaptAbbr}{\hyperlink{dim:ContAdapt}{Cont. Adapt.\xspace}}
\newcommand{\build}{Build Capability\xspace}
\newcommand{\buildAbbr}{\hyperlink{dim:build}{Build Capab.\xspace}}
\newcommand{\rob}{Robustness\xspace}
\newcommand{\robAbbr}{\hyperlink{dim:rob}{Rob.\xspace}}
\newcommand{\compEff}{Computational Efficiency\xspace}
\newcommand{\compEffAbbr}{\hyperlink{dim:compEff}{Comp. Eff.\xspace}}

\newcommand{\taskCompEffAbbr}{\{\taskAbbr, {\compEffAbbr}\}\xspace}

\newcommand{\taskAlgnReasAbbr}{\{\taskAbbr, \algnAbbr, {\reasAbbr}\}\xspace}

\newcommand{\taskAlgnSafeAbbr}{\{\taskAbbr, \algnAbbr, {\safetyAbbr}\}\xspace}

\newcommand{\taskRobAbbr}{\{\taskAbbr, {\robAbbr}\}\xspace}
\newcommand{\priv}{Privacy Preservation\xspace}
\newcommand{\privAbbr}{\hyperlink{dim:priv}{Priv. Preserv.\xspace}}
\newcommand{\explain}{Explainability\xspace}
\newcommand{\explainAbbr}{\hyperlink{dim:explain}{Expl.\xspace}}
\newcommand{\userPerson}{Personalization\xspace}
\newcommand{\userPersonAbbr}{\hyperlink{dim:userPerson}{Person.\xspace}}

\newcommand{\augKPersonAbbr}{\{\augKAbbr, {\userPersonAbbr}\}\xspace}

\newcommand{\taskAugKAbbr}{\{\taskAbbr, {\augKAbbr}\}\xspace}

\newcommand{\taskDomAdaptAbbr}{\{\taskAbbr, {\domAdaptAbbr}\}\xspace}

\newcommand{\augKBehavAbbr}{\{\augKAbbr, {\behavControlAbbr}\}\xspace}

\newcommand{\ContAdaptDriftRemedAbbr}{\{\ContAdaptAbbr, {\DriftRemedAbbr}\}\xspace}

\newcommand{\taskBehavAbbr}{\{\taskAbbr, {\behavControlAbbr}\}\xspace}

\newcommand{\ext}{External Corpus\xspace}
\newcommand{\extAbbr}{\hyperlink{dim:ext}{Ext. Corpus\xspace}}
\newcommand{\smallD}{Small Labeled\xspace}
\newcommand{\smallDAbbr}{\hyperlink{dim:smallD}{Small Labeled\xspace}}
\newcommand{\zeroSh}{Zero-Shot\xspace}
\newcommand{\zeroShAbbr}{\hyperlink{dim:zeroSh}{Zero-Shot\xspace}}
\newcommand{\multiD}{Paired Multimodal\xspace}
\newcommand{\multiDAbbr}{\hyperlink{dim:multiD}{Pair. Multimod.\xspace}}
\newcommand{\wght}{Parameter-Only\xspace}
\newcommand{\wghtAbbr}{\hyperlink{dim:wght}{Param.-Only\xspace}\xspace}
\newcommand{\tstOnly}{Test-Only\xspace}
\newcommand{\tstOnlyAbbr}{\hyperlink{dim:tstOnly}{Test-Only\xspace}}
\newcommand{\pref}{Preference Pairs\xspace}
\newcommand{\prefAbbr}{\hyperlink{dim:pref}{Pref. Pairs\xspace}}
\newcommand{\forget}{Forget Set Specification\xspace}
\newcommand{\forgetAbbr}{\hyperlink{dim:forget}{Forget Set Spec.\xspace}}
\newcommand{\few}{Few Demonstrations\xspace}
\newcommand{\fewAbbr}{\hyperlink{dim:few}{Few Demo.\xspace}}
\newcommand{\largeD}{Large Labeled\xspace}
\newcommand{\largeDAbbr}{\hyperlink{dim:largeD}{Large Labeled\xspace}}
\newcommand{\verify}{Verifiable Task\xspace}
\newcommand{\verifyAbbr}{\hyperlink{dim:verify}{Verif. Task\xspace}}
\newcommand{\unlabD}{Unlabeled\xspace}
\newcommand{\unlabDAbbr}{\hyperlink{dim:unlabD}{Unlabeled\xspace}}

\newcommand{\unlabSmallDAbbr}{\{\unlabDAbbr, {\smallDAbbr}\}\xspace}
\newcommand{\unlabDWghtAbbr}{\{\unlabDAbbr, {\wghtAbbr}\}\xspace}

\newcommand{\SmallLargeDAbbr}{\{\smallDAbbr, {\largeDAbbr}\}\xspace}

\newcommand{\decent}{Decentralized\xspace}
\newcommand{\decentAbbr}{\hyperlink{dim:decent}{Decent.\xspace}}
\newcommand{\seqIncD}{Sequential / Incremental\xspace}
\newcommand{\seqIncDAbbr}{\hyperlink{dim:seqIncD}{Seq./Inc.\xspace}}
\newcommand{\taskDist}{Task Distribution\xspace}
\newcommand{\taskDistAbbr}{\hyperlink{dim:taskDist}{Task Dist.\xspace}}
\newcommand{\synthTeacher}{Teacher-Derived\xspace}
\newcommand{\synthTeacherAbbr}{\hyperlink{dim:synthTeacher}{Teacher-Deriv.\xspace}}
\newcommand{\synthRule}{Constitution-Derived\xspace}
\newcommand{\synthRuleAbbr}{\hyperlink{dim:synthRule}{Constitution-Deriv.\xspace}}
\newcommand{\synthEnv}{Environment-Derived\xspace}
\newcommand{\synthEnvAbbr}{\hyperlink{dim:synthEnv}{Environment-Deriv.\xspace}}
\newcommand{\userInteract}{User Interaction Data\xspace}
\newcommand{\userInteractAbbr}{\hyperlink{dim:userInteract}{User Interact.\xspace}}

\newcommand{\extUserAbbr}{\{\extAbbr, {\userInteractAbbr}\}\xspace}

\newcommand{\zeroFewAbbr}{\{\zeroShAbbr, {\fewAbbr}\}\xspace}

\newcommand{\smallFewAbbr}{\{\smallDAbbr, {\fewAbbr}\}\xspace}
\newcommand{\instrResp}{Instruction-Response Corpus\xspace}
\newcommand{\instrRespAbbr}{\hyperlink{dim:instrResp}{Instr.-Resp. Corpus\xspace}}
\newcommand{\pipeInherit}{Pipeline-Dependent\xspace}
\newcommand{\pipeInheritAbbr}{\hyperlink{dim:pipeInherit}{Pipe.-Dep.\xspace}}

\newcommand{\ScheduledPerm}{Scheduled Permanent\xspace}
\newcommand{\ScheduledPermAbbr}{\hyperlink{dim:ScheduledPerm}{Sched. Perm.\xspace}}
\newcommand{\AdhocPerm}{Ad-hoc Permanent\xspace}
\newcommand{\AdhocPermAbbr}{\hyperlink{dim:AdhocPerm}{Ad-hoc Perm.\xspace}}
\newcommand{\eph}{Session-Ephemeral\xspace}
\newcommand{\ephAbbr}{\hyperlink{dim:eph}{Sess.-Eph.\xspace}}
\newcommand{\verPersistent}{Version-Persistent\xspace}
\newcommand{\verPersistentAbbr}{\hyperlink{dim:verPersistent}{Ver.-Persist.\xspace}}
\newcommand{\boundedCumul}{Bounded Cumulative\xspace}
\newcommand{\boundedCumulAbbr}{\hyperlink{dim:boundedCumul}{Bound. Cumul.\xspace}}
\newcommand{\unboundedCumul}{Unbounded Cumulative\xspace}
\newcommand{\unboundedCumulAbbr}{\hyperlink{dim:unboundedCumul}{Unbound. Cumul.\xspace}}
\newcommand{\trans}{Transient\xspace}
\newcommand{\transAbbr}{\hyperlink{dim:trans}{Transient\xspace}}
\newcommand{\pipeDep}{Pipeline-Dependent\xspace}
\newcommand{\pipeDepAbbr}{\hyperlink{dim:pipeDep}{Pipe.-Dep.\xspace}}
\newcommand{\fairness}{Fairness\xspace}
\newcommand{\fairnessAbbr}{\hyperlink{dim:fairness}{Fair.\xspace}}
\newcommand{\reliability}{Reliability\xspace}
\newcommand{\reliabilityAbbr}{\hyperlink{dim:reliability}{Reliab.\xspace}}
\newcommand{\behavControl}{Behavior Control\xspace}
\newcommand{\behavControlAbbr}{\hyperlink{dim:behavControl}{Behav. Ctrl.\xspace}}

\newcommand{\boundedUnboundedCumulAbbr}{\{\boundedCumulAbbr, {\unboundedCumulAbbr}\}\xspace}

\newcommand{\SchedAdhocPermAbbr}{\{\ScheduledPermAbbr, {\AdhocPermAbbr}\}\xspace}

\newcommand{\unboundedCumulSchedPermAbbr}{\{\unboundedCumulAbbr, {\ScheduledPermAbbr}\}\xspace}

\newcommand{\ephVerPersistentAbbr}{\{\ephAbbr, {\verPersistentAbbr}\}\xspace}

\newcommand{\partl}{Partial\xspace}
\newcommand{\partlAbbr}{\hyperlink{dim:partl}{Part.\xspace}}
\newcommand{\whole}{Whole-Model\xspace}
\newcommand{\wholeAbbr}{\hyperlink{dim:whole}{Whole\xspace}}
\newcommand{\modSwap}{Modular\xspace}
\newcommand{\modSwapAbbr}{\hyperlink{dim:modSwap}{Modular\xspace}}
\newcommand{\dIVdist}{Distributed\xspace}
\newcommand{\dIVdistAbbr}{\hyperlink{dim:dIVdist}{Dist.\xspace}}

\newcommand{\partWholeAbbr}{\{\partlAbbr, {\wholeAbbr}\}\xspace}

\newcommand{\partWholeSurrogAbbr}{\{\partlAbbr, \wholeAbbr, {\surrogateAbbr}\}\xspace}
\newcommand{\fusedComp}{Fused Composition\xspace}
\newcommand{\fusedCompAbbr}{\hyperlink{dim:fusedComp}{Fused\xspace}}
\newcommand{\extContxt}{Input/Output-Space\xspace}
\newcommand{\extContxtAbbr}{\hyperlink{dim:extContxt}{I/O Space\xspace}}
\newcommand{\surrogate}{Surrogate Model\xspace}
\newcommand{\surrogateAbbr}{\hyperlink{dim:surrogate}{Surrog. Model\xspace}}
\newcommand{\compressed}{Compressed Artifact\xspace}
\newcommand{\compressedAbbr}{\hyperlink{dim:compressed}{Compress. Artif.\xspace}}
\newcommand{\actSpace}{Activation-Space\xspace}
\newcommand{\actSpaceAbbr}{\hyperlink{dim:actSpace}{Activ.-Space\xspace}}
\newcommand{\pipelineDepScope}{Pipeline-Dependent\xspace}
\newcommand{\pipelineDepScopeAbbr}{\hyperlink{dim:pipelineDepScope}{Pipe.-Dep.\xspace}}

\newcommand{\partlModAbbr}{\{\partlAbbr, {\modSwapAbbr}\}\xspace}

\newcommand{\PartlFusedCompAbbr}{\{\partlAbbr, {\fusedCompAbbr}\}\xspace}

\newcommand{\mllm}{MLLM\xspace}
\newcommand{\lm}{\{LLM, MLLM\}\xspace}
\newcommand{\FM}{\{FM, MLLM\}\xspace}
\newcommand{\df}{\{DL, FM\}\xspace}
\newcommand{\dL}{\{DL, LLM\}\xspace}
\newcommand{\flm}{\{FM, LLM, MLLM\}\xspace}
\newcommand{\dfl}{\{DL, FM, LLM\}\xspace}
\newcommand{\dflm}{\{DL, FM, LLM, MLLM\}\xspace}
\newcommand{\all}{\{ML, DL, FM, LLM, MLLM\}\xspace}

\newcommand{\NbTech}{48\xspace}
\newcommand{\NbChains}{12\xspace}
\newcommand{\nbFamilies}{9\xspace}
\newcommand{\nbDim}{6\xspace}
\newcommand{\nbGreyLit}{20\xspace}

\newcommand{\longContext}{Long-Context Extension\xspace}
\newcommand{\longContextAbbr}{Long-Context Ext.\xspace}

\newcommand{\frameTax}{taxonomy\xspace}
\newcommand{\FrameTax}{Taxonomy\xspace}

\newcommand{\profile}[6]{(#1,\hspace{0.3em plus 0.2em minus 0.1em}\allowbreak #2,\hspace{0.3em plus 0.2em minus 0.1em}\allowbreak #3,\hspace{0.3em plus 0.2em minus 0.1em}\allowbreak #4,\hspace{0.3em plus 0.2em minus 0.1em}\allowbreak #5,\hspace{0.3em plus 0.2em minus 0.1em}\allowbreak #6)}

\newcommand{\TITLE}{A Six-Dimensional Taxonomy of Post-Training Adaptation Techniques with Applications in AI Governance}

\ifacm
\newcommand{\SuppAcronym}{Supplementary Section S1\xspace}
\newcommand{\SuppMethodology}{Supplementary Section S2\xspace} 
\newcommand{\SuppCategory}{Supplementary Section S3\xspace} 
\newcommand{\SuppTech}{Supplementary Section S4\xspace} 
\newcommand{\SuppQuantitative}{Supplementary Section S5\xspace} 
\newcommand{\SuppGov}{Supplementary Section S6\xspace} 
\else 
\newcommand{\SuppMethodology}{Appendix \ref{appendix:methodology}\xspace} 
\newcommand{\SuppCategory}{Appendix \ref{sec:appendix_definitions}\xspace} 
\newcommand{\SuppTech}{Appendix \ref{sec:TechDefinitions}\xspace} 
\newcommand{\SuppQuantitative}{Appendix \ref{appendix:quantitative}\xspace} 
\newcommand{\SuppGov}{Appendix \ref{appendix:governance}\xspace} 
\fi

\usepackage{placeins}

\usepackage{totcount}

\newcounter{backcount}
\newcounter{seedcount}
\newcounter{forwardcount}

\regtotcounter{backcount}
\regtotcounter{seedcount}
\regtotcounter{forwardcount}

\newcommand{\addback}{\stepcounter{backcount}} 
\newcommand{\addseed}{\stepcounter{seedcount}} 
\newcommand{\addforward}{\stepcounter{forwardcount}} 

\begin{document}

\title{
\TITLE
}

\author{Fardin Afdideh}
\email{fardin.afdideh@ki.se}
\orcid{0009-0006-8430-3130}
\affiliation{%
  \institution{Karolinska Institutet}
  \department{Department of Clinical Science, Intervention and Technology}
  \city{Stockholm}
  \country{Sweden}
}

\author{Fernando Seoane}
\email{fernando.seoane@ki.se}
\orcid{0000-0002-6995-967X}
\affiliation{%
  \institution{Karolinska Institutet}
  \department{Department of Clinical Science, Intervention and Technology}
  \city{Stockholm}
  \country{Sweden}
}
\affiliation{%
  \institution{Karolinska University Hospital}
  \department{Department of Clinical Physiology}
  \city{Stockholm}
  \country{Sweden}
}
\affiliation{%
  \institution{University of Bor\aa s}
  \department{Department of Textile Technology}
  \city{Bor\aa s}
  \country{Sweden}
}
\affiliation{%
  \institution{Karolinska University Hospital}
  \department{Department of Medical Technologies}
  \city{Huddinge}
  \country{Sweden}
}

\author{Farhad Abtahi}
\orcid{0000-0001-7807-8682}
\email{farhad.abtahi@ki.se}
\affiliation{%
  \institution{Karolinska Institutet}
  \department{Department of Clinical Science, Intervention and Technology}
  \city{Stockholm}
  \country{Sweden}
}
\affiliation{%
  \institution{Karolinska University Hospital}
  \department{Department of Clinical Physiology}
  \city{Stockholm}
  \country{Sweden}
}
\affiliation{%
  \institution{KTH Royal Institute of Technology}
  \department{Department of Biomedical Engineering and Health System}
  \city{Huddinge}
  \country{Sweden}
}

\renewcommand{\shortauthors}{Afdideh, Seoane, and Abtahi}

\begin{abstract}
Post-training adaptation has become central to modern machine learning practice and includes techniques such as retraining, fine-tuning, parameter-efficient adaptation, alignment, retrieval augmentation, model editing, unlearning, calibration, and Multimodal Instruction Tuning. However, the literature remains fragmented across technique families, model classes and deployment contexts, making it difficult to compare methods or describe how a trained model has been modified. This survey synthesizes the post-training adaptation literature and introduces a six-dimensional taxonomy organized by mechanism, goal, data requirement, persistence, structural scope, and model type. The taxonomy distinguishes commonly conflated terms such as fine-tuning, 
retrieval augmentation, and prompting, and shows how adaptation strategies evolve from traditional machine learning through deep learning, foundation models, large language models, and multimodal large language models. 
It also maps relationships among techniques, including inheritance, supersession, hybridization, and layered deployment stacks. 
The resulting vocabulary can support technical documentation, model-change tracking, and governance analysis. The survey concludes by identifying open challenges in evaluation, reproducibility, persistent inference-time adaptation, unlearning, multimodal adaptation, and governance-aware post-training workflows.
\end{abstract}


\maketitle

\section{Introduction}
\label{sec:Introduction}
\gls{ml} models are rarely static; The environments in which they operate change, e.g., data distributions shift, user behaviors evolve, regulatory requirements change, and new capabilities become desirable. The performance of a model trained on historical data degrades unless the model is adapted to its evolving context. While managing traditional \gls{ml} drift is a standard operational requirement, the recent proliferation of parameter-efficient and inference-time adaptation techniques for \glspl{fm} has drastically accelerated the regulatory and change-control challenges surrounding post-deployment modifications. The landscape of \textit{post-training techniques}, i.e., methods applied to a model that has already been trained, and that may already be deployed or carry regulatory approval, to modify its behavior, capabilities, or deployment characteristics, is rapidly expanding. 
The qualifier ``post-training'' is deliberate, as it identifies the operational context in which compliance obligations regarding modification and change control arise, distinguishing these techniques from the initial training process that precedes deployment. The systematic organization of post-training techniques is a central challenge in both \gls{ml} system design and AI governance.
\subsection{Problem and Contributions}
\label{sec:ProblemAndContributions}
At least 
a dozen techniques exist for modifying \gls{ml} models, ranging from classical retraining and \gls{tl} to modern methods such as \gls{peft}, \gls{rlhf}, \gls{rlvr}, and \gls{rag}. This has produced a fragmented literature characterized by three specific problems: 1) \textbf{terminological ambiguity}, where the same term is used for different concepts, e.g., ``\gls{ft}'' may refer to full parameter updates, partial layer freezing, or adapter training; 2) \textbf{single-axis taxonomies}, which force ambiguous placements (is \gls{peft} an \gls{ft} technique or an efficiency technique?) and obscure cross-dimensional relationships; and 3) \textbf{model-type conflation}, where current frameworks frequently ignore how model type dictates feasible techniques (\gls{pe} is meaningless for a Random Forest; classical \gls{da} has been largely superseded at the \gls{llm} level by \gls{cpt} and \gls{peft}).
 
These problems can carry legal and financial consequences.
For instance, the EU AI Act treats ``substantial modification'' as a concept that may lead to re-classification of high-risk AI systems and specifies that 
\gls{ft} of a third-party model may, depending on context, place the modifier in a \textit{provider}-like role with associated obligations~\cite{noauthor_regulation_2024};
however, it provides no
technical definition of what constitutes ``\gls{ft}'' versus other modifications. The Commission's \gls{gpai} guidelines introduced a compute-based threshold (one-third of training \glspl{flop}), but as Hacker and Holweg argue ~\cite{hacker_regulation_2025}, this proxy may fail to capture low-compute techniques such as Knowledge Editing and Activation Steering that substantially alter behavior with negligible compute~\cite{turner_activation_2023}. Similarly, the \gls{pccp} framework of the \gls{fda}~\cite{health_marketing_2025} allows pre-specified model modifications; however, audits of historical~\cite{wu_how_2021} and current~\cite{mehta_evaluating_2025} submissions reveal a persistent lack of transparency and omission of technical details required to evaluate model reliability and change.

This paper addresses these problems by proposing a \textit{six-dimensional characterization taxonomy} for \gls{ml} model adaptation. Rather than organizing techniques into a single hierarchy, the taxonomy assigns every technique a coordinate on six dimensions: 
1) \textbf{Mechanism (\dimens{D}{1})}: What changes in the system;
2) \textbf{Goal (\dimens{D}{2})}: Why adapt;
3) \textbf{Data Requirements (\dimens{D}{3})}: What data is needed;
4) \textbf{Persistence (\dimens{D}{4})}: How long the change lasts;
5) \textbf{Scope (\dimens{D}{5})}: How much of the model structure is affected; and
6) \textbf{Model Type (\dimens{D}{6})}: What model is being adapted (Modulating Filter).
Dimensions~\dimens{D}{1}--\dimens{D}{5} \textit{characterize} 
the declared adaptation intervention through canonical practical profiles;
\dimens{D}{6} \textit{modulates} and filters feasible \dimens{D}{1}--\dimens{D}{5} combinations by the target model and resolves the model-type conflation problem. The taxonomy's primary artifact is a \textit{centerpiece table} (Table~\ref{tab:centerpiece}) mapping all \NbTech techniques to their six-dimensional profiles,
complemented by a 
governance mapping that connects the technical dimensions to regulatory requirements under the \gls{nist} \gls{ai} \gls{rmf}, the EU AI Act, the EU \gls{mdr}/\gls{ivdr}, and the \gls{fda} \gls{pccp}.

\subsection{Scope}
\label{sec:Scope}
The inclusion principle governing this survey is that \textit{a technique is in scope if it modifies the behavior, capabilities, or deployment characteristics of an already-existing \gls{ml} model}, that is, one that has been trained and may already be deployed or carry regulatory approval. The inclusion criterion is technical, but the inclusion of a technique in this taxonomy does not imply that the technique always constitutes a retraining event, a substantial modification, or a reportable device change. A technique is in scope because it can influence the behavior, consistency, reliability, or deployment characteristics of an already-existing model and may therefore need to be described in regulated change-control documentation. Whether a specific application of any technique triggers re-certification, a new submission, or a conformity-assessment update is a separate determination that depends on intended purpose, risk profile, clinical or operational impact, and the applicable regulatory pathway. 
However, the taxonomy does not claim to cover all AI system controls relevant to regulation. Deployment-layer controls, e.g., output guardrails, content moderation filters, watermarking, post-processing rules, and human-in-the-loop review, may also be regulatory-relevant system changes.
However, they are excluded here because they do not modify the model's parameters, inference context, or adaptation pipeline as defined in this taxonomy. Their omission is a scope boundary, not a claim of regulatory irrelevance.

The inclusion principle admits two categories of technique. The \textit{core category} comprises methods applied after training to modify a model's parameters, inference context, or outputs (e.g., \gls{ft}, \gls{peft}, alignment, \gls{rag}, Knowledge Editing). 
A \textit{boundary extension} encompasses training-time data-pipeline strategies, e.g., Data Augmentation, Curriculum Learning, 
and Active Learning,
when applied to an already-existing model as an adaptation mechanism rather than as part of capability building. These boundary-extension techniques are classified as \tax{1}{manip} with \tax{4}{pipeDep} in Table~\ref{tab:centerpiece}, making their distinct status visible. 
Foundational learning paradigms such as supervised, unsupervised, and reinforcement learning, low-level mechanical optimizations, and deployment-layer controls such as output guardrails, watermarking, moderation filters that constrain outputs without modifying model behavior are excluded.

The terms \textit{post-training techniques} and \textit{model adaptation} refer to the same \NbTech-technique scope but carry context-dependent emphasis. 
While model adaptation serves as the traditional academic umbrella for modifying a model's behavior or capabilities, post-training techniques are the contemporary industry and regulatory framing. It foregrounds that the model already exists and may carry compliance obligations such as those under the EU AI Act's ``substantial modification'' provisions, the \gls{fda}'s \gls{pccp}, and \gls{mdr} post-market monitoring. Consequently, this paper uses ``model adaptation'' to discuss technical properties and ``post-training techniques'' to address operational or regulatory change-control contexts.

\gls{ssl} is included only in its adaptation role (e.g., \gls{cpt} on domain-specific data). The model type hierarchy spans Traditional \gls{ml}, \gls{dl}, \gls{fm}, \gls{llm}, and \gls{mllm}.
Training is included as a reference row in Table~\ref{tab:centerpiece}, not as a post-training technique, because it establishes a baseline to contrast the inherent properties of post-training techniques.

\subsection{Methodology}
\label{sec:Methodology}
This study adopts a \gls{mlr} methodology. This approach is warranted because AI model adaptation is heavily driven by industry practice, with techniques evolving faster than traditional publication cycles. Emergent techniques often appear in framework documentation, developer blogs, and technical reports months or years before they reach formal academic publication. Consequently, restricting this review to peer-reviewed sources alone would omit rapidly evolving concepts that have achieved widespread practitioner adoption but currently lack formal academic treatment. 

Accordingly, we combined backward and forward snowballing from twelve canonical academic seed clusters (anchored by 18 seed papers) with grey literature from major framework documentation (Hugging Face \gls{peft}, mergekit, EasyEdit, ms-swift), developer reports, and industrial analysis. 
A candidate received a separate row when it represented an independently selectable adaptation decision unit and had a materially distinct six-dimensional profile from existing rows. 
Saturation was reached when additional candidates could be represented as variants, sub-techniques, or compositions of existing rows without losing a practically meaningful distinction.
\SuppMethodology presents the snowballing chains in thematic rather than integration order.

\subsection{Relationship to Existing Surveys}
\label{sec:RelatedSurveys}
Several authoritative surveys address subsets of the adaptation landscape covered by this taxonomy. Table~\ref{tab:survey-comparison} positions the present work relative to the most relevant existing treatments.
The dimensional counts in Table~\ref{tab:survey-comparison} for prior surveys depend on how one parses their taxonomies and reflect this paper's reading.
These approximate counts were coded by the authors using the stated dimension definitions and are intended for positioning rather than exact bibliometric comparison.

{\footnotesize
\begin{table}[htbp]
\caption{Comparison with existing surveys. Columns indicate techniques covered, classification dimensions, whether model type is a structuring axis, governance mapping, and model-type span.}
\label{tab:survey-comparison}
\begin{tabularx}{.75\linewidth}{lccccX}
\toprule
\textbf{Survey} & \textbf{Tech.} & \textbf{Dims.} & \textbf{\dimens{D}{6}} & \textbf{Gov.} & \textbf{Model Span} \\
\midrule 
Pan \& Yang~\cite{pan_survey_2010} & $\sim$8 & 1 & No & No & 
Trad.~\gls{ml}, early \gls{dl} \\
Zhuang et al.~\cite{zhuang_comprehensive_2021} & $\sim$15 & 1 & Partial & No & Trad.~\gls{ml}--\gls{dl} \\
Shi et al.~\cite{shi_continual_2024} & $\sim$12 & 2 & Partial & No & \gls{dl}--\gls{llm} \\
Liu et al.~\cite{liu_pre-train_2023} & $\sim$10 & 1 & No & No & \gls{llm} \\
Hospedales et al.~\cite{hospedales_meta-learning_2021} & $\sim$8 & 3 & No & No & \gls{dl} \\
Gao et al.~\cite{gao_retrieval-augmented_2023} & $\sim$6 & 1 & No & No & \gls{llm} \\
Wehner et al.~\cite{wehner_representation_2025} & $\sim$10 & 1 & No & No & \gls{llm} \\
\textbf{This work} & \textbf{\NbTech} & \textbf{6} & \textbf{Yes} & \textbf{Yes} & \textbf{Trad.~\gls{ml}--\gls{mllm}} \\
\bottomrule
\end{tabularx}
\end{table}
}

Compared with prior surveys, this taxonomy combines six-dimensional classification, coverage from Traditional \gls{ml} to \glspl{mllm}, and an explicit governance mapping.

\subsection{Paper Organization}
\label{sec:Organization}
Section~\ref{sec:FoundationalConcepts} establishes foundational concepts. Section~\ref{sec:Framework} presents the six-dimensional taxonomy with the centerpiece table. 
Section~\ref{sec:TechniqueDefinitions} (\SuppTech) provides 
definitions and characterizations of techniques organized by family. Section~\ref{sec:DecisionFramework} provides terminology navigation guidance. Section~\ref{sec:Governance} maps taxonomy dimensions to regulatory regimes. Section~\ref{sec:Conclusion} offers some conclusions, followed by a list of abbreviations and the bibliography. 
The Supplementary 
Materials
provide the 
detailed methodology (\SuppMethodology),
extended taxonomical category definitions (\SuppCategory),
technique definitions 
(\SuppTech),
exploratory structural consistency analysis 
(\SuppQuantitative),
and a
governance mapping 
(\SuppGov).
\section{Foundational Concepts: Training, Drift, and Retraining}
\label{sec:FoundationalConcepts}
This section establishes the baseline adaptation cycle, corresponding to Rows~\ref{tech:training} and~\ref{tech:retraining} of the centerpiece table (Table~\ref{tab:centerpiece}).
Model training is the process by which an algorithm learns to recognize patterns from data through iterative parameter adjustment to minimize a loss function~\cite{mitchell_machine_1997}. The objective is \textit{generalization}: accurate predictions on new, unseen data, rather than memorization of training examples.

Models degrade once deployed because real-world data diverges from the training distribution~\cite{gama_survey_2014,lu_learning_2019,sculley_hidden_2015}. 
The \gls{ml} community classifies this \textit{drift} by which property of the joint distribution changes~\cite{moreno-torres_unifying_2012}. \textbf{Covariate shift} (or virtual drift) changes the input feature distribution while the 
underlying relationship between inputs and outputs
remains stable~\cite{shimodaira_improving_2000}; it can often be detected with unsupervised drift monitoring~\cite{dos_reis_fast_2016,rabanser_failing_2019}. \textbf{Concept drift} (or real drift) changes 
the input-output mapping
itself, invalidating the model's learned behavior~\cite{widmer_learning_1996}; remediation typically requires labeled data and is costly~\cite{lu_learning_2019}. 
At the \gls{fm} level, degradation often manifests as \textbf{prompt drift}, where user interaction patterns evolve \cite{seegmiller_measuring_2026,wang_detecting_2025}, or \textbf{alignment drift}, where the model's static behavior diverges from shifting safety or cultural standards \cite{das_tracealign_2025}. 
Drift detection is fundamentally a governance practice. The \gls{nist} \gls{ai} \gls{rmf} explicitly treats the failure to monitor for drift as a threat to safety and reliability, making retraining not just a technical optimization, but a necessary compliance control~\cite{noauthor_nist_2023, noauthor_taxonomy_nodate}.

Model retraining updates a previously trained model on newer or additional data and is a central component of \gls{mlops} and \gls{ct} methodologies~\cite{steidl_pipeline_2023,testi_mlops_2022}. 
In the six-dimensional taxonomy, retraining occupies the profile 
\profile{\tax{1}{full}}{\tax{2}{DriftRemed}}{\tax{3}{seqIncD}}{\taxNo{4}{SchedAdhocPerm}}{\tax{5}{whole}}{\taxNo{6}{all}};
Its universality across model types reflects its status as the most general-purpose adaptation technique. Triggers include scheduled intervals, performance degradation, drift detection~\cite{ayers_detecting_2025,lu_early_2025,jourdan_handling_2023,poenaru-olaru_retrain_2023}, upstream pipeline changes, and manual decisions. Data strategy varies along a stability--adaptability spectrum from offline batch retraining (full corpus, mitigates catastrophic forgetting~\cite{mccloskey_catastrophic_1989,bertsimas_towards_2025}) through windowed learning to online learning~\cite{hoi_online_2021} and continual learning~\cite{wang_comprehensive_2024}.

Initial training and retraining share identical \tax{1}{full} mechanisms but differ strictly on their intended goals 
(\tax{2}{build} versus \tax{2}{DriftRemed})
and data requirements (\tax{3}{largeD} versus 
\tax{3}{seqIncD}).
The boundary between retraining and continual learning blurs at the incremental end: when retraining uses constrained parameter updates to prevent catastrophic forgetting, it shifts from a 
\taxNo{4}{SchedAdhocPerm}, \tax{5}{whole}
change to a 
\taxNo{4}{boundedUnboundedCumul}, \tax{5}{partl}
one, functionally becoming continual learning (Row~\ref{tech:cl}, Table~\ref{tab:centerpiece}).

\section{The Multi-Dimensional Adaptation \FrameTax}
\label{sec:Framework}
This section presents the six-dimensional 
\frameTax for \gls{ml} model adaptation. Rather than forcing each adaptation technique into a single slot within a hierarchical tree, the \frameTax defines six conceptually independent axes and positions every technique at a specific coordinate across all six. The dimensions are not empirically orthogonal, since certain pairings are far more common than others, 
but each answers a distinct question that cannot be derived from the others, and the non-canonical pairings are precisely where the \frameTax provides its greatest 
analytical value. 

\subsection{\FrameTax Overview}
\label{sec:FrameworkOverview}

\subsubsection{Motivation}
\label{sec:WhySingleFail}
Single-axis taxonomies organize adaptation by one property, e.g., mechanism, domain, or lifecycle stage. They are useful for local comparison but cannot represent techniques that span several functions.
For example, \gls{peft} is simultaneously a \gls{ft} approach, an efficiency strategy, and a model-adaptation method.
Single-axis schemes also obscure distinctions among methods with a shared mechanism. \gls{pe} and \gls{rag} both use Context Injection, while their persistence depends on whether a per-request context or a maintained production configuration is being described.
They also risk conflating technique properties with target-model properties.
\gls{ft} applied to a ResNet and \gls{ft} applied to 
an \gls{llm}
share a mechanism (\dimens{D}{1}) but may differ in data requirements (\dimens{D}{3}), persistence patterns (\dimens{D}{4}), and complementary adaptation layers. 
The resulting differences are captured through the remaining dimensions
\dimens{D}{6} represents the role of model type in constraining practical applicability.

The \frameTax addresses these limitations by assigning each practical adaptation technique a six-dimensional profile.
\subsubsection{The Six Dimensions}
\label{sec:FiveDimensions}
Table~\ref{tab:dimensions} summarizes five \textit{characterizing} dimensions and one \textit{modulating} dimension. \dimens{D}{1}--\dimens{D}{5} describe the canonical profile of the declared intervention rather than immutable properties of every implementation. \dimens{D}{6} describes the target model and constrains practical applicability. This separation reduces confusion between technique properties and model properties.

{\footnotesize
\begin{table}[htbp]
\centering
\setlength{\tabcolsep}{6pt}
\renewcommand{\tabularxcolumn}[1]{p{#1}}
\caption{The six dimensions: definition and regulatory documentation relevance. 
\dimens{D}{1}--\dimens{D}{5} are characterizing dimensions describing 
canonical properties of the adaptation intervention;
\dimens{D}{6} is a modulating dimension that filters feasible \dimens{D}{1}--\dimens{D}{5} combinations. 
Documentation roles are treated in detail in Section~\ref{sec:Governance}.}
\label{tab:dimensions}
\adjustbox{width=.9\linewidth
}{
\begin{tabularx}{\linewidth}{p{.02\linewidth} p{.1\linewidth} p{.2\linewidth} p{.1\linewidth} X}
\toprule
\textbf{\#} & \textbf{Dimension} & \textbf{Core Question} & \textbf{Role} & \textbf{Regulatory documentation relevance} \\
\midrule
\dimens{D}{1} & Mechanism & \textit{What changes?} & Characterizing & Describes the technical modification: weights, adapters, retrieval context, prompts, calibration layer, activation path, or data pipeline. \\
\dimens{D}{2} & Goal & \textit{Why adapt?} & Characterizing & Links the change to intended purpose and risk rationale: performance maintenance, specialization, alignment, reasoning, knowledge correction, privacy, robustness, or knowledge removal. \\
\dimens{D}{3} & Data \mbox{Requirements} & \textit{What data?} & Characterizing & Supports data provenance and evidence traceability: labeled, unlabeled, preference pairs, verifiable task data, external corpus, or forget set. \\
\dimens{D}{4} & Persistence & \textit{How long the change lasts?} & Characterizing & May inform temporal change-control analysis: permanent versioning, cumulative monitoring, transient rollbacks, or ephemeral sessions. \\
\dimens{D}{5} & Scope & \textit{How much of the model structure is modified?} & Characterizing & Can support structural modification impact assessments: whole-model, partial, modular/swappable, derived artifacts, or distributed. \\
\dimens{D}{6} & Model Type & \textit{What model?} & Modulating & Defines baseline capability limits and applicability of \gls{gpai} regulatory tiers. Prevents misapplication of model-specific terminology, e.g., \gls{llm}-specific terms used for non-\gls{llm} systems and vice versa. \\
\bottomrule
\end{tabularx}
}
\end{table}
}
\subsubsection{How to Read the \FrameTax}
\label{sec:HowToRead}
Each practical adaptation decision unit is represented by a six-element profile,
(\dimens{d}{1}, \dimens{d}{2}, \dimens{d}{3}, \dimens{d}{4}, \dimens{d}{5}, \dimens{d}{6}),
where 
\dimens{d}{i}
denotes the technique's position on Dimension~$i$, i.e., 
\dimens{d}{i} $\in$ \dimens{D}{i}, \ $i = 1, \dots, 6$.
For example, \gls{lora}-based \gls{peft} 
is characterized by
\profile{\tax{1}{sel}}{\taxNo{2}{taskCompEff}}{\tax{3}{smallD}}{\taxNo{4}{SchedAdhocPerm}}{\taxNo{5}{partlMod}}{\taxNo{6}{dflm}}.
The full set of profiles is compiled in Table~\ref{tab:centerpiece}.
Practitioners can enter the \frameTax from any dimension. A deployment engineer constrained by data availability enters through \dimens{D}{3}. 
A compliance officer evaluating governance risk can enter through \dimens{D}{2} (Goal) and \dimens{D}{4} (Persistence). 
A researcher comparing techniques might enter through \dimens{D}{1} (Mechanism). 
In all cases, \dimens{D}{6} acts as a preliminary filter: the practitioner first identifies their model type, which eliminates infeasible techniques before the remaining dimensions are consulted.
Sections~\ref{sec:D1Mechanism}--\ref{sec:D5ModelType} define the dimensions, and Table~\ref{tab:centerpiece}  organizes the resulting technique profiles into navigational families.

\subsubsection{Navigating Boundary Cases}
\label{sec:BoundaryCases}
Techniques marked with ~$\S$ in Table~\ref{tab:centerpiece} require conditional interpretation. Three recurring cases are recognized, i.e., a technique may span more than one category, may serve both a foundational and an adaptation role, or may remain technically available after being displaced by a more general method. The assigned profile should therefore be interpreted as canonical and should be replaced by the realized implementation when a deployment is documented.

Relations among rows are described as umbrella, sub-technique, bridge or hybrid, supersession, and complementarity. Umbrella paradigms contain several independently selectable interventions. Sub-techniques refine a parent row, bridges span families, superseded methods remain technically available but are less commonly selected, and complementary techniques may be layered in an ordered deployment stack. A row is retained when it represents an independently selected practical decision unit with a materially distinct profile.
Highly experimental cases may fall outside this representation, since the \frameTax prioritizes practical applicability over exhaustive theoretical coverage.

\subsection{\dMechanism}
\label{sec:D1Mechanism}
The most fundamental axis is what an adaptation technique modifies.  
Table~\ref{tab:D1} organizes the categories, e.g., parameter-updating techniques (full, selective, transient, or distributed updates), structural interventions (architectural modifications), manipulation techniques (parameter composition, transient activation steering, or data pipeline manipulation), and inference-time augmentations (context injection).
The 
\tax{1}{paramUpd}
category is the most populated, spanning classical \gls{ft}, \gls{peft}, prompt learning, alignment (\gls{rlhf}, \gls{dpo}), and reasoning-focused \gls{rl} (\gls{rlvr}). 
At the \gls{llm} level, the \frameTax introduces \tax{1}{infTimeSearch} to formalize test-time compute scaling methods (e.g., OpenAI o1, DeepSeek-R1) and \tax{1}{transAct} (activation steering), both of which are structurally distinct from gradient-based updates of 
\tax{1}{sel}
(\gls{tta}).
Across Tables~\ref{tab:D1}--\ref{tab:d5_scope}, statements about the relevance of regulatory documentation identify potential considerations rather than automatic legal consequences; the applicable obligation depends on the actual implementation, intended use, risk profile, and regulatory pathway.

Additionally, architectural modification isolates interventions that enact purely topological changes to the computational graph, such as dynamically routing tokens through newly initialized expert sub-networks, expressly excluding updates to pre-existing base weights. From a regulatory standpoint, distinguishing topological shifts from parameter updates may be important for EU AI Act substantial-modification analysis, as introducing new routing logic can alter the software's structural control flow even if the functional behavior of the base parameters remains unchanged.

{\footnotesize
\begin{table}[htbp]
\centering
\renewcommand{\tabularxcolumn}[1]{p{#1}}
\caption{\dMechanism Categories}
\label{tab:D1}
\adjustbox{width=.9\linewidth
}{
\begin{tabularx}{\linewidth}{p{.2\linewidth} p{.4\linewidth} X}
\toprule
\multicolumn{1}{c}{\textbf{Category}} &
\multicolumn{1}{c}{\textbf{Definition}} &
\multicolumn{1}{c}{\textbf{Representative Techniques}}
\\

\midrule
\multicolumn{3}{p{\linewidth}}{\cellcolor{gray!20} \scriptsize \textit{\textbf{I. 
Gradient-Based / Compute-Intensive:}
Computes gradients against stored weights, consuming training \glspl{flop}. 
In the context of the EU AI Act's ``substantial modification'', gradient-based methods are especially relevant to compute-threshold analyses.
May require or support \gls{mdr} Annex II-style regression evidence and \gls{fda} \gls{pccp} pre-specification, depending on device context.
}} \\
\midrule

\makebox[0pt][l]{\hypertarget{dim:paramUpd}{}} \paramUpd
& 
Executes \gls{gd} to alter model weights.
& Training, Retraining, Full/Partial \gls{ft}, \gls{peft}, \gls{mtl}, \gls{ssl}, Prompt Learning, \gls{rlhf}, \gls{dpo}, \gls{rlvr}, \gls{cl}, Meta-Learning, Calibration \\
  
\makebox[0pt][l]{\hypertarget{dim:CrossModelTransf}{}}
\CrossModelTransf 
& \CrossModelTransfConcise & \gls{kd} \\
\midrule
\multicolumn{3}{p{\linewidth}}{\textit{\cellcolor{gray!20} \scriptsize \textbf{II. 
Gradient-Free / Weight Manipulation:}
Alters weights or computational graphs without \gls{gd}. 
May fall outside compute-based threshold proxies despite causing substantial behavioral or structural changes (e.g., altering control flow or creating compressed artifacts).}} \\
\midrule

\makebox[0pt][l]{\hypertarget{dim:WeightSpaceComp}{}}
\WeightSpaceComp 
& \WeightSpaceCompConcise & Task Arithmetic, Model Merging, \gls{leace}, Knowledge Editing (closed-form) \\

\makebox[0pt][l]{\hypertarget{dim:StructReduct}{}}
\StructReduct 
& \StructReductConcise & Compression \\

\makebox[0pt][l]{\hypertarget{dim:arch}{}}
\arch 
& 
Topological alteration of the computational graph (e.g., adding nodes/routing) expressly without modifying existing base weights. 
&\gls{moe} adaptation, dynamic routing \\
\midrule
\multicolumn{3}{p{\linewidth}}{\cellcolor{gray!20} \textit{\scriptsize \textbf{III. 
Inference-Time / Zero-Footprint:}
Operates during inference without weight modification and may therefore fall outside weight-update or training-compute thresholds.
However, Version-Persistent artifacts (e.g., \gls{rag} corpora) may still require system-level transparency documentation under EU AI Act Art. 13 and \gls{fda} \gls{pccp}.
}} \\
\midrule

\makebox[0pt][l]{\hypertarget{dim:ctx}{}}
\ctx 
& Input-context conditioning at inference; zero weight modification &\gls{pe}, \gls{rag}, \gls{icl}, \gls{ce} \\
\makebox[0pt][l]{\hypertarget{dim:transAct}{}}
\transAct 
& 
Direct algebraic manipulation of intermediate forward-pass activations without weight modification.
&Activation Steering \\

\makebox[0pt][l]{\hypertarget{dim:infTimeSearch}{}}
\infTimeSearch 
& \infTimeSearchConcise & \gls{cot} search, Tree of Thoughts, Best-of-$n$ sampling \\
\midrule
\multicolumn{3}{p{\linewidth}}{\cellcolor{gray!20} \textit{\scriptsize \textbf{IV. 
Pipeline-Mediated:}
Adaptation occurs strictly via data pipeline manipulation. 
Regulatory and change-control documentation is often primarily associated with the downstream training method, although pipeline changes may also require documentation.
}} \\
\midrule
\makebox[0pt][l]{\hypertarget{dim:manip}{}}
\manip 
& Characterizes the indirect adaptation pathway for boundary-extension techniques. & Augmentation, Active Learning, Curriculum Learning, \gls{semiSL}
\\
\bottomrule
\multicolumn{3}{p{\linewidth}}{\scriptsize $^\dagger$ \gls{ssl}-as-adaptation (\gls{cpt}) prescribes its own training objective and modifies parameters; see Table~\ref{tab:centerpiece}, Row~\ref{tech:ssl}.}\\ 
\multicolumn{3}{p{\linewidth}}{\scriptsize \textbf{Note:} The \textit{\manip} category explicitly characterizes the indirect adaptation pathway for boundary-extension techniques, modifying the data pipeline rather than directly altering model parameters.}
\end{tabularx}
}
\end{table}
}
\dimens{D}{1} determines computational cost, reversibility, and infrastructure requirements. As models grow, the distinction between parameter-modifying and non-parameter-modifying methods becomes increasingly consequential: inference-time methods scale independently of model size, while training-time methods scale with it. 
Consequently, \dimens{D}{1} is a key technical axis for assessing whether an adaptation event may implicate compute-based thresholds, such as the EU AI Act’s \gls{gpai} guidelines, thereby highlighting the regulatory divergence between parameter-updating methods and inference-time context injection.
Certain techniques exhibit composite \dimens{D}{1} classifications, e.g., \taxNo{1}{archSel} for \gls{moe} adaptation.
 
It is important to note that algorithmic training protocols, specifically the modification of loss functions, reward shaping, or optimization constraints, e.g., Lagrangian-constrained \gls{ppo}, are intentionally excluded from \dimens{D}{1} categories. While altering an optimization objective is a highly consequential regulatory event, it represents a change to the mathematical protocol calculating the error, rather than the structural mechanism of the modification itself. Regardless of how a safety constraint shapes the gradient, the physical mechanism applied to the model artifact remains a parameter update (e.g., \tax{1}{sel}). Consequently, shifts in optimization objectives are captured in this \frameTax through shifts in the corresponding \dimens{D}{2} (Goal) and \dimens{D}{3} (Data Requirements) coordinates, paired with external training protocol documentation.

\subsection{\dGoal}
\label{sec:D2Goal}
This dimension captures the practitioner's intent, the problem that motivates the choice of adaptation technique. It is the natural entry point for most practitioners, who begin not with ``what mechanism should I use?'' but with ``what problem am I trying to solve?'' 
The goal categories are defined in Table~\ref{tab:D2}.

{\scriptsize
\begin{table}[htbp]
\centering
\renewcommand{\tabularxcolumn}[1]{p{#1}}
\caption{\dGoal Categories}
\label{tab:D2}
\adjustbox{width=.9\linewidth
}{
\begin{tabularx}{\linewidth}{ p{.15\linewidth} p{.5\linewidth} X}
\toprule
\multicolumn{1}{c}{\textbf{Category}} &
\multicolumn{1}{c}{\textbf{Definition}} &
\multicolumn{1}{c}{\textbf{Representative Techniques}}
\\
\midrule
\multicolumn{3}{p{\linewidth}}{\cellcolor{gray!20} \textit{\scriptsize \textbf{I. Capability:} Expands or redefines system functionality. 
Capability-expanding changes may trigger or warrant ``substantial modification'' reassessments and intended-use review
under EU AI Act Annex IV and \gls{mdr} Annex I. 
Under \gls{fda} \gls{pccp}, capability boundaries may need to be pre-specified to avoid unapproved scope expansion.
}} \\
\midrule
\makebox[0pt][l]{\hypertarget{dim:build}{}}
\build 
& Establish baseline capability from training data. &Training \\ 

\makebox[0pt][l]{\hypertarget{dim:task}{}}
\task 
& Narrow model expertise to a specific target task. & \gls{ft}, \gls{peft}, \gls{fsl}, \gls{mtl}, \gls{sft}, Active Learning  \\

\makebox[0pt][l]{\hypertarget{dim:capExt}{}}
\capExt 
& Broaden model's operational scope by adding fundamentally new structural capabilities or modalities. &  \gls{cpt}, \gls{moe}, Cross-modal alignment, Multimodal tuning \\
\makebox[0pt][l]{\hypertarget{dim:reas}{}}
\reas 
& 
Enhance multi-step problem-solving, planning, or search capabilities. Often verified through programmatic oracles, simulators, or iterative self-competition.
&\gls{rlvr}/\gls{grpo} \\ 
\midrule
\multicolumn{3}{p{\linewidth}}{\cellcolor{gray!20} \textit{\scriptsize \textbf{II. Lifecycle:} Responds to deployment shifts rather than adding functionality. 
Lifecycle changes may be relevant to \gls{pms} obligations under EU AI Act Art. 72 and \gls{mdr} Arts. 83–84. 
Remediation events may need to be logged as part of \gls{pms} records; bridging distributional gaps (new populations) may require clinical-evaluation updates, especially where intended population or performance claims change.
}} \\
\midrule
\makebox[0pt][l]{\hypertarget{dim:DriftRemed}{}}
\DriftRemed 
& Restore performance degraded by distribution shift. &Retraining, \gls{tta} \\
\makebox[0pt][l]{\hypertarget{dim:ContAdapt}{}}
\ContAdapt 
& Accumulate new knowledge without forgetting prior tasks. & \gls{cl}, Incremental Learning \\

\makebox[0pt][l]{\hypertarget{dim:domAdapt}{}}
\domAdapt$^\ast$ 
& Bridge source-to-target distribution gap (domain shift), same task class. & \gls{da}, Active Learning \\
\midrule
\multicolumn{3}{p{\linewidth}}{\cellcolor{gray!20} \textit{\scriptsize \textbf{III. Trust:} Relates to EU AI Act Chapter III (Arts. 9–15) requirements for risk, bias, transparency, accuracy, and human oversight. 
``Trust''-related adaptations may be relevant to conformity-assessment review, even where the technical mechanism appears low-impact.
}} \\
\midrule
\makebox[0pt][l]{\hypertarget{dim:algn}{}}
\algn 
& 
Conform model behavior to human preferences, tone, and helpfulness.
&\gls{rlhf}, \gls{dpo}, \gls{sft} \\
\makebox[0pt][l]{\hypertarget{dim:safety}{}}
\safety 
& 
Reduce the production of harmful, dangerous, or policy-violating outputs through constitutional-principle enforcement, refusal training, or harm-category filtering. 
& 
Safety \gls{sft}, \gls{rlhf} (Refusal), \gls{dpo} (Refusal), Activation Steering \\
\makebox[0pt][l]{\hypertarget{dim:rob}{}}
\rob 
& Harden model against adversarial or out-of-distribution inputs. & Adversarial Training \\
\makebox[0pt][l]{\hypertarget{dim:explain}{}}
\explain 
& Adapting a model or appending modules to expose internal representations, align reasoning with human logic, or generate post-hoc rationales to satisfy regulatory transparency mandates. & Probing, Concept Bottlenecks, Attention Alignment \\

\makebox[0pt][l]{\hypertarget{dim:fairness}{}}
\fairness 
& Reduce systematic performance or output disparities across demographic or otherwise protected groups (e.g., counterfactual data augmentation, concept erasure). & \gls{leace}\\

\makebox[0pt][l]{\hypertarget{dim:reliability}{}}
\reliability 
& Align expressed model confidence with empirical correctness to support risk-based decision thresholds (e.g., temperature scaling, conformal wrappers).& Calibration \\
\midrule
\multicolumn{3}{p{\linewidth}}{\cellcolor{gray!20} \textit{\scriptsize \textbf{IV. Knowledge Operations:} 
The operational locus for \gls{gdpr} compliance. 
Knowledge-operation techniques may support workflows relevant to \gls{gdpr} Art. 17 and data-provenance disclosures.
May require updated data-governance documentation under EU AI Act Art. 10.}} \\
\midrule
\makebox[0pt][l]{\hypertarget{dim:augK}{}}
\augK 
& Inject, update, or correct model knowledge. &\gls{rag}, \gls{pe}, \gls{icl}, \gls{ce}, Knowledge Editing \\ 

\makebox[0pt][l]{\hypertarget{dim:rem}{}}
\rem 
& Selectively erase targeted factual or behavioral knowledge. & Machine Unlearning \\
\midrule
\multicolumn{3}{p{\linewidth}}{\cellcolor{gray!20} \textit{\scriptsize \textbf{V. Operational Constraints:} Alters deployment profile without changing core tasks, potentially implicating compliance considerations such as:
\gls{gdpr} Art. 25 (privacy), \gls{gdpr} Art. 22 (personalization), AI Act \gls{gpai} (compute threshold rules), and AI Act Art. 13 (transparency). 
May requires distinct configuration-management records.}} \\
\midrule

\makebox[0pt][l]{\hypertarget{dim:compEff}{}}
\compEff 
& Reduce deployment cost, size, or inference latency. &\gls{kd}, Compression, \gls{moe}, Task Arith. / Merging \\

\makebox[0pt][l]{\hypertarget{dim:userPerson}{}}
\userPerson 
& Tailor outputs to individual user preferences and history. &
User-embedding personalization, federated personalization, prefix-tuning per-user \\

\makebox[0pt][l]{\hypertarget{dim:priv}{}}
\priv 
& Train collaboratively without centralizing sensitive data. &\gls{fl} \\

\makebox[0pt][l]{\hypertarget{dim:behavControl}{}}
\behavControl 
& Constrain output format, style, persona, or procedural behavior without targeting underlying knowledge or task capability (e.g., system prompts, structured-output steering). & \gls{pe}\\

\bottomrule
\multicolumn{3}{p{\linewidth}}{\scriptsize \textbf{Note:} \textit{Build Capability} represents the initial training phase; it acts as the baseline anchor from which all post-training goals depart.}\\
\multicolumn{3}{p{\linewidth}}{\scriptsize{
$^\ast$
At the \taxNo{6}{lm} tier, the dedicated \gls{da} \textit{technique} (Row~\ref{tech:da}) is superseded by \gls{cpt} and domain-specific \gls{peft}; however, the \tax{2}{domAdapt} \textit{goal} remains achievable at these tiers via \gls{ssl}/\gls{cpt} (Row~\ref{tech:ssl}).
}}
\end{tabularx}
}
\end{table}
}
\dimens{D}{2} is intentionally practitioner-facing and spans technical, operational, and governance motivations.
To support ease of review and governance mapping, the goal categories in Table~\ref{tab:D2} are organized into five meta-categories, i.e., Capability, Lifecycle, Trust, Knowledge Operations, and Operational Constraints. Regulatory frameworks such as the \gls{nist} \gls{ai} \gls{rmf} and the EU AI Act express requirements in terms of these high-level goals instead of technical mechanisms.
Therefore, \dimens{D}{2} can serve as a technical indicator relevant to intended-purpose or intended-use analysis, providing the critical bridge between the technical taxonomy and the regulatory landscape (Section~\ref{sec:Governance}). 
Crucially, the \frameTax formally recognizes \textit{Explainability} as a standalone adaptation goal. Under frameworks like the EU AI Act Article 13 and the \gls{fda} \gls{pccp} transparency guidance, 
developers may face transparency or explainability obligations in high-risk contexts.
Post-training modifications engineered specifically to satisfy these transparency mandates, e.g., appending probing classifiers, concept bottleneck layers, or attention-alignment fine-tuning, serve an explicit explainability goal that is legally and mechanistically distinct from general \tax{2}{task} or \tax{2}{algn}.

\subsection{\dData}
\label{sec:D3Data}
Data availability is frequently the binding constraint in practice; a practitioner with abundant labeled data faces a fundamentally different set of feasible techniques than one with a handful of examples or no training data at all. 
Crucially, the boundary between a \tax{3}{smallD} and a \tax{3}{largeD} operates as an order-of-magnitude convention relative to the target model and task complexity, rather than an absolute numerical threshold.
Table~\ref{tab:D3} categorizes these distinct data regimes, ranging from traditional \tax{3}{smallD} through specialized inputs like \tax{3}{multiD}, down to data-free interventions that strictly require \tax{3}{wght} or \tax{3}{zeroSh}.

Beyond feasibility, \dimens{D}{3} helps characterize the regulatory data-provenance burden. 
Relying on \tax{3}{pref} introduces annotator-bias governance, e.g., under the EU AI Act; 
querying an \tax{3}{ext} may require copyright, licensing, and provenance controls; and providing a \tax{3}{forget} may implicate privacy-compliance obligations, e.g., the \gls{gdpr} "right to be forgotten".
To maintain taxonomic precision, \tax{3}{userInteract} is bounded by a strict operational and regulatory definition that isolates it from overlapping categories~\cite{noauthor_regulation_2024}. Mechanistically, it is restricted to raw, production-phase behavioral exhaust or implicit telemetry directly generated by an end-user during active deployment, e.g., historical query logs, correction clicks, session paths~\cite{health_marketing_2025}. It is explicitly distinguished from: (i)~\tax{3}{pref}, which requires comparative, explicit human evaluation labels typically gathered in offline curation; (ii)~\tax{3}{ext}, which constitutes a pre-existing, structurally static reference repository independent of the active user session; and (iii)~\tax{3}{smallD}, which relies on human annotators adding intentional task labels to raw data. Crucially, this distinction is enforced by its regulatory profile: under the \gls{gdpr},
\tax{3}{userInteract} 
may carry heightened obligations concerning lawfulness, purpose limitation, and transparency, including the information duties in Articles 13 and 14
that do not apply to curated, non-identifiable training datasets~\cite{noauthor_regulation_2024}.

{\scriptsize
\begin{table}[htbp]
\centering
\renewcommand{\tabularxcolumn}[1]{p{#1}}
\caption{\dData Categories}
\label{tab:D3}
\adjustbox{width=.9\linewidth
}{
\begin{tabularx}{\linewidth}{p{.15\linewidth} p{.5\linewidth} X}
\toprule
\multicolumn{1}{c}{\textbf{Category}} &
\multicolumn{1}{c}{\textbf{Definition}} &
\multicolumn{1}{c}{\textbf{Representative Techniques}}
\\
\midrule
\multicolumn{3}{p{\linewidth}}{\cellcolor{gray!20} \textit{\scriptsize \textbf{I. Human-Annotated Data:} Produced via explicit human judgment. Regulatory burden centers on annotator governance, labeling guidelines, and inter-rater reliability. 
May require documentation of demographic representation and bias mitigation under EU AI Act Art. 10(2)(f–g), plus \gls{gdpr} Arts. 13/14 disclosures if annotations derive from personal data.}}\\
\midrule
\makebox[0pt][l]{\hypertarget{dim:largeD}{}}
\largeD 
& Thousands-to-millions of annotated task examples. & Full \gls{ft}, \gls{mtl}, Adversarial Training \\ 

\makebox[0pt][l]{\hypertarget{dim:smallD}{}}
\smallD 
& Tens-to-hundreds of labeled task examples. & Partial \gls{ft}, \gls{peft}, gradient-based \gls{fsl}, Knowledge Editing \\

\makebox[0pt][l]{\hypertarget{dim:few}{}}
\few 
& One-to-five in-context demonstration examples. & \gls{icl}, metric-based \gls{fsl} \\

\makebox[0pt][l]{\hypertarget{dim:pref}{}}
\pref 
& Human-ranked response pairs encoding behavioral preferences. & \gls{dpo}, \gls{rlhf} \\ 

\makebox[0pt][l]{\hypertarget{dim:taskDist}{}}
\taskDist 
& Distribution of varied tasks for meta-learning episodes or or multi-task training. & Meta-Learning, \gls{mtl} \\
\midrule
\multicolumn{3}{p{\linewidth}}{\cellcolor{gray!20} \textit{\scriptsize \textbf{II. Synthetic / Programmatic Data:} Generated by upstream AI models, simulators, or programmatic rules. Compliance shifts from annotator oversight to provenance tracing of the generating artifact (e.g., teacher biases, constitutional text). Under EU AI Act \gls{gpai} rules, teacher-model risks may propagate to the student and should be assessed.
}}\\
\midrule

\makebox[0pt][l]{\hypertarget{dim:synthTeacher}{}}
\synthTeacher 
& Data generated by a more capable model acting as an annotator or oracle. Requires auditing the provenance and biases of the upstream teacher model. & \gls{kd} \\ 

\makebox[0pt][l]{\hypertarget{dim:synthRule}{}}
\synthRule 
& Data generated iteratively based on explicit textual guidelines or constitutional principles. Requires auditing the constitutional text as a data-governance artifact. & \gls{cai}, \gls{rlaif} \\

\makebox[0pt][l]{\hypertarget{dim:synthEnv}{}}
\synthEnv 
& 
Data generated via a physics engine, simulator, play environment, or a model's own autoregressive self-generation (as an opponent).Requires auditing the simulation parameters and reward mechanics. 
& Self-Play, Classic \gls{rl} \\

\makebox[0pt][l]{\hypertarget{dim:instrResp}{}}
\instrResp 
& 
Heterogeneous prompt-response pairs spanning many tasks, often mixed human-written and synthetic; provenance documentation must cover licensing of scraped sources and the share of model-generated data.
& \gls{sft} (Instruction Tuning) \\
\midrule
\multicolumn{3}{p{\linewidth}}{\cellcolor{gray!20} \textit{\scriptsize \textbf{III. 
Live / Production Data:} 
Generated during active deployment. 
Depending on whether it involves logged interactions or live inference, this data may implicate \gls{gdpr} Arts. 6 and 13/14 (consent, lawfulness, purpose limitation) with non-delegable force, or carry no persistent provenance burden but require documented temporal boundaries for on-the-fly adaptations.}}\\
\midrule
\makebox[0pt][l]{\hypertarget{dim:userInteract}{}}
\userInteract 
& Logged user behavior signals collected during deployment. & Personalization methods, implicit feedback loops \\

\makebox[0pt][l]{\hypertarget{dim:tstOnly}{}}
\tstOnly 
& 
Live, unlabeled test instances encountered on-the-fly during active inference.
& \gls{tta} \\

\makebox[0pt][l]{\hypertarget{dim:seqIncD}{}}
\seqIncD$^\ast$ 
& Time-ordered arriving data batches over a model's lifecycle. & Retraining, \gls{cl} \\ 
\midrule
\multicolumn{3}{p{\linewidth}}{\cellcolor{gray!20} \textit{\scriptsize \textbf{IV. 
External / Targeted Data:} 
Depending on whether the intervention targets an external knowledge base or specific records flagged for deletion, it either requires copyright, licensing, and quality documentation under EU AI Act Art. 13 transparency rules, or may implicate \gls{gdpr} Art. 17 (Right to Erasure) analysis and may require verification of genuine record suppression rather than mere obfuscation.
}}\\
\midrule
\makebox[0pt][l]{\hypertarget{dim:ext}{}}
\ext 
& Queried external knowledge base at inference time. & \gls{rag}, \gls{ce} \\
\makebox[0pt][l]{\hypertarget{dim:forget}{}}
\forget 
& Designated records or facts targeted for erasure. & Machine Unlearning \\
\midrule
\multicolumn{3}{p{\linewidth}}{\cellcolor{gray!20} \textit{\scriptsize \textbf{V. 
Minimal / Decentralized / Delegated Data:} 
Depending on the specific configuration, these approaches carry varying provenance burdens: they may entail the lowest burden as no new data enters the pipeline, limit centralized traceability, and implicate data-sovereignty and federation rules (e.g., \gls{gdpr} Ch. V cross-border transfers; HIPAA), or fully delegate regulatory obligations to downstream training methods.
}}\\
\midrule
\makebox[0pt][l]{\hypertarget{dim:zeroSh}{}}
\zeroSh 
&  Natural language instructions only; no labeled examples. & \gls{pe} \\ 

\makebox[0pt][l]{\hypertarget{dim:wght}{}}
\wght 
& No data; operates exclusively on model parameters. & Task Arithmetic, Model Merging, Compression (data-free) \\

\makebox[0pt][l]{\hypertarget{dim:decent}{}}
\decent 
& Private data distributed across non-collocated clients. & \gls{fl} \\

\makebox[0pt][l]{\hypertarget{dim:pipeInherit}{}}
\pipeInherit 
& The effective data regime is fully inherited from the downstream training method being complemented.  & Data Augmentation, Curriculum Learning, Active Learning \\

\makebox[0pt][l]{\hypertarget{dim:unlabD}{}}
\unlabD
$^\dagger$ 
& Unannotated data used for self-supervised training, continuous adaptation, or distribution sampling for post-training calibration. & \gls{ssl}, \gls{semiSL}, Compression (calibration), Calibration \\ 
\midrule
\multicolumn{3}{p{\linewidth}}{\cellcolor{gray!20} \textit{\scriptsize \textbf{VI. Verifier-Defined Ground Truth:} Ground truth is established by deterministic programmatic verifiers (e.g., unit tests, math checkers, simulators) rather than human judgment or synthetic generation. This shifts the regulatory burden from annotator-bias tracking to verifier-validity documentation. 
The verification function acts as a ``test oracle'' and may require documented validation of the verifier’s soundness and limitations under EU AI Act Art. 15 and \gls{fda}/IEC 62304 software-validation standards.
}}\\
\midrule

\makebox[0pt][l]{\hypertarget{dim:verify}{}}
\verify 
& Tasks with programmatically assessable ground truth. & \gls{rlvr}/\gls{grpo} \\
\midrule
\multicolumn{3}{p{\linewidth}}{\cellcolor{gray!20} \textit{\scriptsize \textbf{VII. Multi-Modal Paired Provenance:} Each training example carries independent provenance chains, requiring separate licensing, copyright, and bias documentation for each individual modality. 
The cross-modal alignment itself should be audited or may need to be assessed for introduced biases. 
If the pairing involves biometric data used for identification or other special-category data, it may implicate \gls{gdpr} Art. 9.
Furthermore, 
EU AI Act data-governance provisions may require relevance and representativeness documentation, potentially per modality or pairing,
significantly multiplying the compliance burden.}}\\
\midrule
\makebox[0pt][l]{\hypertarget{dim:multiD}{}}
\multiD
& Aligned cross-modal pairs (e.g., image-text, audio-text). & Cross-modal alignment, Multimodal instr. tuning \\

\bottomrule
\multicolumn{3}{p{\linewidth}}{\scriptsize $^\ast$ Beyond any applicable \gls{gdpr} purpose-limitation concerns, 
each task or domain boundary should generally be logged as a lifecycle event where \gls{pms} or post-market monitoring obligations apply.
}\\
\multicolumn{3}{p{\linewidth}}{\scriptsize $^\dagger$ When the unlabeled sample is drawn from personal health records or user behavioral logs, \gls{gdpr} Articles 9 (special categories of personal data) and 13/14 obligations apply, elevating the provenance burden irrespective of the absence of explicit labels.}
\\
\end{tabularx}
}
\end{table}
}
Similarly, the \frameTax uses operational boundaries to separate overlapping unlabeled data regimes: offline batch samples used to calibrate weights or steering vectors before deployment are classified as  
\taxNo{3}{unlabSmallDAbbr}, 
whereas live queries driving on-the-fly updates are classified as a \tax{3}{tstOnly}.

The \dimens{D}{3} dimension also illustrates the interaction with target models (\dimens{D}{6}): \tax{3}{multiD} 
emerges at the \tax{6}{fm} level for 
\glspl{repfm}
(e.g., 
\gls{clip}~\cite{radford_learning_2021}), 
but it becomes a dominant, defining operational constraint at the \tax{6}{mllm} level.

Crucially, a shift in \dimens{D}{3} can expand or narrow a technique's operational profile even when structural mechanics are identical. For example, while \gls{pe} and \gls{icl} share identical structural coordinates (\tax{1}{ctx}, \tax{5}{extContxt}, \taxNo{6}{lm}), their differing data requirements (\tax{3}{zeroSh} versus \tax{3}{few}) drive distinct variations in their goal and persistence profiles. While both overlap by targeting \tax{2}{augK} within a \tax{4}{eph} context, \gls{pe}'s profile broadens to include \tax{2}{behavControl} and \tax{4}{verPersistent} artifacts, whereas \gls{icl} uniquely introduces \tax{2}{task}.

\subsection{\dPersistence}
\label{sec:D4_Persistence}

Dimension 4 characterizes the temporal lifespan and operational durability of an adaptation. 
A central motivation for isolating this dimension is the regulatory and operational requirement for \textit{model consistency}, defined here as the preservation or controlled modification of model behavior, performance, output characteristics, and intended use, and the maintenance of relevant reliability across versions, updates, deployment contexts, and adaptation layers. Consistency in this sense is not a synonym for stability or accuracy: a model can remain stable under noise while becoming inconsistent across versions and can remain accurate on a benchmark while exhibiting inconsistent behavior on edge cases that matter for regulatory or clinical use.
Rather than evaluating structural elements, \dimens{D}{4} isolates the state-retention behavior of the model across inference sessions and operational cycles. 
From a regulatory perspective, this dimension is a primary  
technical input to version-control and \gls{pms} strategy.
For example, a modification that alters parameters across multiple deployment intervals 
may require validation against behavioral drift, whereas transient shifts may require validation of initialization and reset boundaries.
To provide clear taxonomic and compliance precision, we categorize the temporal states of adaptation into distinct operational profiles, as detailed in Table~\ref{tab:d4_persistence}.

{\footnotesize
\begin{table}[hbt!]
\centering
\renewcommand{\tabularxcolumn}[1]{p{#1}}
\caption{\dPersistence Categories}
\label{tab:d4_persistence}
\renewcommand{\arraystretch}{1.3}
\adjustbox{width=.9\linewidth
}{
\begin{tabularx}{\linewidth}{p{.15\linewidth} p{.5\linewidth} X}
\toprule
\multicolumn{1}{c}{\textbf{Category}} &
\multicolumn{1}{c}{\textbf{Definition}} &
\multicolumn{1}{c}{\textbf{Representative Techniques}}
\\
\midrule
\multicolumn{3}{p{\linewidth}}{\cellcolor{gray!20} \textit{\scriptsize \textbf{I. Pre-Specifiable Baselines:} Scope, cadence, and boundaries can be fully declared pre-deployment in a \gls{pccp} or technical documentation. 
Depending on the intervention, it may declare a fixed temporal cadence, a finite task boundary, or non-parametric system updates—the latter requiring corpus or prompt validation protocols rather than parametric regression testing.
}}\\
\midrule

\makebox[0pt][l]{\hypertarget{dim:ScheduledPerm}{}}
\ScheduledPerm 
& Pre-planned, periodic parameter updates executed on a fixed temporal or data-volume cadence. Eligible for pre-specification in regulatory plans (e.g., \gls{pccp}) as they require predictable, periodic validation rhythms. & Scheduled Retraining, Periodic \gls{ft} \\

\makebox[0pt][l]{\hypertarget{dim:boundedCumul}{}}
\boundedCumul 
& Incremental parameter accumulation across a fixed, pre-enumerable set of new tasks or domains. Enables pre-specification of boundaries in regulatory change-control plans (e.g., \gls{pccp}). & \gls{til} \\

\makebox[0pt][l]{\hypertarget{dim:verPersistent}{}}
\verPersistent 
& Context-only modification at the model-parameter level, but maintained as a version-controlled, persistent system artifact whose update is a system-level change event. & \gls{rag}, \gls{ce}, Production system prompts, \gls{rag} corpora, persistent steering-vector stores, retrieval pipelines \\
\midrule
\multicolumn{3}{p{\linewidth}}{\cellcolor{gray!20} \textit{\scriptsize \textbf{II. Event-Driven Unplanned Modification:} Triggered by unpredictable events (e.g., sudden drift, emergent vulnerabilities, arbitrary developer decisions). 
Such modifications may require independent validation, re-clearance review, or conformity-preservation assessment depending on pathway.
}}\\
\midrule
\makebox[0pt][l]{\hypertarget{dim:AdhocPerm}{}}
\AdhocPerm 
& Unplanned, event-driven parameter updates triggered by sudden data drift, emergent vulnerabilities, or arbitrary developer decisions. 
May require event-driven validation and review under the applicable regulatory pathway.
& Emergency Patching, Knowledge Editing, Ad-hoc \gls{sft} \\
\midrule
\multicolumn{3}{p{\linewidth}}{\cellcolor{gray!20} \textit{\scriptsize \textbf{III. Continuous Surveillance Required:} Open-ended, streaming parameter accumulation fundamentally exceeds pre-specifiable \gls{pccp} bounds. 
Such approaches may require enhanced \gls{pms}, logging, and behavioral drift monitoring.
}}\\
\midrule
\makebox[0pt][l]{\hypertarget{dim:unboundedCumul}{}}
\unboundedCumul 
& Open-ended, continuous parameter updates over an infinite or unknown horizon of streaming data. 
Exceeds pre-specifiable limits, 
which may motivate enhanced post-market monitoring, logging, and drift controls.
& \gls{dil} \\ 
\midrule
\multicolumn{3}{p{\linewidth}}{\cellcolor{gray!20} \textit{\scriptsize \textbf{IV. Session-Scoped:} These interventions clear completely and do not persist beyond the inference session. While they may present a lower change-control burden, they may require software validation  
of reset fidelity to ensure base weights are successfully rolled back and no state carries across sessions.}}\\
\midrule
\makebox[0pt][l]{\hypertarget{dim:eph}{}}
\eph 
& Context-only modification supplied ad hoc per request or session; state clears post-inference. & User prompts, \gls{icl} demonstrations supplied at query time, Test-Time Compute Scaling. \\

\makebox[0pt][l]{\hypertarget{dim:trans}{}}
\trans 
& 
Temporary inference-session change. Parameter-modifying variants require explicit weight rollback post-inference; activation-modifying variants naturally clear at session end.
& \gls{tta}, Activation Steering \\
\midrule
\multicolumn{3}{p{\linewidth}}{\cellcolor{gray!20} \textit{\scriptsize \textbf{V. Inherited Burden:} Change-control and persistence documentation are entirely delegated to and subsumed by the downstream training event being fed. 
A separate persistence entry may not be required when the pipeline change is fully covered by the downstream training event; this should be assessed case by case.
}}\\
\midrule
\makebox[0pt][l]{\hypertarget{dim:pipeDep}{}}
\pipeDep 

& Persistence fully inherited from complementary downstream training method. & Data Augmentation, Active Learning, Curriculum Learning, \gls{semiSL} 
\\
\bottomrule
\end{tabularx}
}
\end{table}
}
This temporal classification can help \gls{pccp} authors declare whether an adaptation establishes an enduring structural baseline (\taxNo{4}{SchedAdhocPerm}), updates dynamically across streaming operations (\taxNo{4}{boundedUnboundedCumul}), or 
does not necessarily create a persistent parametric modification
(\taxNo{4}{ephVerTrans}).

The distinction between \tax{4}{eph} and \tax{4}{trans} persistence is structurally profound. 
\tax{4}{eph} denotes a temporary input context or inference-search state that clears after the request. \tax{4}{trans} denotes temporary internal model state during execution and includes both rolled-back parameter updates, as in \gls{tta}, and activation-space interventions, as in Activation Steering. The distinction is therefore temporal and operational rather than purely parametric.

Techniques operating on the data pipeline, e.g., Data Augmentation, Active Learning, Curriculum Learning, 
\gls{semiSL}, 
are classified as \tax{4}{pipeDep}. 
This indicates that their temporal persistence is delegated to, and inherited by, the downstream training method they complement. 
\dimens{D}{4} records the persistence of the realized intervention. Inference-only techniques can still depend on \tax{4}{verPersistent} system artifacts, so model-parameter persistence and system-level persistence should be documented separately.

\subsection{\dScope}
\label{sec:D5_Scope}

Dimension 5 delineates the structural boundaries and topological localization of the modification within the \gls{ml} architecture. Dimension 5 isolates \textit{where} and \textit{to what extent} the system graph is altered. 
For compliance officers and engineers executing risk impact assessments (such as under EU \gls{mdr} Annex II), this dimension determines the structural footprint of a modification, and can inform the scale of regression testing, safety re-validation, and software component separation.

The architectural boundaries of post-training interventions are partitioned into six 
structural meta-categories, as formalized in Table~\ref{tab:d5_scope}.
For techniques operating strictly via data-pipeline manipulation, the structural scope is classified as \tax{5}{pipelineDepScope}, explicitly signaling that the architectural footprint, and the corresponding component 
component-testing emphasis
is primarily inherited from the downstream training method that consumes the data.
Crucially, the introduction of the \tax{5}{extContxt} and \tax{5}{actSpace} categories explicitly isolates techniques (e.g., \gls{rag}, \gls{pe}) that have zero parametric footprint. 
From a compliance perspective, particularly under \gls{mdr} Annex II and \gls{fda} \gls{pccp} impact assessments, these classifications 
help distinguish systems that may primarily require behavioral validation from those that may also require structural regression testing.

{\scriptsize
\begin{table}[hbt!]
\centering
\renewcommand{\tabularxcolumn}[1]{p{#1}}
\caption{\dScope Categories}
\label{tab:d5_scope}
\renewcommand{\arraystretch}{1.3}
\adjustbox{width=.9\linewidth
}{
\begin{tabularx}{\textwidth}{p{.15\linewidth} p{.5\linewidth} X}
\toprule
\multicolumn{1}{c}{\textbf{Category}} &
\multicolumn{1}{c}{\textbf{Definition}} &
\multicolumn{1}{c}{\textbf{Representative Techniques}}
\\
\midrule
\multicolumn{3}{p{\linewidth}}{\cellcolor{gray!20} \textit{\scriptsize \textbf{I. Full-System Validation:}
Modifies the entire computational graph or merges models into lineage-opaque artifacts.  
May warrant full-system regression testing and re-validation, as targeted component rollback is impossible without full checkpoint reversion. 
May trigger or support the need for stringent provenance and traceability documentation.
}}\\
\midrule
\makebox[0pt][l]{\hypertarget{dim:whole}{}}
\whole 
& All model layers and components modified in full. & Training, Retraining, Full \gls{ft}, \gls{mtl}, Adversarial Training \\

\makebox[0pt][l]{\hypertarget{dim:fusedComp}{}}
\fusedComp 
& 
Derived artifact produced by 
parameter combination of multiple parents, whose behavior is not attributable to any single parent (lineage-opaque).
& Task Arithmetic, Model Merging, Machine Unlearning (vector negation) \\
\midrule
\multicolumn{3}{p{\linewidth}}{\cellcolor{gray!20} \textit{\scriptsize \textbf{II. Targeted Component Validation:}  
Imposes a partial structural footprint, allowing targeted regression testing of specific components. Rollback feasibility varies by implementation depth. Depending on the intervention, it may be evaluated as a change-control amendment to an existing device, 
or may require an independent conformity-assessment pathway
as a newly created standalone artifact.}}\\
\midrule
\makebox[0pt][l]{\hypertarget{dim:partl}{}}
\partl 
& 
Only a designated component or layer subset modified. 
For regulatory rollback documentation, partial modifications vary in reversibility; isolated head updates permit targeted rollback, whereas embedded layer updates require full-model checkpoint restoration.
& Partial \gls{ft}, \gls{peft}, Prompt Learning, Machine Unlearning, Knowledge Editing \\ 

\makebox[0pt][l]{\hypertarget{dim:surrogate}{}}
\surrogate 
& A completely new, independent model artifact trained to replicate a teacher model's outputs or to provide a learned scoring signal as an intermediate training artifact (e.g., reward model). 
A surrogate model may need to be assessed as a distinct artifact, depending on its role, intended use, and regulatory pathway.
& \gls{kd}, \gls{rlhf} \\

\makebox[0pt][l]{\hypertarget{dim:compressed}{}}
\compressed
& A structurally reduced, pruned, or lower-precision version of the original base model. 
May be assessed as a modified derivative artifact, depending on its role, intended use, and regulatory pathway.
& Model Compression \\
\midrule
\multicolumn{3}{p{\linewidth}}{\cellcolor{gray!20} \textit{\scriptsize \textbf{III. Modular Isolation:} 
Modifies the system via decoupled additions while the base architecture remains completely untouched. 
Modular isolation can provide a comparatively clear rollback documentation pattern (by simply swapping or removing the module).
}}\\
\midrule
\makebox[0pt][l]{\hypertarget{dim:modSwap}{}}
\modSwap 
& Decoupled module; base model unchanged and fully rollback-able. 
& \gls{lp}, \gls{moe}, \gls{til}, Swappable \gls{lora} adapters, Modality-Specific Adapters, Calibration \\
\midrule
\multicolumn{3}{p{\linewidth}}{\cellcolor{gray!20} \textit{\scriptsize \textbf{IV. No Parametric Footprint:}  
Modifies system behavior entirely through input manipulation, output selection, internal search, or transient activation modulation, leaving all internal parameters untouched. 
Validation may emphasize I/O correctness and behavioral boundaries while retaining integration and end-to-end behavioral checks where appropriate.
}}\\

\midrule
\makebox[0pt][l]{\hypertarget{dim:extContxt}{}}
\extContxt 
& 
No modification to any model parameter or internal state; behavioral conditioning occurs exclusively through external input-space manipulation, output selection, or inference search.
& \gls{pe}, \gls{rag}, \gls{icl}, \gls{ce}, Test-Time Compute Scaling \\

\makebox[0pt][l]{\hypertarget{dim:actSpace}{}}
\actSpace 
& Direct modulation of intermediate forward-pass representations (e.g., adding learned steering vectors to residual streams) with zero modification to the underlying parameter weights. 
& Activation Steering \\
\midrule
\multicolumn{3}{p{\linewidth}}{\cellcolor{gray!20} \textit{\scriptsize \textbf{V. Distributed Aggregation:}
Synthesizes global model state from decentralized local updates. Shifts the validation focus from single-system structural regression to cross-client behavioral consistency, differential privacy budget auditing, and aggregation integrity. 
May implicate cross-border data-transfer regulations where personal data or cross-jurisdictional processing are involved.
}}\\
\midrule
\makebox[0pt][l]{\hypertarget{dim:dIVdist}{}}
\dIVdist 
& Global model state from federated local-update aggregation. & \gls{fl} \\

\midrule
\multicolumn{3}{p{\linewidth}}{\cellcolor{gray!20} \textit{\scriptsize \textbf{VI. Inherited Burden:}
The structural footprint and primary component-validation burden are inherited from the downstream training event. The pipeline intervention should still be documented and assessed where it materially changes the data presented to that event.
}}\\
\midrule

\makebox[0pt][l]{\hypertarget{dim:pipelineDepScope}{}}
\pipelineDepScope 
& Structural scope is fully inherited from the complementary downstream training method. & Data Augmentation, Active Learning, Curriculum Learning, \gls{semiSL} \\

\bottomrule
\end{tabularx}
}
\end{table}
}
The interplay between decoupled structural scope (\tax{5}{modSwap}) and varying persistence levels 
(\taxNo{4}{SchedAdhocPerm} versus \tax{4}{eph}) 
has grown rapidly in practical importance with the rise of \glspl{fm}. Modern deployment patterns frequently layer a permanent, modular \gls{peft} adapter 
(\taxNo{4}{SchedAdhocPerm}, 
\tax{5}{modSwap}) onto a frozen base model, while simultaneously employing ephemeral techniques like \gls{pe} and \gls{rag} (\taxNo{4}{ephVerPersistent}, \tax{5}{extContxt}) to inject session-specific context
through a composite, layered architecture characteristic of the \gls{llm} and \gls{mllm} tiers.

\subsection{Dimension 6 (\dimens{D}{6}): Model Type}
\label{sec:D5ModelType}
Unlike Dimensions~\dimens{D}{1}--\dimens{D}{5}, which characterize inherent properties of an adaptation technique, Dimension~\dimens{D}{6} characterizes the \textit{target model} to which a technique is applied. It functions not as an independent characterizing axis but as a \textit{modulating dimension}: a filter that constrains which combinations of \dimens{D}{1}--\dimens{D}{5} are feasible for a given model. \gls{pe} is meaningless for a Random Forest; cross-modal alignment is irrelevant for a text-only \gls{llm}. This section defines the model type hierarchy and traces how the available adaptation landscape changes at each level.

\subsubsection{The Model Type Hierarchy}
\label{sec:ModelHierarchy}
The space of \gls{ml} models forms a tree rather than a linear scale. At the coarsest level, \gls{ml} divides into \textit{Traditional \gls{ml}} (shallow models such as support vector machines, random forests, logistic regression, and k-nearest neighbors) and \textit{\gls{dl}} (multi-layer neural networks that learn hierarchical representations, including convolutional, recurrent, and standard Transformer architectures). Within \gls{dl}, a distinct paradigm has emerged: \textit{\glspl{fm}}, defined by Bommasani et al.~\cite{bommasani_opportunities_2021} as large-scale models trained via \gls{ssl} on broad data and subsequently adapted to a wide range of downstream tasks. While the original definition emphasized breadth of adaptation, \glspl{fm} in 2024 and beyond are increasingly generative and multimodal.

At the \gls{fm} level, the hierarchy branches into two sibling categories: \textit{\glspl{llm}}, which process and generate text and code (e.g., GPT, LLaMA, Mistral), and \textit{\glspl{mllm}}, which process and generate across multiple modalities simultaneously, including text, images, audio, and video (e.g., GPT-4o, Gemini, LLaVA). 
Structurally, \glspl{llm} and \glspl{mllm} are siblings under the broader \glspl{fm} umbrella. While the \gls{llm} designation refers text-centric architectures, a model transitions into the \gls{mllm} category once it processes additional modalities, such as images or audio.

While this \frameTax classifies \glspl{fm} under a unified structural tier \tax{6}{fm}, it is operationally critical to recognize that this node encompasses two distinct architectural lineages: \glspl{repfm} and \glspl{genfm}. \glspl{repfm}, such as encoder-only architectures (e.g., BERT) and modern embedding models, are optimized for feature extraction, classification, and retrieval. Because they lack decoder blocks, they are structurally incompatible with generation-dependent adaptation mechanisms, such as \tax{1}{ctx}. Conversely, \glspl{genfm} (e.g., T5, BART, and early GPT models) incorporate generative decoders, enabling the use of \gls{pe}, \gls{icl}, and \gls{rag}. In this \frameTax, the unified \gls{fm} node represents the baseline pre-training scale, while the subsequent \gls{llm} and \gls{mllm} sibling categories act as the direct, specialized descendants of the \gls{genfm} lineage.
\subsubsection{Model Type as Modulating Dimension}
\label{sec:D5Modulating}
Model type does not simply add an orthogonal axis of classification; it \textit{constrains} the feasible space of adaptation. Three properties distinguish \dimens{D}{6} from \dimens{D}{1}--\dimens{D}{5} and give the adaptation landscape a \textit{relation structure}, encompassing inheritance, supersession, and emergent category creation, that cannot be captured by a single hierarchy.

First, \dimens{D}{6} is a pragmatic applicability and navigation hierarchy rather than a claim that all model classes are mutually exclusive. The hierarchy Traditional \gls{ml}~$\rightarrow$~\gls{dl}~$\rightarrow$~\gls{fm}~$\rightarrow$~\{\gls{llm}~$|$~\gls{mllm}\} represents common operational applicability: techniques available at a parent node are often, but not invariably, available at descendant nodes. 

Second, \dimens{D}{6} exhibits \textit{technique supersession}. 
Moving down the hierarchy not only \textit{adds} techniques; it can also \textit{replace} earlier methods with more general alternatives.
\gls{da} (classical feature alignment) is largely superseded at the \gls{llm} level by \gls{cpt} and domain-specific \gls{peft}. Meta-learning (bi-level optimization for rapid adaptation) is functionally replaced by \gls{icl}. Incremental learning is subsumed by continual learning. ``Superseded'' is not binary: a technique may remain technically available at a higher-level while being practically uncommon because a more general alternative covers the same goal more efficiently. 

Third, \dimens{D}{6} \textit{creates new \dimens{D}{1}--\dimens{D}{5} categories} at specific branch points. 
\tax{1}{ctx} as a \dimens{D}{1} mechanism exists only at the \taxNo{6}{lm} level. 
\tax{3}{multiD} 
is most characteristic at the \tax{6}{mllm} level, while related paired-modality regimes may also occur for multimodal representation \glspl{fm}.
\tax{4}{eph} as a \dimens{D}{4} category becomes a dominant deployment pattern only at the \taxNo{6}{lm} level. 
\dimens{D}{6} does not operate independently of \dimens{D}{1}--\dimens{D}{5}; 
it modulates which categories within each dimension are practically relevant.

\subsection{Multi-Dimensional Technique Profiles}
\label{sec:CenterpieceTable}
Table~\ref{tab:centerpiece} presents the complete six-dimensional profile of each adaptation technique discussed in this paper. Each row provides the technique's classification on all six dimensions simultaneously, enabling direct comparison across any axis. 
The \dimens{D}{6} column uses set notation to indicate applicable model types, making inheritance patterns visible: moving down the model hierarchy adds techniques without removing them, except where techniques are superseded by more general alternatives (see Section~\ref{sec:D5Modulating}).
The entries are canonical practical profiles rather than universal invariants. Actual deployments should be classified from the realized implementation; 

To maintain dimensional precision, inclusion in Table~\ref{tab:centerpiece} is strictly limited to \textit{distinct practical adaptation decision units}, defined as techniques, strategies, or operational patterns that 
have a materially distinct six-dimensional profile from existing rows
and that a practitioner would consider independently. 
Consequently, three categories of methods are 
excluded:
(1) Umbrella paradigms that subsume multiple rows, such as \gls{tl} (which encompasses \gls{ft}, \gls{da}, and \gls{fsl}; see 
\SuppTech);
(2) Pure goals realized through varying mechanisms, such as Personalization, which is instead treated as a \dimens{D}{2} category;
and (3) Low-level optimizations that constrain the training loop without constituting an adaptation choice in their own right.

{
\scriptsize
\centering
\setlength{\tabcolsep}{3pt}   

\begin{xltabular}{.9\textwidth}{
c
X
X
X
X
X
X
X
}
\caption{\textbf{Centerpiece Table:} Six-Dimensional Adaptation Technique Profiles.
} 
\label{tab:centerpiece}
\\
\toprule
\textbf{\#} & \textbf{Technique} & 
\makecell{\textbf{\dimens{D}{1}}\\\textbf{Mechanism}} & \makecell{\textbf{\dimens{D}{2}}\\\textbf{Goal}} & \makecell{\textbf{\dimens{D}{3}}\\\textbf{Data}} & \makecell{\textbf{\dimens{D}{4}}\\\textbf{Persistence}} & \makecell{\textbf{\dimens{D}{5}}\\\textbf{Scope}} & \makecell{\textbf{\dimens{D}{6}}\\\textbf{Model Type}} \\
\midrule
\endfirsthead

\multicolumn{8}{c}
{{\tablename\ \thetable\ -- continued from previous page}}\\
\toprule
\textbf{\#} & \textbf{Technique} & \makecell{\textbf{\dimens{D}{1}}\\\textbf{Mechanism}} & \makecell{\textbf{\dimens{D}{2}}\\\textbf{Goal}} & \makecell{\textbf{\dimens{D}{3}}\\\textbf{Data}} & \makecell{\textbf{\dimens{D}{4}}\\\textbf{Persistence}} & \makecell{\textbf{\dimens{D}{5}}\\\textbf{Scope}} & \makecell{\textbf{\dimens{D}{6}}\\\textbf{Model Type}} \\
\midrule
\endhead

\midrule
\multicolumn{8}{r}{{Continued on next page}}\\
\endfoot

\bottomrule
\endlastfoot


\tech{training} & Training & \fullAbbr & \buildAbbr & \largeDAbbr & 
\SchedAdhocPermAbbr
& \wholeAbbr & \all \\

\midrule
\multicolumn{8}{c}{\cellcolor{gray!20} \textit{\scriptsize \textit{I. Knowledge Transfer and Task Specialization}}}\\
\midrule
\tech{fullFT} & \gls{ft} (full) & \fullAbbr & \taskAbbr & 
\largeDAbbr
&
\SchedAdhocPermAbbr
& \wholeAbbr & \dflm \\
\tech{partFT} & \gls{ft} (partial) & \selAbbr & \taskAbbr & \smallDAbbr & 
\SchedAdhocPermAbbr
& \partlAbbr & \dflm \\
\tech{peft} & \gls{peft} (\gls{lora}, adapters) & \selAbbr & 
\taskCompEffAbbr &
\smallDAbbr & 
\SchedAdhocPermAbbr
& \partlModAbbr & 
\dflm
\\
\tech{reft} & 
\gls{reft} &
\paramUpdAbbr & \taskAbbr & \smallDAbbr & \SchedAdhocPermAbbr & \modSwapAbbr & \lm
\\
\tech{da} & \gls{da}$^\S$ & \selAbbr & 
\domAdaptAbbr &
\unlabSmallDAbbr &
\SchedAdhocPermAbbr
& \partlAbbr & \df \\
\tech{fsl} & \gls{fsl}$^\P$ & 
\selAbbr
& 
\taskAbbr
&
\smallFewAbbr
& 
\SchedAdhocPermAbbr
& 
\partlAbbr
&
\df
\\
\midrule

\multicolumn{8}{c}{\cellcolor{gray!20} \textit{\scriptsize \textit{II. Temporal Adaptation and Maintenance}}}\\
\midrule

\tech{retraining} & Retraining & \fullAbbr & \DriftRemedAbbr & 
\seqIncDAbbr
& 
\SchedAdhocPermAbbr
& \wholeAbbr & \all \\

\tech{cl} & \gls{cl} & 
\selAbbr &
\ContAdaptAbbr & 
\seqIncDAbbr
& 
\boundedUnboundedCumulAbbr
& \partlAbbr
& \dflm \\

\tech{til} & \gls{til} & 
\selAbbr &
\ContAdaptAbbr
&  
\seqIncDAbbr
& \boundedCumulAbbr & \partlModAbbr & \df \\

\tech{dil} & \gls{dil} & 
\selFullAbbr &
\ContAdaptDriftRemedAbbr
&  
\seqIncDAbbr
& \unboundedCumulAbbr & \partWholeAbbr & \df \\

\midrule
\multicolumn{8}{c}{\cellcolor{gray!20} \textit{\scriptsize \textit{III. 
Alignment, Reasoning, and Trustworthiness
}}}\\

\midrule

\tech{sft} & \gls{sft} & \selFullAbbr & 
\taskAlgnSafeAbbr
& \instrRespAbbr & \SchedAdhocPermAbbr & \partWholeAbbr & \lm \\

\tech{rlhf} & \gls{rlhf} (\gls{llm} Alignment)$^*$ 
& \selAbbr & 
\algnSafetyAbbr
&
\prefAbbr
& 
\SchedAdhocPermAbbr
& 
\partWholeSurrogAbbr
& \lm \\
\tech{dpo} & \gls{dpo} & \selAbbr & 
\algnSafetyAbbr
&
\prefAbbr & 
\SchedAdhocPermAbbr
& 
\partWholeAbbr
& \lm \\
\tech{rlaif} & \gls{rlaif}/\gls{cai} 
& \selAbbr & 
\algnSafetyAbbr
& 
\synthRuleAbbr
& 
\SchedAdhocPermAbbr
& 
\partWholeAbbr
& \lm \\
\tech{rlvr} & \gls{rlvr}/\gls{grpo}$^\S$ & 
\selAbbr &
\reasAbbr &
\verifyAbbr & 
\SchedAdhocPermAbbr
& 
\partWholeAbbr
& \lm \\

\tech{lp} & \gls{lp}$^\S$ & \selAbbr & \explainAbbr & \smallDAbbr & 
\SchedAdhocPermAbbr
& \modSwapAbbr & \dflm \\

\tech{leace} & \gls{leace} & \WeightSpaceCompAbbr & \fairnessAbbr & \smallDAbbr & \AdhocPermAbbr & \partlAbbr & \dflm \\

\tech{AdvrsTrn} & Adversarial Training & 
\fullAbbr &
\robAbbr &
\largeDAbbr &
\SchedAdhocPermAbbr
& \wholeAbbr & \dfl \\

\midrule
\multicolumn{8}{c}{\cellcolor{gray!20} \textit{\scriptsize \textit{IV. Training Strategies}}}\\
\midrule

\tech{metaLrn} & Meta-Learning & 
\selAbbr &
\taskAbbr
& \taskDistAbbr & 
\SchedAdhocPermAbbr
& \partlAbbr & \df \\
\tech{mtl} & \gls{mtl} & 
\fullAbbr &
\taskAbbr &
\taskDistAbbr
&
\SchedAdhocPermAbbr
& \wholeAbbr & \dflm \\

\tech{selfplay} & Self-Play$^\ddagger$ & 
\selAbbr
&
\taskAlgnReasAbbr
&
\synthEnvAbbr
& 
\SchedAdhocPermAbbr
& \wholeAbbr & \dL \\

\tech{curriLrn} & Curriculum Learning & 
\manipAbbr &
\taskAbbr
&
\pipeInheritAbbr
&
\pipeDepAbbr & 
\pipelineDepScopeAbbr
& \all \\

\tech{activeLrn} & Active Learning & 
\manipAbbr &
\taskDomAdaptAbbr
& 
\pipeInheritAbbr
&
\pipeDepAbbr
& 
\pipelineDepScopeAbbr
& \all \\

\midrule
\multicolumn{8}{c}{\cellcolor{gray!20} \textit{\scriptsize \textit{V. Data-Centric and Privacy-Preserving Methods}}}\\
\midrule

\tech{dataAug} & Data Augmentation & 
\manipAbbr &
\taskRobAbbr
& 
\pipeInheritAbbr
&
\pipeDepAbbr
& 
\pipelineDepScopeAbbr
& \all \\

\tech{semiSl} & \gls{semiSL} & 
\manipAbbr &
\taskAbbr
& 
\unlabSmallDAbbr & 
\pipeDepAbbr & 
\pipelineDepScopeAbbr
& \all \\

\tech{ssl} & \gls{ssl} / 
\gls{cpt}$^\dagger$ & 
\selFullAbbr
&
\domAdaptCapExtAbbr
&
\unlabDAbbr &
\unboundedCumulSchedPermAbbr
& \partWholeAbbr & \dflm \\

\tech{dpft} & \gls{dpft} & \paramUpdAbbr & \privAbbr & \SmallLargeDAbbr & 
\SchedAdhocPermAbbr
& \partWholeAbbr & \dflm \\

\tech{fl} & \gls{fl} & 
\selFullAbbr
& \privAbbr & 
\decentAbbr &
\ScheduledPermAbbr
& \dIVdistAbbr & \all \\

\midrule
\multicolumn{8}{c}{\cellcolor{gray!20} \textit{\scriptsize \textit{VI. Efficiency and Composition}}}\\
\midrule

\tech{kd} & 
\gls{kd} & 
\CrossModelTransfAbbr &
\compEffAbbr &
\synthTeacherAbbr
&
\SchedAdhocPermAbbr
& \surrogateAbbr & \dflm \\
\tech{compress} & Model Compression & 
\StructReductAbbr &
\compEffAbbr &
\unlabDWghtAbbr
&
\SchedAdhocPermAbbr
& 
\compressedAbbr
& \dflm \\

\tech{taskArith} & Task Arith. \& Model Merging & 
\WeightSpaceCompAbbr &
\capExtCompEffAbbr
&
\wghtAbbr & 
\SchedAdhocPermAbbr
& \fusedCompAbbr & \flm \\

\tech{lce} & \longContextAbbr & \paramArchAbbr & \capExtAbbr & \unlabSmallDAbbr & \SchedAdhocPermAbbr & \partlAbbr & \flm \\

\tech{moe} & \gls{moe}$^\S$ & \paramArchAbbr & 
\capExtCompEffAbbr
&
\unlabSmallDAbbr &
\SchedAdhocPermAbbr
& 
\partlModAbbr
&
\flm \\

\midrule
\multicolumn{8}{c}{\cellcolor{gray!20} \textit{\scriptsize \textit{VII. Inference-Time Adaptation}}}\\
\midrule

\tech{pe} & \gls{pe}$^\S$ & \ctxAbbr & 
\augKBehavAbbr
&
\zeroShAbbr & 
\ephVerPersistentAbbr
& \extContxtAbbr & \lm \\

\tech{pl} & Prompt Learning & 
\selAbbr & 
\taskAbbr & \smallDAbbr & 
\SchedAdhocPermAbbr
& \partlAbbr & \flm \\

\tech{apo} & \gls{apo} & \ctxAbbr & \taskBehavAbbr & \zeroFewAbbr & \verPersistentAbbr & \extContxtAbbr & \lm \\

\tech{rag} & \gls{rag} & \ctxAbbr & 
\augKAbbr &
\extAbbr & 
\verPersistentAbbr
& \extContxtAbbr & \lm \\
\tech{icl} & \gls{icl}$^\S$ & \ctxAbbr & 
\taskAugKAbbr
&
\fewAbbr & \ephAbbr & \extContxtAbbr & \lm \\
\tech{ce} & \gls{ce} & \ctxAbbr & 
\augKPersonAbbr
&
\extUserAbbr
&
\verPersistentAbbr
& \extContxtAbbr & \lm \\
\tech{tta} & \gls{tta} & 
\selAbbr
& \DriftRemedAbbr & 
\tstOnlyAbbr &
\transAbbr & 
\partWholeAbbr
& \df \\

\tech{ttCompute} & Test-Time Compute Scaling & \infTimeSearchAbbr & \reasAbbr & \zeroShAbbr & \ephAbbr & \extContxtAbbr & \lm \\

\midrule
\multicolumn{8}{c}{\cellcolor{gray!20} \textit{\scriptsize \textit{VIII. Calibration, Personalization, and Multimodal Adaptation}}}\\
\midrule

\tech{calib} & Calibration & 
\selAbbr
& 
\reliabilityAbbr
&
\unlabSmallDAbbr
&
\SchedAdhocPermAbbr
& \modSwapAbbr & \all \\

\tech{crossModel} & Cross-Modal Alignment & 
\selAbbr &
\capExtAbbr &
\multiDAbbr & 
\SchedAdhocPermAbbr
& \partlAbbr & 
\FM
\\
\tech{modelAdapter} & Modality-Spec. Adapters & 
\selAbbr &
\capExtCompEffAbbr
&
\multiDAbbr &
\SchedAdhocPermAbbr
& \modSwapAbbr & \mllm \\

\tech{multimodInstr} & Multimodal Instr. Tuning & \selFullAbbr & 
\capExtAlgnAbbr
&
\multiDAbbr &
\SchedAdhocPermAbbr
& \partWholeAbbr & \mllm \\

\midrule
\multicolumn{8}{c}{\cellcolor{gray!20} \textit{\scriptsize \textit{IX. Knowledge Modification and Activation-Based Adaptation}}}\\
\midrule

\tech{mUnLrn} & Machine Unlearning & 
\selWeightSpaceAbbr &
\remKPrivAbbr &
\forgetAbbr & 
\AdhocPermAbbr
& 
\PartlFusedCompAbbr
& \dflm \\
\tech{ke} & Knowledge Editing$^\S$ & 
\selWeightSpaceAbbr
&
\augKAbbr &
\smallDAbbr &
\AdhocPermAbbr
& \partlAbbr & \lm \\

\tech{actSteer} & Activation Steering & 
\transActAbbr &
\algnSafetyAbbr 
&
\unlabSmallDAbbr
&
\transAbbr & 
\actSpaceAbbr
& \lm \\

\end{xltabular}
\vspace{-1.5em}
 \begin{flushleft}  
 \par\smallskip
{\footnotesize
    \textbf{Note:} 
    \ifacm
\else
Abbreviations are listed in the manuscript before the bibliography.
    \fi
     Highly flexible techniques such as \gls{ce} (e.g., prompting for step-by-step rationales) and \gls{rlhf} (e.g., penalizing unexplainable refusals) can dynamically adopt \tax{2}{explain} as a primary goal when engineered specifically to satisfy regulatory transparency mandates.
     \par
      $^\dagger$~\gls{ssl} is classified in its adaptation role (\gls{cpt} on domain-specific unlabeled data). As a foundational pre-training paradigm, it falls outside the adaptation taxonomy.
     \par
     $^\ddagger$~Self-play skips the \gls{fm} tier because standard \gls{fm} pre-training and \gls{ft} pipelines do not employ self-play; it re-emerges at the \gls{llm} level through \gls{rlhf}-adjacent techniques (e.g., self-play \gls{ft}, \gls{spin}~\cite{chen_spin_2024}) that exploit autoregressive generation as an opponent model.
     \par
     $^\S$~Classification tensions involving these techniques are discussed in 
      \SuppTech.
     \par
     $^\P$~For in-context \gls{fsl} (Row~\ref{tech:fsl}) at the \gls{llm} level, see \gls{icl} (Row~\ref{tech:icl}).
     \par
     $^*$~While preference-based \gls{rl} predates foundation models,
     this row explicitly scopes its modern instantiation for generative alignment at the \taxNo{6}{lm} tiers. 
     \\
}
\end{flushleft} 
}
\normalsize

Table~\ref{tab:centerpiece} makes several relationships visible. \gls{llm} and \gls{mllm} adaptation introduces \tax{1}{ctx}, \tax{1}{infTimeSearch}, \tax{1}{transAct}, \tax{2}{algn}, \tax{2}{reas}, \tax{2}{rem}, and \tax{4}{eph} persistence.
\gls{mllm}-specific rows are distinguished by \tax{3}{multiD} data and \gls{mllm} applicability.
Machine Unlearning and Knowledge Editing represent different goals, \tax{2}{rem} and \tax{2}{augK}, but both may connect to Task Arithmetic through \tax{1}{WeightSpaceComp}.
These relationships illustrate the value of comparing complete dimensional profiles rather than single family labels.

\subsection{Technique Definitions}
\label{sec:TechniqueDefinitions} 
Table~\ref{tab:centerpiece} organizes the techniques into \nbFamilies navigational families:
1) Knowledge Transfer and Task Specialization (Rows~\ref{tech:fullFT}--\ref{tech:fsl});
2) Temporal Adaptation and Maintenance (Rows~\ref{tech:retraining}--\ref{tech:dil});
3) Alignment, Reasoning, and Trustworthiness (Rows~\ref{tech:sft}--\ref{tech:AdvrsTrn});
4) Training Strategies (Rows~\ref{tech:metaLrn}--~\ref{tech:activeLrn});
5) Data-Centric and Privacy-Preserving Methods (Rows~\ref{tech:dataAug}--\ref{tech:fl});
6) Efficiency and Composition (Rows~\ref{tech:kd}--\ref{tech:moe});
7) Inference-Time Adaptation (Rows~\ref{tech:pe}--\ref{tech:ttCompute});
8) Calibration, Personalization, and Multimodal Adaptation (Rows~\ref{tech:calib}--\ref{tech:multimodInstr}); 
and 
9) Knowledge Modification and Activation-Based Adaptation (Rows~\ref{tech:mUnLrn}--\ref{tech:actSteer}).
Detailed \dimens{D}{1}--\dimens{D}{5} category definitions are provided in \SuppCategory.
Detailed technique definitions and boundary cases are provided in {\SuppTech}.

\subsection{Exploratory Structural Consistency Analysis}
\label{sec:topological}

To explore how the six-dimensional characterizations in Table~\ref{tab:centerpiece} are represented geometrically
we computed a Gower-style dissimilarity matrix over \dimens{D}{1}--\dimens{D}{6} and projected the results using UMAP (full implementation details and geometric mappings are provided in 
\SuppQuantitative).
This analysis serves as a diagnostic for whether the proposed functional families occupy coherent regions of the categorical profile space and where boundary cases arise. As illustrated in 
\SuppQuantitative,
the projection reveals substantial overlap among the predefined functional families, quantified by a near-zero mean silhouette score. This is expected: 
the \nbFamilies functional families introduced in Section~\ref{sec:TechniqueDefinitions} (\SuppTech) are strictly navigational groupings designed to aid practitioner discovery, not validated statistical clusters.
This result supports the need for multi-dimensional characterization: techniques often share one anchor dimension while differing on others. For example, inference-time methods are anchored by ephemeral persistence and context injection, whereas \gls{mllm}-specific methods are anchored by paired multimodal data and the \gls{mllm} model type.
 This exploratory representation analysis does not independently validate the \frameTax.
\section{Terminology Navigation for Describing Adaptation Pathways}
\label{sec:DecisionFramework}
The \frameTax is intended to name and compare adaptation interventions rather than prescribe a preferred method. The workflow below expresses choices in a common vocabulary for research, technical documentation, and governance.
\subsection{Entry Points}
Four entry points are commonly used. \dimens{D}{6} filters methods by target model, \dimens{D}{2} identifies the adaptation purpose, \dimens{D}{3} constrains the choice by available data, and \dimens{D}{1} with \dimens{D}{4} identifies methods compatible with access, compute, and persistence constraints.
For example, \tax{2}{DriftRemed} points to retraining or \gls{tta}, \tax{2}{reas} points to \gls{rlvr}, \tax{2}{algn} points to \gls{rlhf} or \gls{dpo}, \tax{2}{augK} points to \gls{pe}, \gls{icl}, \gls{rag}, \gls{ce}, or Knowledge Editing, and \tax{2}{rem} points primarily to Machine Unlearning.
Data and access constraints then narrow the set further. \tax{3}{smallD} data may support \gls{peft},  
\tax{3}{few} support \gls{icl}, and API-only access generally limits adaptation to \tax{1}{ctx} or \tax{1}{infTimeSearch}. Activation steering or calibration may require intermediate access, while parameter updates require weight access.
\subsection{The Model-Type-First Decision Path}
\label{sec:DecisionPath}
A robust documentation sequence may begin with \dModel, followed by \dGoal and \dData, and concludes with \dMechanism, \dPersistence, and \dScope.
For an \gls{llm}, \dimens{D}{6} first defines the applicable rows in Table~\ref{tab:centerpiece}.
The resulting set is then filtered by the primary adaptation goal and data regime.
An \gls{llm} intervention for \tax{2}{augK}, e.g., may be classified as \gls{pe} with \tax{3}{zeroSh} instructions, \gls{icl} with \tax{3}{few}, or \gls{rag} and \gls{ce} with an \tax{3}{ext}.
\dimens{D}{5} finally records whether the intervention is \tax{5}{whole}, \tax{5}{partl}, \tax{5}{modSwap}, or confined to the \tax{5}{extContxt}, which supports rollback and change-control documentation.
\subsection{Complementary Combinations}
\label{sec:Combinations}
Many techniques are complementary rather than mutually exclusive. The ordered composition operator ($\oplus$) denotes a deployment stack in which sequence may influence the resulting behavior and documentation profile.
 
A representative \gls{llm} stack is \gls{rlvr} (\tax{2}{reas}) $\oplus$ \gls{rlhf} (\tax{2}{algn}) $\oplus$ \gls{peft} (\tax{2}{task}) $\oplus$ \gls{rag} (\tax{2}{augK}) $\oplus$ \gls{pe} (\tax{2}{behavControl}) through different mechanisms and persistence profiles.  
\gls{fl} (\tax{2}{priv}) $\oplus$ \gls{peft} (\tax{2}{task}) provides a second example, in which lightweight adapters are trained across distributed data sources.
Privacy claims still depend on safeguards such as secure aggregation, differential privacy, access controls, and an explicit threat model.
Each layer should be named and documented separately even when the layers are deployed as one system.
\subsection{Common Sources of Terminological and Regulatory Ambiguity}
\label{sec:AntiPatterns}
Three ambiguities are particularly important for technical and regulatory documentation.
First, \gls{ft} does not distinguish full \gls{ft} (\tax{5}{whole}), partial \gls{ft} (\tax{5}{partl}), and modular \gls{peft} (\taxNo{5}{partlMod}).
These interventions differ in scope and should be named explicitly.
Second, using \gls{rag} is incomplete unless the corpus, retrieval method, and update cadence are documented.
Third, classical labels such as \gls{da} or meta-learning may obscure the actual \gls{llm} mechanism, e.g., \gls{cpt}, domain-specific \gls{peft}, or \gls{icl}. \dimens{D}{6} helps prevent these terminology mismatches.

\section{Governance and Compliance Context}
\label{sec:Governance}
This section does not provide legal advice and does not determine whether a specific model intervention is a substantial modification, a significant change, or a reportable device modification. It shows how the taxonomy can supply the technical vocabulary needed to describe such interventions before regulatory assessment. Determinations of regulatory consequence remain the responsibility of manufacturers, regulators, and notified or auditing bodies acting under the applicable legal framework.

As established in Section~\ref{sec:D4_Persistence}, the six-dimensional \frameTax supports the description of changes that affect model consistency, without specifying which changes are acceptable in any particular regulatory context.
\subsection{Mapping \FrameTax Dimensions to Regulatory Regimes}
\label{sec:RegMapping}
Table~\ref{tab:dimensions} summarizes the common documentation role of \dimens{D}{1}--\dimens{D}{6}. The discussion below focuses on how these dimensions are used within the \gls{nist} \gls{ai} \gls{rmf}, the EU AI Act, the EU \gls{mdr} and \gls{ivdr}, and the \gls{fda} \gls{pccp}. Detailed lifecycle mappings are provided in \SuppGov.

\textbf{\gls{nist} \gls{ai} \gls{rmf}.} 
The seven-stage lifecycle of \gls{ai} \gls{rmf} 1.0 (Plan and Design, Collect and Process Data, Build and Use Model, Verify and Validate, Deploy and Use, Operate and Monitor, Use or Impacted By) and the four core functions (Govern, Map, Measure, Manage) 
identify stage-specific risk-management activities relevant to adaptation~\cite{noauthor_nist_2023};
complementary management standards include ISO/IEC~42001 and ISO/IEC~23894. 
Furthermore, risk management mapping for \glspl{llm} should now align with the ``Generative AI Profile'' (NIST-AI-600-1)~\cite{autio_artificial_2024} and the April 2026 concept note for the ``Trustworthy AI Profile'' for critical infrastructure~\cite{stanley_concept_2026}, which address post-training risks like prompt injection, data poisoning, and model extraction.
The ``Operate and Monitor'' stage invokes \tax{2}{DriftRemed} and \tax{2}{ContAdapt} goals, which can be mapped to the separate temporal constraints of \dimens{D}{4} (Persistence) and structural limits of \dimens{D}{5} (Scope) \cite{noauthor_nist_2023}. This isolation maps cleanly to the EU \gls{mdr} Annex II significant-change protocol, which 
may require distinct change-control review, 
depending on whether a modification is structurally localized (\tax{5}{partl} or \tax{5}{modSwap}) and either temporally enduring (\taxNo{4}{SchedAdhocPerm}) or incrementally accumulated (\taxNo{4}{boundedUnboundedCumul}) \cite{european_parliament_and_of_the_council_regulation_2017}.
``Verify and Validate'' invokes \tax{2}{algn}, \tax{2}{reas}, and \tax{2}{rob}, including adversarial-threat evaluation across the lifecycle~\cite{vassilev_adversarial_2025}; 
``Collect and Process Data'' invokes \dimens{D}{3}, with technique-specific obligations for 
\tax{3}{pref}
(annotator governance), 
\tax{3}{verify}
(oracle provenance), and 
\tax{3}{decent}
(data sovereignty).

\textbf{EU AI Act.} The Act establishes risk-based requirements and post-market obligations that may be informed by the adaptation goal, persistence, and structural scope~\cite{noauthor_regulation_2024}. The \gls{gpai} compute threshold can identify some large \gls{ft} events, but it may not capture low-compute interventions that materially alter behavior~\cite{hacker_regulation_2025}. For example, full \gls{ft} may consume substantial \glspl{flop} and change the \tax{5}{whole} scope, whereas targeted Knowledge Editing may consume little compute while directly changing a high-stakes factual output. A domain-specific \gls{lora} adapter can also have significant deployment impact despite a small computational footprint.
The \frameTax separates compute intensity from purpose, data, persistence, and scope. It also supports documentation of \tax{1}{ctx}, Trust-related interventions (\tax{2}{fairness}), and Fused Composition artifacts (\tax{5}{fusedComp}). 
\gls{leace}, e.g., may provide a specified fairness intervention outside gradient-based compute thresholds, while \gls{rag} and production prompts may require documentation when they materially shape a high-risk system. 
\gls{lp} can similarly support explainability evidence when internal states are exposed without modifying the validated base model.

Furthermore, structural modifications mapped to \tax{5}{fusedComp} create lineage-opaque derived artifacts and may heighten the need for provenance and technical documentation to trace parent-model contributions under Annex IV of the Act.
Crucially, this applies equally to models merged via Task Arithmetic (Row~\ref{tech:taskArith}) and to models subjected to task-vector-negation Machine Unlearning (Row~\ref{tech:mUnLrn}), 
helping avoid inconsistent documentation of mathematically similar operations.

\textbf{EU \gls{mdr}/\gls{ivdr}.} Under the \gls{mdr}, AI/ML-driven \gls{mdsw} is subject to lifecycle obligations beyond initial conformity assessment. Article~83 requires a \gls{pms} system proportionate to the risk class~\cite{european_parliament_and_of_the_council_regulation_2017}, Article~84 requires a \gls{pms} plan as part of the technical documentation, and the \gls{ivdr} establishes parallel obligations~\cite{european_parliament_and_of_the_council_regulation_2017_746}. The \frameTax supports compliance documentation across these obligations: \dimens{D}{1}, \dimens{D}{4}, and \dimens{D}{5} support the technical-documentation requirement (a statement that a model ``was updated'' is insufficient when \gls{mdr} Annex II/III-style 
documentation may call for more precise description, such as ``the model received a
\tax{1}{paramUpd} producing a 
\taxNo{4}{SchedAdhocPerm}, \taxNo{5}{partlMod}
change via a specified \gls{peft} method''); 
\dimens{D}{2} and \dimens{D}{3} support benefit-risk and clinical-evaluation documentation, distinguishing a change driven by \tax{2}{DriftRemed} from one driven by \tax{2}{task} or \tax{2}{capExt} even when \dimens{d}{1} and \dimens{d}{4} are identical; 
and \dimens{D}{4} with \dimens{D}{5} 
can help assess whether a planned modification may fit
within an approved change-control framework, as a substantial modification, or through \gls{pms}-driven updates. The \gls{mdr}'s vigilance reporting obligations (Articles~87--92) and the \gls{ivdr}'s corresponding obligations (Articles~82--87) are not adaptation events themselves but may trigger them, with the appropriate response depending on the implicated \dimens{D}{2} goal.

\textbf{\gls{fda} \gls{pccp}.} The \gls{fda}'s \gls{pccp} framework~\cite{health_marketing_2025} allows manufacturers to pre-specify anticipated model modifications and the methodology for implementing them, enabling certain adaptations without new submissions. The broader \gls{gmlp} principles, harmonized internationally through the \gls{imdrf}, emphasize lifecycle data management and performance monitoring that connect to \dimens{D}{3} and \dimens{D}{4}. Table~\ref{tab:frameworkPCCP} maps \gls{pccp} components to \frameTax dimensions.

{\footnotesize
\begin{table}[htbp]
\centering
\caption{\gls{fda} \gls{pccp} component $\leftrightarrow$ taxonomy dimension}
\label{tab:frameworkPCCP}
\adjustbox{width=.9\linewidth
, max totalheight=0.95\textheight
}{
\begin{tabularx}{\linewidth}{l X}
\toprule
\textbf{\gls{pccp} Component} & \textbf{Relevant Taxonomy Dimensions} \\
\midrule
Description of modification & \dimens{D}{1} (what changes), \dimens{D}{4} (how lasting), \dimens{D}{5} (structural scope), \dimens{D}{6} (on what model) \\
Modification protocol & \dimens{D}{3} (what evidence is required) plus separately stated validation/acceptance criteria \\
Impact assessment & \dimens{D}{2} (intent), \dimens{D}{4} (reach of change), and intended-purpose / safety-effectiveness analysis (outside the framework) \\
\bottomrule
\end{tabularx}
}
\end{table}
}

A \gls{pccp} might specify: ``The model may be adapted via \tax{1}{sel} (\gls{peft}), for \taxNo{2}{taskCompEff}, using \tax{3}{smallD} ($n \geq 200$), with 
\taxNo{4}{SchedAdhocPerm}, \taxNo{5}{partlMod},
on a \tax{6}{fm} model type.'' This specificity is impossible with single-axis descriptions such as ``the model may be fine-tuned,'' as illustrated by the comparative change-log scenario in Table~\ref{tab:before_after_changelog}.

{\scriptsize
\begin{table}[h]
\centering
\caption{Before/After: Translating a vague change-log entry into a {\nbDim}D regulatory specification.}
\label{tab:before_after_changelog}
\adjustbox{width=.9\linewidth
, max totalheight=0.95\textheight
}{
\begin{tabularx}{\textwidth}{p{.15\linewidth} X}
\toprule
\multicolumn{2}{c}{\textbf{Example: Medical AI System Update (Change-Log Entry)}} \\
\midrule
\textbf{Before (Standard Prose)} & \textit{``Update v2.1: The model was fine-tuned on recent hospital data to improve performance and incorporate new clinical guidelines.''} \\
\addlinespace
Regulatory Ambiguity & The term ``fine-tuned'' fails to specify if the base weights were altered, \dScope, if the change is a transient patch or permanent, \dPersistence, or if the guidelines were learned parametrically or retrieved, \dMechanism. This prevents an auditor from determining if a full system re-validation is required. \\
\midrule
\textbf{After ({\nbDim}D \FrameTax)} & \textbf{Layer 1 (Parameter Update):} \gls{peft} applied to base model. \newline 
Profile: 
\profile{\tax{1}{paramUpd}}{\tax{2}{DriftRemed}}{\tax{3}{smallD}}{\tax{4}{AdhocPerm}}{\tax{5}{modSwap}}{\tax{6}{llm}}.
\newline\newline
\textbf{Layer 2 (Inference Update):} RAG corpus updated with new guidelines. \newline
Profile: 
\profile{\tax{1}{ctx}}{\tax{2}{augK}}{\tax{3}{ext}}{\tax{4}{verPersistent}}{\tax{5}{extContxt}}{\tax{6}{llm}}
\\
\addlinespace
Compliance Payoff & 
The {\nbDim}D specification records that the base weights are intended to remain frozen and identifies the adapter and retrieval artifacts as the primary modified components (\tax{5}{modSwap}, \tax{5}{extContxt}).
\\
\bottomrule
\end{tabularx}
}
\end{table}
}

The \frameTax can support targeted review and rollback planning, but it does not itself prove implementation fidelity or replace end-to-end validation.
\dimens{D}{4} is particularly relevant to \gls{pccp} planning. \tax{4}{trans}, \tax{4}{verPersistent}, and \taxNo{4}{boundedUnboundedCumul} may require different validation, logging, and monitoring arrangements, although the applicable procedure remains case specific.

The \frameTax supplies a specification vocabulary rather than an operational governance system. \gls{aegis}-type infrastructures~\cite{aegis_preprint_2025} may use these profiles as inputs to monitoring and release workflows, but their technical integration remains prospective.

\subsection{Why No Single Dimension Determines Regulatory Consequence: Intended Use and Inference-Time Interventions}
\label{sec:frameworkFeatures}

Inference-time interventions, e.g., \gls{rag}, \gls{pe}, \gls{ce}, and Activation Steering, may alter deployed behavior without changing stored parameters. Two principles are important.
\textbf{First}, a maintained corpus, prompt, index, retriever, or steering artifact can be \tax{4}{verPersistent} even when the per-request context is \tax{4}{eph}.
\textbf{Second}, regulatory consequence depends on intended use, risk class, and the applicable pathway in addition to the \dimens{D}{1}--\dimens{D}{6} profile. The same \gls{rag} architecture can therefore have different consequences in customer support and clinical decision support.

Existing change-control frameworks do not always state clearly how updates to inference-time artifacts should be treated.
The \frameTax provides defined intervention types that can be used in future guidance, while it does not resolve the legal classification.

Practitioners should therefore document recurring updates to persistent inference-time artifacts even when they are not treated as parameter modifications.
A quarterly clinical-guideline corpus update, e.g., is a recurring system-level adaptation.

\subsection{Operational Scenario: Medical \gls{fm} with Layered Adaptation}
\label{sec:GovernanceVignette}
Consider a hospital deploying a foundation model for radiology report generation across three adaptation layers:

\textbf{Layer 1: Domain-specific \gls{peft}.} A \gls{lora} adapter trained on 5,000 institutional reports specializes the base model. Profile: 
\profile{\tax{1}{sel}}{\tax{2}{task}}{\tax{3}{smallD}}{\tax{4}{ScheduledPerm}}{\taxNo{5}{partlMod}}{\tax{6}{llm}}.
Under the EU AI Act, this layer is 
likely to be a strong 
candidate for substantial-modification analysis; under \gls{fda} \gls{pccp}, it 
may be appropriate to pre-specify
\dimens{D}{3}, \dimens{D}{4} and \dimens{D}{5} constraints.

\textbf{Layer 2: \gls{rag} over clinical guidelines.} The system retrieves sections from clinical protocols at inference. 
Profile: 
\profile{\tax{1}{ctx}}{\tax{2}{augK}}{\tax{3}{ext}}{\tax{4}{verPersistent}}{\tax{5}{extContxt}}{\tax{6}{llm}}.
Although the model's parameters remain unchanged, its \dPersistence profile is characterized as \tax{4}{verPersistent}; therefore, transparency requirements may still apply where retrieval sources materially shape clinical outputs;
corpus updates represent a distinct change pathway from parameter updates.
  
\textbf{Layer 3: \gls{pe} for output formatting.} System prompts enforce structured reporting templates. 
Profile: 
\profile{\tax{1}{ctx}}{\tax{2}{behavControl}}{\tax{3}{zeroSh}}{\tax{4}{verPersistent}}{\tax{5}{extContxt}}{\tax{6}{llm}}.
This shares \dimens{d}{1} and \dimens{d}{4} coordinates with Layer~2 but differs on \dimens{d}{3}, a distinction invisible in single-axis taxonomies but consequential for change control.

The \frameTax makes visible that these three layers operate on different \dimens{D}{1} mechanisms, serve different \dimens{D}{2} goals, and exhibit different \dimens{D}{4} persistence profiles, even though all three contribute to the system's clinical behavior. A compliance officer can use this decomposition to identify which layers may require separate documentation, separate risk assessment, or separate regulatory review under the applicable pathway. The taxonomy does not determine which procedures apply; it ensures the decomposition needed to answer that question is explicit. Without it, all three layers would be conflated under ``the model was adapted,'' obscuring the distinct dimensional profiles, and therefore the distinct regulatory considerations, of each.

\subsection{Trustworthiness and Responsible Adaptation}
\label{sec:Trustworthiness}
Beyond specific regulatory regimes, the broader trustworthiness agenda for \gls{fm} deployments intersects with adaptation at three points that the \frameTax keeps separable. 
Within the medical domain, this agenda is increasingly formalized by clinical-AI-governance frameworks such as the FUTURE-AI consensus guidelines~\cite{lekadir_future-ai_2025} and AI-specific reporting standards like TRIPOD+AI~\cite{collins_tripodai_2024}, which emphasize traceability, fairness, and continuous evaluation.
\textit{Alignment, reasoning, and specialization are distinct \dimens{D}{2} goals, not synonyms.} 
A model well-specialized for a clinical domain (\tax{2}{task}) but poorly aligned (\tax{2}{algn}) may produce harmful outputs; 
a well-aligned model that lacks reasoning depth (\tax{2}{reas}) may fail at complex cases; a model that reasons well but is unspecialized may be unhelpful. \gls{rlhf}, \gls{dpo}, and \gls{rlvr} target these goals through different mechanisms (\dimens{D}{1}) and data regimes (\dimens{D}{3}), so trustworthiness governance should specify which goal a given intervention addresses rather than collapsing all three under ``post-training.''

Furthermore, explicitly classifying adaptations under \tax{2}{fairness}—and employing targeted gradient-free interventions like \gls{leace} (Row~\ref{tech:leace}) to  
apply a mathematically specified projection intended to remove selected protected-attribute concepts from model
representations—may support evidence relevant to \gls{fda} \gls{gmlp} principles, \gls{mdr} clinical-evaluation equity expectations, and EU AI Act bias-mitigation requirements.
Meanwhile, \tax{2}{reliability} 
may help establish
risk-based decision thresholds for medical \gls{cdss}. 
The clinical regulatory-science community has actively highlighted that static evaluation of such systems is inadequate; models require structured protocols for continual monitoring and updating post-deployment to maintain reliability safely~\cite{muehlematter_approval_2021,gilbert_algorithm_2021}.

\textit{\dimens{D}{3} is the data-provenance and annotator-bias axis.} 
Techniques using \pref (\gls{rlhf}, \gls{dpo}) introduce annotator-selection and labeling-guideline documentation that techniques using only existing labeled data do not~\cite{hausenloy_towards_2025}. 
Techniques using \tax{3}{verify} (\gls{rlvr}) shift the documentation burden to oracle correctness rather than annotator bias. 
Techniques using \tax{3}{ext} (\gls{rag}) require documentation of retrieval-source quality, currency, and licensing. 
Dynamic benchmarking platforms that evaluate models across multiple \dimens{D}{2} goals simultaneously~\cite{huang_trustworthiness_2025} are the operational complement: an adaptation that improves the target \dimens{D}{2} goal must not regress on safety-related \dimens{D}{2} goals.

Assigning an intervention \tax{2}{rem} 
may serve as one technical component
of ``right to be forgotten'' workflows.
\gls{gdpr} and similar regimes may motivate technical workflows involving Machine Unlearning (Row~\ref{tech:mUnLrn}) or Knowledge Editing (Row~\ref{tech:ke}), but these methods do not by themselves establish legal compliance.
Current methods are approximate rather than provable: benign relearning attacks can reverse unlearning~\cite{hu_unlearning_2024}, and \gls{rome}-style edits degrade after $\sim$10 sequential applications~\cite{hsueh_editing_2024}. 
Governance frameworks should therefore specify thresholds for knowledge suppression and post-edit validation rather than assume that ``remove'' is a single operation.

\section{Conclusion and Future Directions}
\label{sec:Conclusion}

\subsection{Summary of Contributions}
This paper proposed a six-dimensional \frameTax for \gls{ml} model adaptation. 
The \frameTax decomposes adaptation into Mechanism (\dimens{D}{1}), Goal (\dimens{D}{2}), Data Requirements (\dimens{D}{3}), Persistence (\dimens{D}{4}), Scope (\dimens{D}{5}), and Model Type (\dimens{D}{6}), with \dimens{D}{1}--\dimens{D}{5} 
reporting canonical properties of the declared intervention
and \dimens{D}{6} modulating feasibility by target model. Table~\ref{tab:centerpiece} classifies \NbTech techniques on all six dimensions.
The governance mapping connects \dimens{D}{2} and \dimens{D}{4} to potential documentation considerations under the \gls{nist} \gls{ai} \gls{rmf}, the EU AI Act, the EU \gls{mdr}/\gls{ivdr}, and the \gls{fda} \gls{pccp}.

\subsection{Scope Boundaries and Limitations}
While the \frameTax is broad, its design incorporates specific boundaries to maintain taxonomic precision:
(a)
\textbf{Model Continuum Simplification.} The \dimens{D}{6} hierarchy discretizes a richer continuum, leaving active-research boundaries (e.g., adversarial training for \glspl{mllm}) fluid;
(b)
\textbf{Isolation vs. Composition.} The \frameTax characterizes individual techniques in isolation, whereas practical deployments layer them in stacks (as modeled in Section \ref{sec:Combinations});
(c)
\textbf{Classificatory vs. Operational Focus.} The \frameTax provides the structural vocabulary but leaves the operational governance of deployment decisions to complementary architectures like \gls{aegis}~\cite{aegis_preprint_2025};
(d)
\textbf{\gls{mllm} Exhaustiveness.} The \gls{mllm}-specific rows (\ref{tech:crossModel}--\ref{tech:multimodInstr}) establish the branch structure but defer a comprehensive survey of the rapidly proliferating \gls{mllm} landscape to future work;
and (e)
\textbf{Empirical Benchmarking.} The \frameTax maps technical properties rather than adoption trends or cross-technique performance comparisons; empirical landscape characterization remains a priority for future work.

\subsection{Future Directions}
\label{sec:FutureDirections}
\textbf{Agentic AI and adaptation orchestration.} 
Agentic AI~\cite{singh_agentic_2025} is not a new adaptation technique but an orchestration pattern composing \gls{ce}, \gls{pe}, and \gls{rag} into a behavioral loop, 
with per-request execution often operating as \tax{4}{eph} through \tax{1}{ctx}, while maintained memory and retrieval configurations can be \tax{4}{verPersistent}.
As agents become more capable, the boundary between inference-time orchestration and persistent adaptation may blur: agents that fine-tune their own sub-models, learn user preferences through interaction, or accumulate long-term memory will require the \frameTax to address \textit{meta-adaptation}, a natural extension of the \dimens{D}{2} and \dimens{D}{4} axes (e.g., \tax{2}{lrnAdpt} and \tax{4}{gradAccm}).

Additional work is needed on \gls{fti} pipeline standardization, privacy-preserving combinations of \gls{fl} and \gls{peft}, and adaptation interfaces for new multimodal data.

An access-regime descriptor, i.e., white-box, grey-box, or black-box access, may also be evaluated as a future governance overlay because access constrains both technical feasibility and the modifications that can be operationally controlled.
\section*{List of Abbreviations}

\printnoidxglossary[
    type=acronym,
    title={}
]
\section*{Author Contributions}
\noindent
\textbf{F. Afdideh}: Methodology, Software, Validation, Investigation, Data Curation, Formal Analysis, Writing -- Original Draft, Writing -- Review \& Editing, Visualization.
\textbf{F. Seoane}: Supervision, Writing -- Review \& Editing.
\textbf{F. Abtahi}: Conceptualization, Methodology, Software, Validation, Investigation, Formal Analysis, Writing -- Original Draft, Writing -- Review \& Editing, Project Administration, Funding Acquisition.

\section*{Code Availability}
\noindent 
All project materials, including the source code, are available through the GitHub repository: \url{https://github.com/ki-smile/post-training-taxonomy}.

\begin{acks}
This study was supported by the Stockholm Medical Artificial Intelligence and Learning Environments (SMAILE) core facility at Karolinska Institutet and by EU grant 240037 from EIT Health. The authors used large language models for English proofreading and readability improvements. These tools did not contribute to conceptual content, data analysis, or scientific conclusions. All cited sources were independently verified, and all interpretations remain the sole responsibility of the authors.
\end{acks}

\bibliographystyle{ACM-Reference-Format} 

\bibliography{acm_references}

\appendix

\section{Methodology and Corpus Flow}
\label{appendix:methodology}

\subsection{Multivocal Literature Review Approach}
\label{appendix:mlr}
This study adopts a \gls{mlr} methodology, integrating peer-reviewed academic literature with practitioner-oriented grey literature to construct a comprehensive taxonomy of \gls{ai} model adaptation techniques~\cite{garousi_guidelines_2019}. The \gls{mlr} approach is warranted because \gls{ai} adaptation is heavily driven by industry practice: emergent techniques frequently appear in framework documentation, developer blogs, and industry technical reports months or years before formal academic publication. Restricting the review to peer-reviewed sources alone would omit rapidly evolving concepts, such as \gls{ce}, that have already achieved widespread practitioner adoption but lack formal academic treatment.
The rationale for grey literature inclusion is developed in Section~\ref{sec:grey-rationale}.

\subsection{Search Strategy and Snowball Sampling}
\label{sec:SearchStrategy}
Because terms such as \textit{\gls{ft}}, \textit{adaptation}, and \textit{\gls{tl}} are heavily overloaded across the \gls{ml} literature, standard database keyword searches yield fragmented and inconsistent results. We therefore adopted a purposive snowball sampling methodology, starting from canonical seed papers and tracing the semantic evolution of adaptation terminology through two complementary directions: \textit{backward snowballing}, examining reference lists of seed papers to identify foundational works that established the core terminology and theoretical underpinnings of each adaptation paradigm; and \textit{forward snowballing}, identifying citing papers via Google Scholar, Semantic Scholar, and arXiv to discover emergent variants, extensions, and successor methods that refine or supersede the seed techniques.

Twelve academic snowballing chains were traced, each anchored by a canonical seed paper or survey cluster:
\begin{enumerate}
\item \gls{peft} 
(\cite{hu_lora_2022});
\item \gls{tl} 
(\cite{pan_survey_2010} and~\cite{zhuang_comprehensive_2021});
\item Model Merging and Task Arithmetic 
(\cite{ilharco_editing_2023});
\item \gls{rl} from Verifiable Rewards 
(\cite{deepseek_r1_2025} and~\cite{shao_deepseekmath_2024});
\item Alignment Methods 
(\cite{rafailov_direct_2023} and~\cite{ouyang_training_2022});
\item \gls{cl} 
(\cite{shi_continual_2024});
\item Meta-Learning 
(\cite{finn_model-agnostic_2017} and~\cite{hospedales_meta-learning_2021});
\item \gls{rag} 
(\cite{lewis_retrieval-augmented_2020});
\item Knowledge Editing 
(\cite{meng_locating_2022});
\item Machine Unlearning 
(\cite{cao_towards_2015});
\item \gls{repe} and Activation Steering 
(\cite{zou_representation_2023} and~\cite{turner_activation_2023});
\item Data Augmentation and Synthetic Data Generation 
(\cite{wang_survey_2024} and~\cite{zhou_survey_2024}).
\end{enumerate}

\subsection{Inclusion and Exclusion Criteria}
Sources were included if they met at least one of the following criteria: (a) they introduced, formalized, or significantly extended an adaptation technique applicable to \gls{ml} models; (b) they provided a systematic survey or taxonomic organization of adaptation methods; or (c) they constituted authoritative framework documentation or technical reports from major \gls{ai} organizations describing adaptation tooling or methodology. Sources were excluded if they addressed foundational learning paradigms (e.g., supervised, unsupervised, or reinforcement learning) without an adaptation context, or if they focused exclusively on low-level optimization mechanics (e.g., \gls{gd} algorithms, learning rate scheduling) rather than adaptation strategies.

\subsection{Grey Literature}
To capture practitioner terminology and tooling that precedes formal publication, the review incorporated four categories of grey literature:
\begin{itemize}
\item \textit{Framework documentation}: official documentation for Hugging Face \gls{peft} (v0.18.0), MergeKit (Arcee AI), \gls{trl}, ms-swift (ModelScope), and EasyEdit (Zhejiang University);
\item \textit{Developer blogs and technical reports} from Anthropic, LangChain, LlamaIndex, Weaviate, and the \gls{pe} Guide describing emergent techniques and best practices;
\item \textit{Industry analysis}, including reports from organizations such as Gartner tracking the adoption trajectory of adaptation terminology (e.g., the shift from \gls{pe} to \gls{ce});
\item \textit{Synthetic data and augmentation surveys}, comprising preprints and workshop papers covering data-centric adaptation methods that bridge academic and industrial practice.
\end{itemize}

\subsection{Corpus and Synthesis}
The corpus was assembled through a staged process: seed identification, 
backward snowballing,
forward snowballing, 
and grey literature integration.
Academic sources include foundational papers, tier-1 venue publications (NeurIPS, ICML, ICLR, ACL, IEEE TPAMI, ACM Computing Surveys), and high-impact arXiv preprints with substantial citation counts. Grey literature sources were included only if they met at least one of three criteria: (a) they constituted official framework documentation from a major AI organization; (b) they introduced or formalized a technique or term not yet covered in peer-reviewed literature; or (c) they provided implementation-level detail necessary to distinguish technique variants that appear identical in academic descriptions. Promotional material, opinion pieces, and sources without verifiable technical content were excluded.

\subsection{Saturation Criterion and Stopping Rule}
Two distinct criteria governed corpus growth. A candidate required a new row when its \dimens{D}{1}--\dimens{D}{6} profile, i.e., the combination of category values across all six dimensions, was not already represented, and it constituted an independently selectable adaptation decision unit. A candidate required a new dimension category when it could not be characterized by any existing value on some dimension. Saturation was defined as the point at which no further candidate required either.
The category structure was developed jointly with the corpus rather than fixed in advance, and was extended where the literature demanded. Several category values are instantiated solely by techniques traced through a single chain and entered the framework with that chain's integration: \tax{1}{actSpace} through the activation-steering chain (Row~\ref{tech:actSteer}), as direct algebraic intervention on intermediate representations is structurally distinct from gradient-based parametric update; 
\tax{2}{reas} and \tax{2}{verify} through the \gls{rlvr}/\gls{grpo} chain (Row~\ref{tech:rlvr}), whose seed literature also motivated \tax{1}{infTimeSearch} (Row~\ref{tech:ttCompute}); 
and \tax{2}{rem} with \tax{3}{forget} through the machine-unlearning chain (Row~\ref{tech:mUnLrn}).
The final \NbTech-technique set satisfies both criteria: every candidate identified through forward snowballing and grey-literature review, including those integrated late in the process, was either assignable to an existing row or accommodated by a row differing from all existing rows on at least one dimension, without requiring further extension of the category structure. Candidate paradigms that were evaluated and not added, including agentic orchestration patterns, were determined to represent compositions of existing techniques or open research questions rather than distinct adaptation techniques.
Chains are presented in Section \ref{sec:SearchStrategy} in thematic order; 
integration order differed and is not load-bearing for this criterion.
The \NbTech-technique count is the saturation count under the \dimens{D}{1}--\dimens{D}{6} criterion, not a claim that no further techniques exist.
Techniques whose \dimens{D}{1}--\dimens{D}{6} profile is shared with an existing row are subsumed under that row even when they carry distinct names in the literature. 
New rows are required only when a candidate presents a profile no existing row provides.

\subsection{Classification Process}
The synthesis proceeded iteratively: individual technique definitions were first extracted and cross-referenced across sources to resolve terminological inconsistencies, then organized into the multi-dimensional characterization framework.
Technique classifications were drafted by the first author and independently reviewed by all authors against the \dimens{D}{1}--\dimens{D}{6} category definitions; disagreements were resolved by discussion until consensus.
Classification tensions, that is, cases where techniques resist clean single-category assignment on one or more dimensions, were identified during this process and are documented transparently in Section~\ref{sec:ClassTensions}.
Taxonomy gaps, defined as areas where the reviewed literature revealed adaptation paradigms not yet integrated into existing frameworks, were identified through systematic comparison of the snowballing chains against the emerging taxonomy structure. 
Five such gaps were identified: \gls{rlvr}/\gls{grpo}, machine unlearning, knowledge editing, \gls{ce}, and activation steering. 
All five were subsequently resolved and integrated into the final framework.

\subsection{Literature Trace by Snowballing Chain}
\label{sec:LiteratureTrace}
The purposive snowball sampling strategy described in Section~\ref{sec:SearchStrategy}, combined with the grey-literature rationale of Section~\ref{appendix:mlr}, produced twelve independent chains. 
Tables~\ref{tab:peft-chain}--\ref{tab:augment-chain} report the full trace for each chain, separating backward-snowballing foundations, seed papers, and forward-snowballing successors, with the key development each source contributed.

\subsection{Search and Selection Process}\label{sec:search-process}

The corpus was assembled through the following staged process, summarized in Section~\ref{sec:prisma}:

\begin{enumerate}
\item \textbf{Seed identification.} \NbChains snowball chains were defined, each corresponding to a distinct adaptation paradigm (Tables~\ref{tab:peft-chain}--\ref{tab:augment-chain}). Each chain was anchored by one or two canonical seed papers selected for citation impact and foundational status; where a paradigm required both a seminal method paper and a comprehensive survey, two co-seeds were used. This yielded \total{seedcount} distinct seed papers across \NbChains chains.

\item \textbf{Backward snowballing.} For each seed, reference lists were examined to identify foundational works that established the core terminology and theoretical underpinnings of each adaptation paradigm. All \NbChains chains received backward snowballing; foundational references are listed per-chain in Tables~\ref{tab:peft-chain}--\ref{tab:augment-chain}. This yielded \total{backcount} foundational sources across all chains.

\item \textbf{Forward snowballing.} Citing papers were traced via Google Scholar, Semantic Scholar, and arXiv to discover emergent variants, extensions, and successor methods. This yielded \total{forwardcount} additional academic sources.

\item \textbf{Grey literature integration.} Framework documentation, developer blogs, technical reports, and industry analyses were incorporated where they captured practitioner terminology or tooling that precedes or complements formal publication. This yielded \nbGreyLit grey literature sources (see Section~\ref{sec:grey}).

\item 
\textbf{Exclusion.} 
Sources were excluded if they: (a)~addressed foundational learning paradigms without an adaptation context; (b)~focused exclusively on low-level optimization mechanics; or (c)~constituted promotional material without technical substance. 

\item \textbf{Final corpus.} Searches were conducted through July 2026.
The resulting corpus comprises \the\numexpr\totvalue{backcount}+\totvalue{seedcount}+\totvalue{forwardcount}\relax~academic sources,
spanning 1985--2026, 
and \nbGreyLit grey literature sources. 
Academic sources include foundational papers, tier-1 venue publications (NeurIPS, ICML, ICLR, ACL, IEEE TPAMI, ACM Computing Surveys), and high-impact arXiv preprints with substantial citation counts.
\end{enumerate}

\textbf{Dating convention and apparent temporal inversions.} Snowball chains operate on the citation graph, which is driven by \textit{preprint availability} rather than venue publication dates. Because of preprint-to-publication lag (typically 6--18 months for NeurIPS/ICML/ICLR, and up to two years for journal venues such as IEEE TPAMI or ACM Computing Surveys), a paper can legitimately appear in the reference list of a work whose venue year precedes or equals its own. For example, 
DeepSeekMath (2024, \cite{shao_deepseekmath_2024}) cites \textit{Let's Verify Step by Step} (ICLR 2024, \cite{lightman_lets_2023}) via its May 2023 preprint. Throughout this appendix, chain chronology 
follows first-preprint dates, while tables report 
its publication year. 

\subsection{Corpus Flow Summary} 
\label{sec:prisma}

While this review follows an \gls{mlr} rather than strict systematic literature review protocol, the following flow summarizes the corpus assembly process:

\begin{itemize}
\item Snowball chains: \NbChains (Tables~\ref{tab:peft-chain}--\ref{tab:augment-chain})
\item Seed: \total{seedcount} sources
\item Backward snowballing: \total{backcount} sources 
\item Forward snowballing: \total{forwardcount} sources
\item Grey literature: \nbGreyLit sources 
\item Final corpus: \the\numexpr\totvalue{backcount}+\totvalue{seedcount}+\totvalue{forwardcount}\relax~academic sources + \nbGreyLit grey literature sources
\item Techniques extracted: \NbTech distinct adaptation techniques (Rows~\ref{tech:fullFT}--\ref{tech:actSteer} of the centerpiece table; Row~\ref{tech:training}, Training, is a baseline reference, not a post-training technique)
\end{itemize}

\subsection{Snowballing Results by Adaptation Paradigm}
\label{sec:snowballing}
This section presents the results of our snowball sampling methodology, organized across \NbChains tables (Tables~\ref{tab:peft-chain}--\ref{tab:augment-chain}) that correspond to the core adaptation paradigms evaluated in this study. Each table maps the methodological evolution of a specific paradigm, structured around three distinct phases of our literature trace: 
(1) the foundational works identified through backward snowballing;
(2) the canonical seed papers selected to anchor the chain; and 
(3) the emergent variants, extensions, and successor methods discovered via forward snowballing. 
To contextualize each source, every entry provides the method's name, its reference, and a concise summary of the key development it contributed to the post-training adaptation landscape.

{\scriptsize
\begin{table}[htbp]
\renewcommand{\tabularxcolumn}[1]{p{#1}}
\caption{\gls{peft} snowballing results.}
\label{tab:peft-chain}
\adjustbox{width=.9\linewidth
, max totalheight=0.9\textheight
}{
\begin{tabularx}{\linewidth}{p{0.1\linewidth} X p{0.78\linewidth}}
\toprule
\multicolumn{1}{c}{\textbf{Method}} & \multicolumn{1}{c}{\textbf{Ref.}} & \multicolumn{1}{c}{\textbf{Key Development}} 
\\
\midrule
\multicolumn{3}{c}{\cellcolor{gray!20} \scriptsize \textit{\textbf{I. Backward Snowballing}}} \\
\midrule
---&\addback\cite{aghajanyan_intrinsic_2020}&
Shows that the success of \gls{ft} \glspl{llm} on small datasets is due to their extremely low intrinsic dimension and that larger models inherently possess lower intrinsic dimensions.
\\
Adapter Modules &\addback\cite{houlsby_parameter-efficient_2019}&
Enable highly parameter-efficient transfer learning by injecting a small number of trainable, task-specific parameters into a frozen pre-trained model.
\\
\midrule
\multicolumn{3}{c}{\cellcolor{gray!20} \scriptsize \textit{\textbf{II. Seed}}} \\
\midrule
\gls{lora}& \addseed\cite{hu_lora_2022} &
Reduces the computational cost and memory requirements of \gls{ft} \glspl{llm} by freezing pre-trained weights and injecting small, trainable rank decomposition matrices.
\\
\midrule
\multicolumn{3}{c}{\cellcolor{gray!20} \scriptsize \textit{\textbf{III. Forward Snowballing}}} \\
\midrule
\gls{dora} & 
\addforward\cite{liu_dora_2024}
& 
Bridges the accuracy gap between \gls{lora} and full \gls{ft} by decomposing pre-trained weights into magnitude and direction, using \gls{lora} solely for efficient directional updates to improve learning capacity without adding inference overhead.
\\
\gls{adalora} & 
\addforward\cite{zhang_adalora_2023}
& 
Dynamically allocates the parameter budget across weight matrices based on their importance, utilizing \gls{svd} to efficiently prune less critical updates.
\\
\gls{vera} & 
\addforward\cite{kopiczko_vera_2023}
& 
Dramatically reduces the parameter and storage footprint of \gls{lora} by sharing a single, frozen pair of random low-rank matrices across all layers and learning only small, layer-specific scaling vectors.
\\
\gls{qlora} & 
\addforward\cite{dettmers_qlora_2023} &  
Enables the \gls{ft} of \glspl{llm} (up to 65B parameters) on a single 48GB GPU by combining a frozen, 4-bit quantized base model with \gls{lora}, maintaining full 16-bit performance via memory-saving innovations like 4-bit NormalFloat, Double Quantization, and Paged Optimizers.
\\
\gls{rslora} & 
\addforward\cite{kalajdzievski_rank_2023}
& 
Modifies the traditional scaling factor,  
thereby unlocking the effective use of higher-rank adapters to achieve better \gls{ft} performance.
\\
\gls{lora}+ & 
\addforward\cite{hayou_lora_2024}
& 
Improves the \gls{ft} speed and performance of large-width models by assigning different learning rates to the adapter matrices A and B.
\\
\gls{pissa} & 
\addforward\cite{meng_pissa_2024} & 
Accelerates \gls{peft} by initializing adapter matrices with the original model's principal singular components and freezing the residual weights, resulting in faster convergence, reduced quantization error, and superior task performance.
\\
\gls{lora}-Pro & 
\addforward\cite{wang_lora-pro_2024} & 
Adjusts the gradients of \gls{lora}'s low-rank matrices to more accurately approximate the full \gls{ft} gradient.
\\
\gls{dude} & 
\addforward\cite{han_dual_2025}
& 
Decomposes weight matrices into magnitude and direction components and utilizes \gls{svd} for principled initialization.
\\
\gls{fedpara} & 
\addforward\cite{hyeon-woo_fedpara_2021}
& 
A communication-efficient parameterization technique for \gls{fl} that utilizes a low-rank Hadamard product to dramatically reduce model upload and download costs without sacrificing model capacity, while also offering an extension for highly efficient personalized \gls{fl}.
\\
\gls{dylora} &  
\addforward\cite{valipour_dylora_2023} & 
Dynamically trains adapter modules across a range of ranks simultaneously rather than a single fixed rank, eliminating the need for exhaustive rank search and enabling flexible rank adjustments post-training without the need to retrain from scratch.
\\
\gls{moelora} & 
\addforward\cite{luo_moelora_2024} & 
Treats \gls{lora} modules as a \gls{moe} and employs contrastive learning to encourage distinct feature learning among these experts.
\\
\gls{ia3} 
& 
\addforward\cite{liu_few-shot_2022} & 
Scales internal activations by learned vectors, and pairs it with the T-Few recipe.
\\
\gls{milora_svd} & \addforward\cite{wang_milora_2025} & Improves \gls{ft} by restricting updates to the minor singular components of the weight matrix, which represent noisy or long-tail information, while keeping the principal components frozen to preserve essential pre-trained knowledge.
\\
\bottomrule
\end{tabularx}
}
\end{table}
}

{\scriptsize
\begin{table}[htbp]
\renewcommand{\tabularxcolumn}[1]{p{#1}}
\caption{\gls{tl} snowballing results.}
\label{tab:tl-chain}
\adjustbox{width=.9\linewidth
}{
\begin{tabularx}{\linewidth}{p{0.1\linewidth} X p{0.78\linewidth}}
\toprule
\multicolumn{1}{c}{\textbf{Method}} & \multicolumn{1}{c}{\textbf{Ref.}} & \multicolumn{1}{c}{\textbf{Key Development}} 
\\
\midrule
\multicolumn{3}{c}{\cellcolor{gray!20} \scriptsize \textit{\textbf{I. Backward Snowballing}}} \\
\midrule
\gls{mtl}
& \addback\cite{caruana_multitask_1997}&
Improves model generalization by simultaneously training related tasks on a shared representation, allowing the domain information from one task to act as a beneficial inductive bias for the others.\\
Mixture Model &\addback\cite{iii_domain_2006}&
Uses conditional expectation maximization to adapt maximum entropy classifiers to new domains, effectively leveraging abundant out-of-domain training data alongside scarce in-domain data to improve predictive performance.\\
---&\addback\cite{shimodaira_improving_2000}&
Improves predictive inference under covariate shift by weighting the log-likelihood function using the ratio of population to observation covariate densities, demonstrating asymptotic optimality under model misspecification and introducing an information criterion to select optimal weights for moderate sample sizes.\\
Self-Taught Learning &\addback\cite{raina_self-taught_2007}&
Leverages vast amounts of random, unstructured unlabeled data
using sparse coding to construct higher-level feature representations that significantly improve performance on supervised classification tasks.\\
---&\addback\cite{yosinski_how_2014}&
Experimentally quantifies feature transferability across deep neural network layers, revealing that while transfer is hindered by higher-layer specialization and the splitting of co-adapted neurons, initializing networks with transferred features yields a lasting generalization boost even across distant tasks.\\
\gls{bert}&\addback\cite{devlin_bert_2019}&
Pre-trains deep bidirectional representations that condition on both left and right context across all layers.
\\
\midrule
\multicolumn{3}{c}{\cellcolor{gray!20} \scriptsize \textit{\textbf{II. Seed}}} \\
\midrule
---& \addseed\cite{pan_survey_2010} &
Formalizes how knowledge can be effectively transferred across different feature spaces and data distributions to overcome the standard \gls{ml} assumption of identical training and testing distributions.
\\
---& \addseed\cite{zhuang_comprehensive_2021} &
Reviews over 40 representative approaches from data and model perspectives and provides an extensive empirical evaluation of more than 20 models across multiple datasets to guide practical model selection.
\\
\midrule
\multicolumn{3}{c}{\cellcolor{gray!20} \scriptsize \textit{\textbf{III. Forward Snowballing}}} \\
\midrule
---&\addforward\cite{ma_deep_2024}&
Measures differences in internal dependence structures separately from marginal distributions, optimizing their relative weights to robustly enhance transferability where domain shifts matter most.
\\
---&\addforward\cite{guo_comprehensive_2024}&
Reviews  
\gls{ftl}, outlining how 
it solves challenges like data heterogeneity and scarcity while preserving privacy constraints across decentralized nodes.
\\
---&\addforward\cite{blanchard_generalizing_2011}&
Introduces a distribution-free, kernel-based framework that utilizes reproducing kernel Hilbert spaces and regularized empirical risk minimization to achieve universal consistency when generalizing classification labels from multiple related training datasets to a new, unlabeled sample.
\\
---&\addforward\cite{wang_generalizing_2023}&
Reviews theoretical foundations and categorizes existing algorithms into data manipulation, representation learning, and learning strategies to facilitate the development of models that maintain robust performance when deployed in unseen, out-of-distribution environments.
\\
\bottomrule
\end{tabularx}
}
\end{table}
}

{\scriptsize
\begin{table}[htbp]
\caption{Model merging snowballing results.}
\label{tab:merging-chain}
\adjustbox{width=.9\linewidth
}{
\begin{tabularx}{\linewidth}{p{0.1\linewidth} X p{0.78\linewidth}}
\toprule
\multicolumn{1}{c}{\textbf{Method}} & \multicolumn{1}{c}{\textbf{Ref.}} & \multicolumn{1}{c}{\textbf{Key Development}} 
\\
\midrule
\multicolumn{3}{c}{\cellcolor{gray!20} \scriptsize \textit{\textbf{I. Backward Snowballing}}} \\
\midrule
Model Soups &\addback\cite{wortsman_model_2022}&
Improves model accuracy and robustness by averaging the weights of multiple models fine-tuned with different hyperparameter configurations,  
without increasing inference or memory costs.\\
Fisher Merging &\addback\cite{matena_merging_2022}&
Combines models by computing a weighted average of their parameters based on their Fisher information matrices.
\\
\gls{swa} &\addback\cite{izmailov_averaging_2019}&
Improves generalization and finds flatter loss optima by averaging multiple weights sampled along the \gls{sgd} training trajectory.
\\
\midrule
\multicolumn{3}{c}{\cellcolor{gray!20} \scriptsize \textit{\textbf{II. Seed}}} \\
\midrule
Task Arithmetic &\addseed\cite{ilharco_editing_2023}&
A paradigm for model editing that represents task-specific behavioral shifts as "task vectors"
(the difference between fine-tuned and pre-trained model weights)
which can be intuitively added, negated, or combined to steer model performance across multiple tasks without additional training.
\\
\midrule
\multicolumn{3}{c}{\cellcolor{gray!20} \scriptsize \textit{\textbf{III. Forward Snowballing}}} \\
\midrule
\gls{ties} & 
\addforward\cite{yadav_ties-merging_2024} & 
Enables the effective combination of multiple task-specific models into a single multitask model by mitigating parameter interference through a three-step process: pruning redundant parameter changes, resolving cross-model sign conflicts, and exclusively merging parameters that align with the agreed-upon sign. 
\\
\gls{dare} & 
\addforward\cite{yu_language_2024}
& 
Eliminates the vast majority of redundant \gls{ft} deltas to mitigate interference, enabling multiple task-specific language models to be fused into a single, high-performing model without the need for retraining.
\\
Model Breadcrumbs & 
\addforward\cite{davari_model_2025}
&  
Creates sparse, task-specific weight updates (derived from the delta between pre-trained and fine-tuned weights) to enable efficient, hyperparameter-free integration of multiple task adaptations into a single foundation model.
\\
\gls{della}, \gls{magprune} & 
\addforward\cite{deep_della-merging_2024}
&  
Assigns higher dropout probabilities to parameters with smaller values and applies a compensatory rescaling factor to preserve embedding integrity.
\\
\gls{emr} & 
\addforward\cite{huang_emr-merging_2024} 
& 
Aligns multiple task-specific models with a unified base model by generating lightweight, task-specific masks and rescalers, thereby enabling multi-task capabilities without requiring additional data or training.
\\
\gls{slerp} & \addforward\cite{shoemake_animating_1985} & 
Interpolates between compatible model checkpoints along aspherical path. The method originates in spherical interpolation for rotations and is implemented in contemporary \gls{llm} model-merging toolchains, including MergeKit. It has also been evaluated empirically as a model-merging strategy \cite{shoemake_animating_1985,lu_fine-tuning_2025}.
\\
\gls{lata} & \addforward\cite{chen_layer-aware_2025} & 
Assigns layer-specific weights to task vectors, isolating task-specific knowledge from shared instruction-following components.
\\
\bottomrule
\end{tabularx}
}
\end{table}
}

{\scriptsize
\begin{table}[htbp]
\caption{\gls{rlvr}/\gls{grpo} snowballing results.}
\label{tab:rlvr-chain}
\adjustbox{width=.9\linewidth
}{
\begin{tabularx}{\linewidth}{p{0.1\linewidth} X p{0.78\linewidth}}
\toprule
\multicolumn{1}{c}{\textbf{Method}} & \multicolumn{1}{c}{\textbf{Ref.}} & \multicolumn{1}{c}{\textbf{Key Development}} 
\\
\midrule
\multicolumn{3}{c}{\cellcolor{gray!20} \scriptsize \textit{\textbf{I. Backward Snowballing}}} \\
\midrule
---&\addback\cite{christiano_deep_2017}&
Enables complex goal specification through human preference feedback between trajectory segments.
\\
InstructGPT &\addback\cite{ouyang_training_2022}& 
Aligns \glspl{llm} with human intent by using \gls{sft} on demonstrations followed by \gls{rlhf}.
\\
\gls{ppo}&\addback\cite{schulman_proximal_2017}&
A surrogate objective function that facilitates multiple epochs of minibatch updates.
\\
---&\addback\cite{lightman_lets_2023}&
Demonstrates how providing human feedback at each intermediate step of a logical chain 
outperforms feedback provided only at the final result 
for training reliable \glspl{llm} on complex mathematical reasoning tasks.
\\
\midrule
\multicolumn{3}{c}{\cellcolor{gray!20} \scriptsize \textit{\textbf{II. Seed}}} \\
\midrule
DeepSeekMath, \gls{grpo} &\addseed\cite{shao_deepseekmath_2024}&
Utilizes a curated pipeline for high-quality web data and introducing \gls{grpo}, a memory-efficient alternative to \gls{ppo} that optimizes reasoning performance.
\\
DeepSeek-R1 &\addseed\cite{deepseek_r1_2025}&
Incentivizes advanced reasoning capabilities, such as self-reflection and dynamic verification, directly in \glspl{llm} without relying on human-annotated reasoning trajectories.
\\
\midrule
\multicolumn{3}{c}{\cellcolor{gray!20} \scriptsize \textit{\textbf{III. Forward Snowballing}}} \\
\midrule
\gls{dapo} & 
\addforward\cite{yu_dapo_2025}
& 
Enhances long-chain-of-thought \gls{rl} through four technical innovations (Clip-Higher, Dynamic Sampling, Token-Level Policy Gradient Loss, and Overlong Reward Shaping) designed to improve training stability, efficiency, and reasoning performance.
\\
\gls{gspo} & \addforward\cite{zheng_group_2025} & 
Shifts from traditional token-level importance ratios to sequence-level optimization
to achieve enhanced stability and performance, particularly for \gls{moe} architectures.
\\
\gls{sapo} & \addforward\cite{zheng_sapo_2026} &  
Treats individual reasoning steps as the fundamental unit for credit assignment, which stabilizes training and improves performance.
\\
\gls{cispo} & \addforward\cite{minimax_minimax-m1_2025} & Enhances training efficiency by clipping importance sampling weights rather than token updates. 
\\
\gls{rloo} & 
\addforward\cite{kool_buy_2019}
& 
Introduces a variance-reducing REINFORCE estimator that draws multiple samples without replacement per datapoint to establish a natural baseline.
\\
REINFORCE++ & \addforward\cite{hu_reinforce_2025} & 
a critic-free framework that utilizes global advantage normalization across the entire batch rather than local prompt-specific groups, providing an effectively unbiased and stable estimate.
\\
\bottomrule
\end{tabularx}
}
\end{table}
}

{\scriptsize
\begin{table}[htbp]
\caption{Alignment snowballing results.}
\label{tab:alignment-chain}
\adjustbox{width=.9\linewidth
}{
\begin{tabularx}{\linewidth}{p{0.1\linewidth} X p{0.78\linewidth}}
\toprule
\multicolumn{1}{c}{\textbf{Method}} & \multicolumn{1}{c}{\textbf{Ref.}} & \multicolumn{1}{c}{\textbf{Key Development}} 
\\
\midrule
\multicolumn{3}{c}{\cellcolor{gray!20} \scriptsize \textit{\textbf{I. Backward Snowballing}}} \\
\midrule
---&\addback\cite{ziegler_fine-tuning_2020}&
Introduces a framework for applying \gls{rl} to natural language tasks by training a reward model based on human preference comparisons, demonstrating effective performance on stylistic continuation and summarization tasks.
\\
\midrule
\multicolumn{3}{c}{\cellcolor{gray!20} \scriptsize \textit{\textbf{II. Seed}}} \\
\midrule
InstructGPT &
\addseed
\cite{ouyang_training_2022}&
Aligns \glspl{llm} with human intent by using \gls{sft} on demonstrations followed by \gls{rlhf}.
\\
\gls{dpo}&\addseed\cite{rafailov_direct_2023}&
Mathematically bypasses the complex, unstable \gls{rlhf} pipeline by mapping reward functions directly to optimal policies, allowing for stable, single-stage fine-tuning based on human preference classification.
\\
\midrule
\multicolumn{3}{c}{\cellcolor{gray!20} \scriptsize \textit{\textbf{III. Forward Snowballing}}} \\
\midrule

DeepSeekMath, \gls{grpo} &
\addforward
\cite{shao_deepseekmath_2024}&
Utilizes a curated pipeline for high-quality web data and introducing \gls{grpo}, a memory-efficient alternative to \gls{ppo} that optimizes reasoning performance.
\\
\gls{orpo} & 
\addforward\cite{hong_orpo_2024}
& 
Eliminates the need for a separate preference alignment phase by embedding preference optimization directly into \gls{sft} using a minor odds ratio penalty for the disfavored style.
\\
\gls{simpo} & 
\addforward\cite{meng_simpo_2024}
&  
A reference-free offline preference optimization algorithm that uses the average sequence log probability as an implicit reward and incorporates a target reward margin.
\\
\gls{kto} & 
\addforward\cite{ethayarajh_model_2024}
& 
An alignment objective rooted in prospect theory that directly maximizes the human utility of model generations, matching or exceeding preference-based methods using only a simple binary signal of desirability.
\\
\gls{cpo} & 
\addforward\cite{xu_contrastive_2024}
& 
Improves machine translation by teaching models to avoid merely adequate but imperfect translations.
\\
\gls{mallows} & 
\addforward\cite{chen_mallowspo_2024}
&  
An offline preference optimization algorithm inspired by Mallows' ranking theory that incorporates a dispersion index to model the diversity of human preferences.
\\
\gls{dapo} & 
\addforward
\cite{yu_dapo_2025}
& 
Utilizes four key techniques to scale \gls{rl} and elicit complex reasoning in large language models.
\\
\gls{ipo} & 
\addforward\cite{azar_general_2024}
& 
A general objective that bypasses the pointwise reward approximations by learning directly from pairwise preferences.
\\
\bottomrule
\end{tabularx}
}
\end{table}
}

{\scriptsize
\begin{table}[htbp]
\caption{\gls{cl} snowballing results.}
\label{tab:cl-chain}
\adjustbox{width=.9\linewidth
}{
\begin{tabularx}{\linewidth}{p{0.1\linewidth} X p{0.78\linewidth}}
\toprule
\multicolumn{1}{c}{\textbf{Method}} & \multicolumn{1}{c}{\textbf{Ref.}} & \multicolumn{1}{c}{\textbf{Key Development}} 
\\
\midrule
\multicolumn{3}{c}{\cellcolor{gray!20} \scriptsize \textit{\textbf{I. Backward Snowballing}}} \\
\midrule
Catastrophic Interference &\addback\cite{mccloskey_catastrophic_1989}&
Sequential learning in connectionist networks causes new information to destructively overwrite previously learned representations. 
This phenomenon explains the catastrophic forgetting observed in modern full \gls{ft}.
\\
\gls{ews} &\addback\cite{kirkpatrick_overcoming_2017}& 
A method inspired by synaptic consolidation in neuroscience that overcomes catastrophic forgetting by identifying and selectively slowing down updates to weights that are critical for previously learned tasks.
\\
---&\addback\cite{bengio_empirical_2014}&
Compares various gradient-based training algorithms and activation functions, identifying dropout as the regularization technique that consistently provides the best stability–plasticity trade-off.
\\
\midrule
\multicolumn{3}{c}{\cellcolor{gray!20} \scriptsize \textit{\textbf{II. Seed}}} \\
\midrule
---&\addseed\cite{shi_continual_2024}&
A survey, 
defining the landscape through "vertical" (general-to-specific) and "horizontal" (across time/domains) continuity, and structuring research into three distinct stages: Continual Pre-Training, Domain-Adaptive Pre-training, and Continual \gls{ft}.
\\
\midrule
\multicolumn{3}{c}{\cellcolor{gray!20} \scriptsize \textit{\textbf{III. Forward Snowballing}}} \\
\midrule
---&\addforward\cite{wang_comprehensive_2024}&
A survey that 
establishes a taxonomy of methods, clarifies the stability-plasticity trade-off, and evaluates how various strategies (such as regularization, replay, and architecture-based approaches) manage the stability-plasticity trade-off.
\\
\bottomrule
\end{tabularx}
}
\end{table}
}

{\scriptsize
\begin{table}[htbp]
\caption{Meta-learning snowballing results.}
\label{tab:meta-chain}
\adjustbox{width=.9\linewidth
}{
\begin{tabularx}{\linewidth}{p{0.1\linewidth} X p{0.78\linewidth}}
\toprule
\multicolumn{1}{c}{\textbf{Method}} & \multicolumn{1}{c}{\textbf{Ref.}} & \multicolumn{1}{c}{\textbf{Key Development}} 
\\
\midrule
\multicolumn{3}{c}{\cellcolor{gray!20} \scriptsize \textit{\textbf{I. Backward Snowballing}}} \\
\midrule
Meta-Learning &\addback\cite{thrun_learning_1998}&
Enables machines to improve their own generalization capabilities through experience, mirroring the human ability to acquire "bias" and learn effectively from extremely sparse data.\\
Matching Networks &\addback\cite{vinyals_matching_2016}&
Combines metric learning with external memory to enable one-shot learning by directly predicting labels from a support set of examples without requiring task-specific \gls{ft}.
\\
Prototypical Networks &\addback\cite{snell_prototypical_2017}&
Simplifies few-shot classification by learning a metric space where new examples are classified based on their distance to class-specific prototype representations, demonstrating that a straightforward inductive bias can outperform complex meta-learning architectures.
\\
\midrule
\multicolumn{3}{c}{\cellcolor{gray!20} \scriptsize \textit{\textbf{II. Seed}}} \\
\midrule
\gls{maml}&\addseed\cite{finn_model-agnostic_2017}&
An optimization-based meta-learning algorithm that explicitly trains model parameters to be highly sensitive to gradient updates, allowing a model to achieve strong generalization on new tasks after only a few training samples and minimal gradient steps.
\\
---   
&\addseed\cite{hospedales_meta-learning_2021}& This survey establishes a comprehensive meta-learning taxonomy and evaluates the field's progress in overcoming deep learning bottlenecks, such as data scarcity and generalization challenges, by enabling algorithms to improve themselves through experience across multiple tasks.\\
\midrule
\multicolumn{3}{c}{\cellcolor{gray!20} \scriptsize \textit{\textbf{III. Forward Snowballing}}} \\
\midrule
\gls{maml}++ & 
\addforward\cite{antoniou_how_2018}
& 
Improvements to the \gls{maml} framework that resolves its inherent training instability, hyperparameter sensitivity, and computational overhead while significantly improving generalization and convergence speed.
\\
\gls{fomaml}, Reptile & 
\addforward\cite{nichol_first-order_2018}
& 
Explicitly optimizes a model's initial parameters so that produces rapid adaptation and high generalization across diverse learning tasks.
An efficient \gls{fomaml} that learns an adaptable parameter initialization by repeatedly moving the initial weights toward the weights learned on sampled tasks, eliminating the need for computationally expensive second-order derivatives.
\\
Meta-\gls{sgd} & 
\addforward\cite{li_meta-sgd_2017}
& 
Extends upon \gls{maml} by simultaneously learning a model's optimal initialization, update directions, and learning rates to enable highly efficient, one-step adaptation to new tasks.
\\
\gls{leo} & 
\addforward\cite{rusu_meta-learning_2018}
& 
Bypasses the limitations of high-dimensional parameter spaces in \gls{fsl} by learning a data-dependent generative representation and performing gradient-based meta-adaptation entirely within a low-dimensional latent space.
\\
\gls{cavia} & \addforward\cite{zintgraf_fast_2019} & 
Improves upon \gls{maml} by partitioning model parameters into globally shared weights and task-specific context parameters, restricting test-time adaptation solely to the low-dimensional context to reduce overfitting and improve efficiency.
\\
\bottomrule
\end{tabularx}
}
\end{table}
}

{\scriptsize
\begin{table}[htbp]
\caption{\gls{rag} snowballing results.}
\label{tab:rag-chain}
\adjustbox{width=.9\linewidth
}{
\begin{tabularx}{\linewidth}{p{0.1\linewidth} X p{0.78\linewidth}}
\toprule
\textbf{Method} 
& \textbf{Ref.} & \textbf{Key Development} 
\\
\midrule
\multicolumn{3}{c}{\cellcolor{gray!20} \scriptsize \textit{\textbf{I. Backward Snowballing}}} \\
\midrule
\gls{dpr}&\addback\cite{karpukhin_dense_2020}&
A method for open-domain question answering that replaces traditional sparse retrieval models (like TF-IDF or BM25) with dense embeddings learned via a dual-encoder framework. 
\\
\gls{prf}&\addback\cite{robertson_probabilistic_2009}&
Presents the \gls{prf} conceptually,
incorporating document metadata, such as structure and link-graph information, culminating in the BM25F algorithm for highly effective web and corporate search retrieval.
\\
\gls{retro}&\addback\cite{borgeaud_improving_2022}&
Achieves performance comparable to massive models like \gls{gpt}-3 with 25 times fewer parameters by augmenting an auto-regressive language model with a frozen \gls{bert} retriever and chunked cross-attention to condition predictions on document chunks retrieved from a 2-trillion-token database.
\\
\midrule
\multicolumn{3}{c}{\cellcolor{gray!20} \scriptsize \textit{\textbf{II. Seed}}} \\
\midrule
\gls{rag}&\addseed\cite{lewis_retrieval-augmented_2020}&
Combines pre-trained parametric memory (a seq2seq model) with non-parametric memory (a dense vector index accessed via a neural retriever).
\\
\midrule
\multicolumn{3}{c}{\cellcolor{gray!20} \scriptsize \textit{\textbf{III. Forward Snowballing}}} \\
\midrule
--- 
& 
\addforward\cite{gao_retrieval-augmented_2023} & 
A comprehensive review of the evolution of \gls{rag} paradigms from Naive to Modular.
\\
Agentic \gls{rag} & 
\addforward\cite{singh_agentic_2025} & 
A comprehensive analytical survey of Agentic \gls{rag}, introducing a principled taxonomy that details how embedding autonomous agents into \gls{rag} pipelines overcomes static workflow limitations through dynamic retrieval, multi-step reasoning, and adaptive collaboration.
\\
\gls{selfrag} & 
\addforward\cite{asai_self-rag_2023}
& 
Trains a language model to adaptively retrieve information on-demand and use special reflection tokens to evaluate both the retrieved passages and its own generations, improving factuality and inference-time controllability.
\\
\gls{graphrag} & \addforward\cite{edge_local_2025} & 
A graph-based \gls{rag} framework that addresses the limitations of standard \gls{rag} on global, corpus-wide queries by constructing an entity knowledge graph and pregenerating community summaries that are aggregated to synthesize comprehensive responses across massive text collections.
\\
\gls{hipporag} & \addforward\cite{gutierrez_hipporag_2024} & 
Inspired by human long-term memory that orchestrates \glspl{llm} and knowledge graphs to enable efficient, single-step knowledge integration across new experiences.
\\
\gls{raft} & \addforward\cite{zhang_raft_2024} & 
A post-training recipe that enhances in-domain, open-book question answering by teaching models to ignore irrelevant distractor documents and cite verbatim passages from relevant texts using chain-of-thought reasoning.
\\
\bottomrule
\end{tabularx}
}
\end{table}
}

{\scriptsize
\begin{table}[htbp]
\caption{Knowledge editing snowballing results.}
\label{tab:editing-chain}
\adjustbox{width=.9\linewidth
}{
\begin{tabularx}{\linewidth}{p{0.1\linewidth} X p{0.78\linewidth}}
\toprule
\multicolumn{1}{c}{\textbf{Method}} & \multicolumn{1}{c}{\textbf{Ref.}} & \multicolumn{1}{c}{\textbf{Key Development}} 
\\
\midrule
\multicolumn{3}{c}{\cellcolor{gray!20} \scriptsize \textit{\textbf{I. Backward Snowballing}}} \\
\midrule
Knowledge Neurons &\addback\cite{dai_knowledge_2022}&
Identifies the specific neurons responsible for storing and expressing factual knowledge within pretrained Transformers, enabling direct knowledge editing without model \gls{ft}.
\\
KnowledgeEditor &\addback\cite{de_cao_editing_2021}&
Corrects specific incorrect or obsolete factual knowledge in language models without expensive retraining by utilizing a hyper-network trained with constrained optimization to predict precise weight updates that generalize to paraphrases while leaving unrelated knowledge intact.
\\
---&\addback\cite{geva_transformer_2021}&
Reveals that feed-forward layers in transformer language models operate as interpretable key-value memories, where keys detect specific textual patterns and values induce corresponding output vocabulary distributions to predict subsequent tokens.
\\
\midrule
\multicolumn{3}{c}{\cellcolor{gray!20} \scriptsize \textit{\textbf{II. Seed}}} \\
\midrule
\gls{rome}&\addseed\cite{meng_locating_2022}&
Uses causal interventions to locate factual associations within the mid-layer feed-forward modules of autoregressive transformers and directly updates these specific weights to alter factual predictions without sacrificing specificity or generalization.\\
\midrule
\multicolumn{3}{c}{\cellcolor{gray!20} \scriptsize \textit{\textbf{III. Forward Snowballing}}} \\
\midrule
\gls{memit} & 
\addforward\cite{meng_mass-editing_2023} & 
A scalable method for directly updating \glspl{llm} that enables the simultaneous insertion of thousands of memories.
\\
\gls{mend} & 
\addforward\cite{mitchell_fast_2022}
& 
An efficient post-hoc editing approach that utilizes small auxiliary networks and low-rank gradient decomposition to enable fast, targeted corrections to massive pre-trained models using just a single input-output pair.
\\
\gls{serac} & 
\addforward\cite{mitchell_memory-based_2022}
& 
Stores edits in an explicit memory and reasons over them to dynamically modulate the base model's predictions.
\\
\gls{pmet} & \addforward\cite{li_pmet_2024} & 
Simultaneously optimizes Transformer component hidden states but selectively updates only the Feed-Forward Network weights, thereby preserves the general knowledge extraction patterns inherently encoded within the Multi-Head Self-Attention mechanism.
\\
WISE 
& \addforward\cite{wang_wise_2024} & 
A dual parametric memory scheme that resolves the impossible triangle of lifelong model editing by employing a router to direct queries between a main pre-trained memory and a side memory, where continual edits are seamlessly integrated via a conflict-free knowledge-sharding mechanism.
\\
\gls{make} & \addforward\cite{park_make_2025} & 
Leverages a language model's own internal memory to recall and update indirectly associated knowledge, ensuring consistent knowledge transfer without the need for external databases.
\\
\bottomrule
\end{tabularx}
}
\end{table}
}

{\scriptsize
\begin{table}[htbp]
\caption{Machine unlearning snowballing results.}
\label{tab:unlearning-chain}
\adjustbox{width=.9\linewidth
}{
\begin{tabularx}{\linewidth}{p{0.1\linewidth} X p{0.78\linewidth}}
\toprule
\multicolumn{1}{c}{\textbf{Method}} & \multicolumn{1}{c}{\textbf{Ref.}} & \multicolumn{1}{c}{\textbf{Key Development}} 
\\
\midrule
\multicolumn{3}{c}{\cellcolor{gray!20} \scriptsize \textit{\textbf{I. Backward Snowballing}}} \\
\midrule
Machine Unlearning &\addback\cite{ginart_making_2019}&
To comply with privacy mandates like the ``Right to Be Forgotten'', it proposes algorithmic principles and specific $k$-means algorithms that efficiently delete individual data points without the prohibitive computational cost of retraining models from scratch.
\\
\gls{sisa}&\addback\cite{bourtoule_machine_2021}&
Accelerates machine unlearning by strategically partitioning data and isolating its influence during training, significantly reducing the computational overhead required to remove a user's data from deep neural networks compared to retraining from scratch.
\\
\midrule
\multicolumn{3}{c}{\cellcolor{gray!20} \scriptsize \textit{\textbf{II. Seed}}} \\
\midrule
Machine Unlearning &\addseed\cite{cao_towards_2015}&
Enables systems to rapidly and completely forget specific training data and its lineage by transforming learning algorithms into a summation form, achieving data removal asymptotically faster than retraining from scratch.
\\
\midrule
\multicolumn{3}{c}{\cellcolor{gray!20} \scriptsize \textit{\textbf{III. Forward Snowballing}}} \\
\midrule
\gls{ga} & \addforward\cite{yao_machine_2024} & 
Maximize loss on forget set (reverse learning signal)
Demonstrates that integrating \gls{ga} with \gls{gd} on in-distribution data provides a robust, highly efficient mechanism to enforce the "right to be forgotten" without the computational burden of retraining.
\\
Task Vec.\ Negation & 
\addforward
\cite{ilharco_editing_2023} & 
A paradigm for steering neural network behavior by representing tasks as vectors in weight space that can be added, negated, or combined to efficiently edit models and improve multi-task performance.
\\
\gls{rmu} & 
\addforward\cite{li_wmdp_2024} & 
Introduces the \gls{wmdp} benchmark to measure hazardous knowledge in \glspl{llm}, alongside \gls{rmu}, an unlearning method that controls model representations to selectively erase dangerous capabilities while preserving general domain proficiency.
\\
Adaptive \gls{rmu} & 
\addforward\cite{huu-tien_effects_2025} & 
Provides a theoretical framework explaining how \gls{rmu} successfully erases knowledge by reducing token confidence, and introduces Adaptive \gls{rmu}, an enhancement that overcomes standard \gls{rmu}'s limitations to enable highly effective unlearning across all network layers.
\\
\glspl{sae} & 
\addforward\cite{farrell_applying_2024} & 
Demonstrates that negatively scaling specific interpretable features can selectively erase targeted knowledge.
\\
\gls{loku}, \gls{fila}  & 
\addforward\cite{cha_towards_2024} 
& 
Introduces \gls{loku}, a framework that leverages an Inverted Hinge Loss to suppress sensitive tokens while maintaining linguistic fluency and utilizes \gls{fila} to precisely target unlearning without sacrificing a model's broader reasoning and generative capabilities.
\\
\gls{loka} & 
\addforward\cite{zhang_loka_2026} & 
A conflict-aware framework for simultaneous knowledge unlearning and updating that utilizes an adaptive, multi-unit memory architecture and a learning-based router to dynamically integrate new information while effectively managing task conflicts.
\\
\bottomrule
\end{tabularx}
}
\end{table}
}

{\scriptsize
\begin{table}[htbp]
\caption{\gls{repe} / Activation steering snowballing results.}
\label{tab:steering-chain}
\adjustbox{width=.9\linewidth
}{
\begin{tabularx}{\linewidth}{p{0.1\linewidth} X p{0.78\linewidth}}
\toprule
\multicolumn{1}{c}{\textbf{Method}} & \multicolumn{1}{c}{\textbf{Ref.}} & \multicolumn{1}{c}{\textbf{Key Development}}
\\
\midrule
\multicolumn{3}{c}{\cellcolor{gray!20} \scriptsize \textit{\textbf{I. Backward Snowballing}}} \\
\midrule
---&\addback\cite{mikolov_linguistic_2013}&
Demonstrates that vector-space word representations learned by continuous language models implicitly capture deep syntactic and semantic regularities, enabling complex linguistic relationships to be accurately modeled and solved through simple algebraic vector operations—famously illustrated by "King - Man + Woman = Queen."
\\
\gls{tcav}&\addback\cite{kim_interpretability_2018}&
Translate a neural network's internal representations into high-level, human-interpretable concepts and use directional derivatives to quantify their exact importance to classification decisions.
\\
\gls{iti}&\addback\cite{li_inference-time_2023}&
Enhances the truthfulness of \glspl{llm} by shifting internal activations within specific attention heads during inference, offering a computationally inexpensive and data-efficient method to surface internally represented truths.
\\
\midrule
\multicolumn{3}{c}{\cellcolor{gray!20} \scriptsize \textit{\textbf{II. Seed}}} \\
\midrule
\gls{repe} &\addseed\cite{zou_representation_2023}&
A neuroscience-inspired, top-down approach to AI transparency that focuses on monitoring and manipulating population-level representations rather than individual neurons to effectively control safety-relevant behaviors like honesty and harmlessness in \glspl{llm}.
\\
\gls{actadd}&\addseed\cite{turner_activation_2023}&
Steers language model outputs by directly modifying intermediate activations during inference using simple steering vectors derived from prompt pairs, enabling real-time control over high-level properties like sentiment and toxicity without the need for \gls{ft}.
\\
\midrule
\multicolumn{3}{c}{\cellcolor{gray!20} \scriptsize \textit{\textbf{III. Forward Snowballing}}} \\
\midrule
\gls{caa} &  
\addforward\cite{rimsky_steering_2024} & 
Steers the behavior of \glspl{llm} at inference time by modifying their internal hidden state activations.
\\
\glspl{fv} & 
\addforward\cite{todd_function_2023} & 
Demonstrates that \glspl{llm} encode specific tasks as compact, extractable \glspl{fv} within their attention heads, which can be independently manipulated or combined to causally steer the model's behavior across novel contexts.
\\
Conceptor & 
\addforward\cite{postmus_steering_2024} & 
Introduces ``conceptors'' for activation engineering
inference-time steering of \glspl{llm} and the use of Boolean operations to effectively combine complex control goals.
\\
Circuit Breakers & 
\addforward\cite{zou_improving_2024} & 
Introduces ``circuit breakers,'' an alignment technique
that directly disrupts the internal model representations responsible for harmful outputs, creating a robust, attack-agnostic defense for both text and multimodal AI systems without compromising their core utility.
\\
\bottomrule
\end{tabularx}
}
\end{table}
}

{\scriptsize
\begin{table}[htbp]
\caption{Data augmentation and synthetic data generation snowballing results.}
\label{tab:augment-chain}
\adjustbox{width=.9\linewidth
}{
\begin{tabularx}{\linewidth}{p{0.1\linewidth} X p{0.78\linewidth}}
\toprule
\multicolumn{1}{c}{\textbf{Method}} & \multicolumn{1}{c}{\textbf{Ref.}} & \multicolumn{1}{c}{\textbf{Key Development}}
\\
\midrule
\multicolumn{3}{c}{\cellcolor{gray!20} \scriptsize \textit{\textbf{I. Backward Snowballing}}} \\
\midrule
---&\addback\cite{sennrich_improving_2016}&
Introduces a back-translation strategy for \gls{nmt} that pairs target-side monolingual data with automatic synthetic translations to create additional parallel training data.
\\
\gls{eda}&\addback\cite{wei_eda_2019}&
A technique consisting of four simple operations (synonym replacement, random insertion, random swap, and random deletion) that boosts text classification performance and data efficiency, particularly on smaller datasets.
\\
AutoAugment &\addback\cite{cubuk_autoaugment_2019}&
Automates the creation of data augmentation strategies by using a search algorithm to discover optimal, highly transferable policies composed of varied image-processing sub-policies that maximize a neural network's validation accuracy.
\\
\midrule
\multicolumn{3}{c}{\cellcolor{gray!20} \scriptsize \textit{\textbf{II. Seed}}} \\
\midrule
---&\addseed\cite{wang_survey_2024}&
Provides a survey of data synthesis and augmentation techniques across the entire lifecycle of \glspl{llm}.
\\
---&\addseed\cite{zhou_survey_2024}&
This survey systematically categorizes and evaluates large model-driven data augmentation techniques across text, image, and paired data, highlighting how leveraging these models to generate high-quality training data addresses the critical challenge of impending data scarcity across multiple domains.
\\
\midrule
\multicolumn{3}{c}{\cellcolor{gray!20} \scriptsize \textit{\textbf{III. Forward Snowballing}}} \\
\midrule
---&\addforward\cite{long_llms-driven_2024}&
Systematizes the workflow of \gls{llm}-driven synthetic data generation, curation, and evaluation.
\\
\bottomrule
\end{tabularx}
}
\end{table}
}
\FloatBarrier

\subsection{Grey Literature Sources}\label{sec:grey}

\subsubsection{Rationale for Grey Literature Inclusion}\label{sec:grey-rationale}

The inclusion of grey literature in this \gls{mlr} is warranted on three grounds. First, AI adaptation is heavily driven by industry practice: emergent techniques frequently appear in framework documentation, developer blogs, and technical reports months or years before formal academic publication. Restricting the review to peer-reviewed sources alone would omit rapidly evolving concepts. Second, several key techniques in the framework (notably \gls{ce}, model merging via \gls{slerp}, and the \gls{grpo} family algorithms) were first formalized in practitioner documentation and only subsequently (if at all) received academic treatment. Third, framework documentation (Hugging Face \gls{peft}, MergeKit, \gls{trl}, ms-swift, EasyEdit) constitutes the de facto standard reference for implementation details and method comparisons that academic papers often omit or simplify.

Grey literature sources were included only if they met at least one of the following criteria: (a)~they constituted official framework documentation from a major AI organization; (b)~they introduced or formalized a technique or term not yet covered in peer-reviewed literature; or (c)~they provided implementation-level detail necessary to distinguish technique variants that appear identical in academic descriptions. Promotional material, opinion pieces, and sources without verifiable technical content were excluded.

\subsubsection{Grey Literature Inventory}\label{sec:grey-inventory}

The following is a complete enumeration of the \nbGreyLit grey literature sources retained in the final corpus, organized by category:

\textbf{Framework documentation (5):} Hugging Face \gls{peft} (v0.18.0); MergeKit (Arcee AI); \gls{trl} (Hugging Face); ms-swift (ModelScope); EasyEdit (Zhejiang University, ACL 2024 System Demo).

\textbf{\gls{ce} (6):} Anthropic engineering blog; 
Prompt Engineering Guide; LlamaIndex guide; Weaviate guide; Philipp Schmid blog; Kubiya 2025 guide. See Section~\ref{sec:grey-context-eng}.

\textbf{Industry technical reports (5):} 
A. Karpathy
(\gls{ce} definition; \gls{llm}-as-CPU mental model);
Gartner \gls{ce} tracking; OpenAI system card documentation (\gls{rlhf} pipeline details); DeepSeek-R1 technical report appendices (training pipeline implementation details beyond the main paper).

\textbf{Developer blogs and analyses (4):} T. L\"{u}tke (Shopify) \gls{ce} remarks; D. Horthy (HumanLayer) \gls{ce} remarks; \gls{slerp} implementation guides (multiple community sources, consolidated); MergeKit community benchmarking reports.

\subsubsection{\gls{ce} (Grey Literature)}\label{sec:grey-context-eng}

\textbf{Definition:} 
A. Karpathy:
``The delicate art and science of filling the context window with just the right information for the next step'' \cite{andrej_karpathy_karpathy_1_2025}.

\textbf{Distinction from \gls{pe}:} \gls{pe} concerns how to phrase the question (static, single instruction). \gls{ce} concerns designing the system architecture that feeds the \gls{llm} the right information at the right time (dynamic system, runtime assembly)~\cite{schmid_new_2025}.  

\textbf{Components:} System prompt, user input, short-term memory (chat history), long-term memory (vector store retrieval), information from knowledge bases (\gls{rag}, MCP, API calls), tool definitions/responses, structured outputs, global state/workflow context~\cite{peGuide_CE_2025}.

\textbf{Inclusion justification:} Despite the absence of peer-reviewed formalization, \gls{ce} is included in the framework (Row~\ref{tech:ce}) because: (a)~it has achieved widespread practitioner adoption, as evidenced by Gartner's tracking of the term~\cite{noauthor_context_nodate,noauthor_lead_nodate}; (b)~it occupies a distinct six-dimensional profile (\tax{1}{ctx}, \taxNo{3}{extUser}, \tax{4}{verPersistent}, \tax{5}{extContxt}) that is not captured by any other row; and (c)~it represents the operational pattern underlying agentic AI systems~\cite{noauthor_effective_nodate,noauthor_context_nodate-2,noauthor_context_2025}.
Excluding it would leave a significant gap between the taxonomy and current practice.

\subsubsection{Hugging Face \gls{peft} Library}
Supported methods: \gls{lora}, Prefix Tuning, Prompt Tuning, P-Tuning, \gls{adalora}, LoHa, LoKr, AdaptionPrompt, \gls{ia3}, \gls{dora}, \gls{vera}, \gls{cpt_hf}, CorDA, BitFit, Hotswapping (15+ methods)~\cite{noauthor_peft_nodate}. 

\subsubsection{MergeKit (Arcee AI)}
Supported methods: Task Arithmetic, \gls{ties}, \gls{dare}, \gls{slerp}, Model Soups, linear interpolation~\cite{noauthor_arcee-aimergekit_2026}. 

\subsubsection{\gls{trl} Library (Hugging Face)}
Supported methods: \gls{sft}, \gls{dpo}, \gls{kto}, Reward Modeling, \gls{cpo}, \gls{simpo}, \gls{orpo}, \gls{ppo}, REINFORCE~\cite{noauthor_trl_nodate}.

\subsubsection{ms-swift (ModelScope)}
Supported \gls{rlvr} algorithms: \gls{grpo}, \gls{dapo}, \gls{gspo}, \gls{sapo}, \gls{cispo}, CHORD, \gls{rloo}, Reinforce++. Also supports: \gls{dpo}, \gls{kto}, RM, \gls{cpo}, \gls{simpo}, \gls{orpo}~\cite{noauthor_modelscopems-swift_2026}.

\subsubsection{EasyEdit (Zhejiang University)}
Supported methods: \gls{rome}, \gls{memit}, \gls{mend}, \gls{serac}, \gls{pmet}, GRACE, WISE, EMMET, InstructEdit, MELO~\cite{noauthor_zjunlpeasyedit_2026}.

\subsection{Snowball Saturation and Seed-Sensitivity Simulation}
To evaluate the robustness of the purposive snowball sampling methodology and quantify the structural reliance on any single seed paper, we conducted a \gls{mc} sensitivity analysis (5,000 replicates). It is important to note that this simulation models the opportunity structure of the literature network rather than strictly reconstructing the historical citation log. It tests whether the \total{seedcount} selected seeds were mathematically sufficient to reach the final retained corpus (\the\numexpr\totvalue{backcount}+\totvalue{seedcount}+\totvalue{forwardcount}\relax~chain entries across the \NbChains chains plus \nbGreyLit grey-literature sources; Section \ref{sec:prisma})

\subsubsection{Methodological Assumptions}
Reachability was modeled using fixed category-level affinity probabilities rather than deterministic links:
\begin{itemize}
    \item \textbf{Own-chain affinity} (the probability of discovering an item within the same taxonomy category as the seed) was set to 0.90.
\item \textbf{Related-chain affinity} (cross-discovery between heavily overlapping domains, such as RLVR and Alignment, or Merging and Unlearning via task vectors) was parameterized between 0.20 and 0.45, mirroring the cross-chain duplicates observed during the actual review.
\item \textbf{Unrelated-chain affinity} was set to a baseline noise level of 0.03.
\item \textbf{Grey literature accessibility} was modeled at 0.18 from modern, LLM-era "practitioner-adjacent" seeds, and 0.03 from older, traditional academic seeds.
\end{itemize}
Because the simulation's item universe is the retained corpus, it
characterizes internal reachability and redundancy among retrieved sources; it does not estimate coverage of literature that the search did not retrieve. Affinity parameters represent modelled discovery opportunity rather than measured citation probabilities, and results should be read as a structural sensitivity probe rather than as a recall estimate.

\subsubsection{Simulation Results}
The simulation results, visualized in Figure~\ref{fig:SnowballSensitivity}, are consistent with the structural stability of the chosen seed set and characterize the reachability structure of the seed network:

\begin{figure}[htbp]
    \centering
    \includegraphics[width=\linewidth]{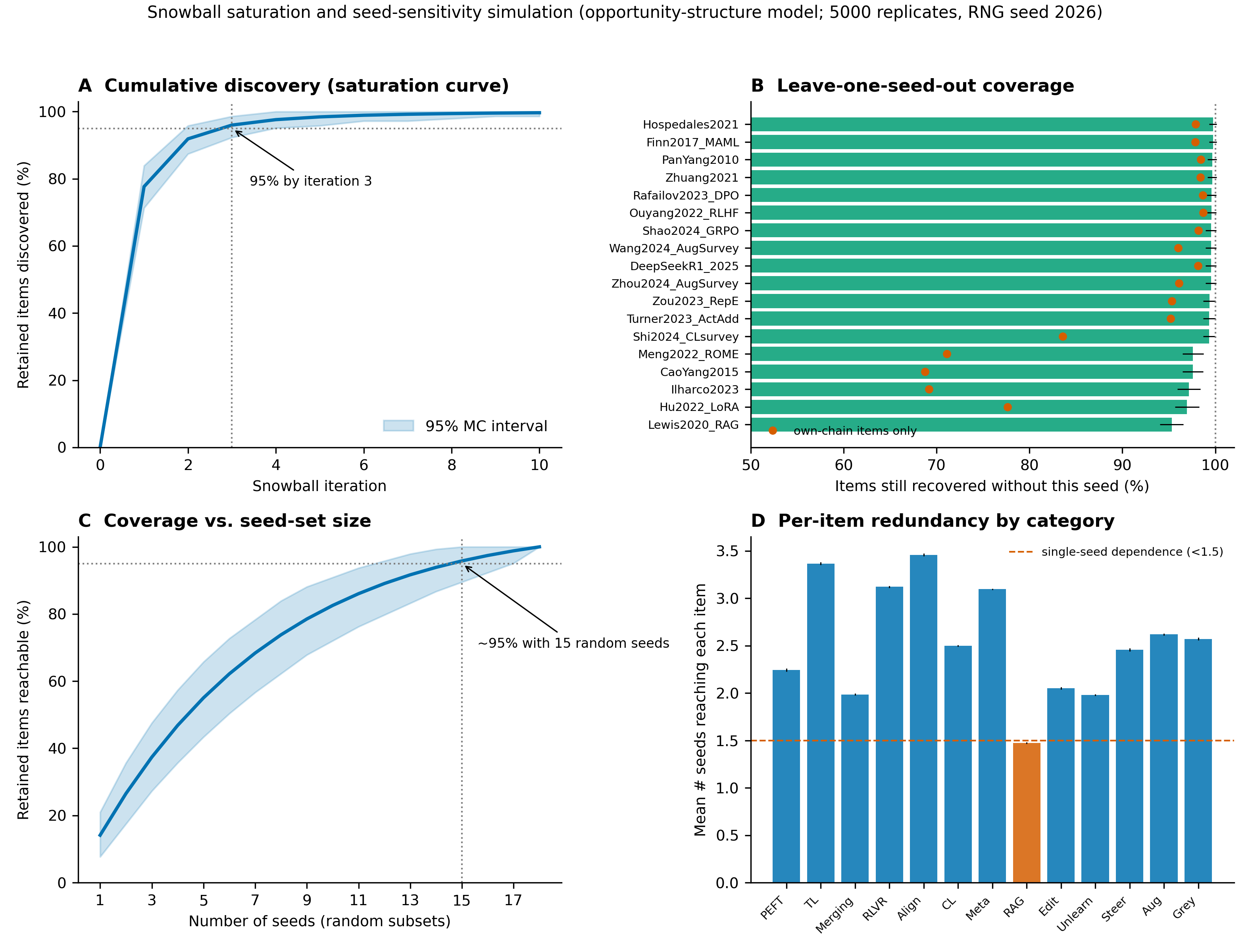}
    \caption{\textbf{Snowball sampling saturation and seed-sensitivity simulation.} 
     \gls{mc} analysis (5,000 replicates) evaluating the structural robustness of the \total{seedcount}-seed literature network. (A) Cumulative fraction of retained items discovered per snowball iteration (shading indicates 95\% confidence intervals). Mean discovery reaches 95\% saturation by iteration 3. (B) \gls{loso} coverage. Bars indicate the fraction of total items still reachable when a specific seed is removed. Orange markers isolate coverage strictly to the removed seed’s own chain, highlighting concentrated dependencies (e.g., \gls{rag} category retention drops to ~40\% without Lewis et al.). (C) Expected coverage as a function of seed-set size, demonstrating diminishing returns beyond approximately 15 seeds. (D) Mean per-item redundancy (number of unique seeds capable of reaching each item) categorized by taxonomy chain. The dashed line denotes the single-seed-dependence threshold (<1.5), which isolated categories like \gls{rag} fall below. (Note: Reachability is modeled on category-level affinity probabilities representing discovery opportunity, rather than strict historical citation traces).
    }
    \label{fig:SnowballSensitivity}
\end{figure}

\begin{itemize}
    \item Rapid Saturation (Panel A): The cumulative fraction of retained items discovered per snowball iteration saturates quickly. The mean discovery rate reaches 95\% expected coverage by the third snowball iteration, consistent with the depth of the manual search.
\item \gls{loso} Robustness (Panel B): To test vulnerability to initial seed selection, we calculated the fraction of items still reachable when each seed is systematically removed. Overall corpus coverage remains highly robust, staying above 95\% under any single seed removal. However, isolating the coverage to the removed seed's own chain reveals concentrated dependencies. For example, removing Lewis et al. drops \gls{rag}-specific coverage to approximately 40\%, demonstrating why certain foundational papers were strictly mandatory inclusions.
\item Diminishing Returns (Panel C): Expected coverage plotted against seed-set size demonstrates severe diminishing returns beyond approximately 15 random seeds. The purposive selection of \total{seedcount} seeds safely captures the asymptote of this discovery curve.
\item Network Redundancy (Panel D): Mean per-item redundancy tracks how many unique seeds can eventually reach a given item. While most taxonomy categories enjoy robust, multi-seed discovery paths, structurally isolated categories (such as \gls{rag}) fall below the 1.5 single-seed-dependence threshold, further validating the necessity of a highly purposive, multi-chain seed strategy over a single, centralized query.
\end{itemize}

\section{Detailed Dimensional Category Definitions}
\label{sec:appendix_definitions}

This section provides the granular definitions for the categories of our proposed taxonomy. 
By detailing the \dMechanism, \dGoal, \dData, \dPersistence, and \dScope, these tables allow for a clearer, multi-dimensional view of the current adaptation landscape.

{\scriptsize
\begin{table}[htbp]
\centering
\renewcommand{\tabularxcolumn}[1]{p{#1}}
\caption{\dimens{D}{1} (Mechanism of Adaptation) Categories}
\label{tab:D1_detailed}
\adjustbox{width=.9\linewidth
}{
\begin{tabularx}{\linewidth}{p{.1\linewidth} X}
\toprule
\multicolumn{1}{c}{\textbf{Category}} &
\multicolumn{1}{c}{\textbf{Detailed Definition}} 
\\
\midrule
\multicolumn{2}{c}{\cellcolor{gray!20} \scriptsize \textit{\textbf{I. 
Gradient-Based / Compute-Intensive}
}} \\
\midrule
\makebox[0pt][l]{\hypertarget{dim:paramUpd}{}} 
\paramUpd
& 
A gradient-based optimization procedure that computes loss-driven parameter gradients—using a task-specific or dataset-specific objective—and applies them to alter the stored weight tensors of one or more model components. The mechanism can operate across the full parameter space simultaneously (full \gls{ft}), or be constrained to a designated subset of layers, adapter matrices, or task-specific heads (partial and parameter-efficient variants); in all cases it consumes forward and backward passes—and therefore training-time floating-point operations—to modify the model's learned representations. No architectural constraints are assumed: the defining property is that \gls{gd} acts directly on stored parameters.
 \\
  
\makebox[0pt][l]{\hypertarget{dim:CrossModelTransf}{}}
\CrossModelTransf 
& A training procedure in which a new, independently deployable student model is optimized to replicate the output distributions, intermediate representations, or decision boundaries of a pre-existing teacher model. Rather than updating the teacher's parameters, the mechanism generates a derived artifact—the student—by minimizing a surrogate objective (e.g., \gls{kl} divergence over soft logits, feature-map matching, or contrastive alignment) that transfers the teacher's generalization surface without requiring access to the original training data. The student and teacher are structurally distinct deployable artifacts.
 \\
\midrule
\multicolumn{2}{c}{\textit{\cellcolor{gray!20} \scriptsize \textbf{II. 
Gradient-Free / Weight Manipulation}
}} \\
\midrule

\makebox[0pt][l]{\hypertarget{dim:WeightSpaceComp}{}}
\WeightSpaceComp 
& A gradient-free algebraic operation applied directly to the parameter vectors of one or more pre-trained model checkpoints—such as task-vector addition, negation, 
or weight averaging—that derives a new behavioral configuration by combining or transforming weight-space representations without executing any training loop on task data. No forward passes over labeled examples and no loss function are required; the mechanism operates entirely in parameter space.  \\

\makebox[0pt][l]{\hypertarget{dim:StructReduct}{}}
\StructReduct 
& A structural or numerical intervention that reduces a single existing model's parameter count, memory footprint, or arithmetic precision—through techniques such as magnitude pruning, structured sparsification, or weight quantization—producing a deployment-optimized derivative artifact. Unlike cross-model transfer, no independent teacher model is required; the compression operation modifies the model's representational capacity directly by eliminating or reducing its own parameter tensors.  \\

\makebox[0pt][l]{\hypertarget{dim:arch}{}}
\arch 
& 
A topological alteration of the model's computational graph that introduces new nodes, routing pathways, or expert sub-networks—such as dynamically initializing \gls{moe} routing modules, inserting gating components, or appending new expert layers—without modifying the pre-trained weights of existing base components. The mechanism expands the model's capacity or specialization range through structural graph surgery rather than parameter optimization on existing tensors. 
 \\
\midrule
\multicolumn{2}{c}{\cellcolor{gray!20} \textit{\scriptsize \textbf{III. 
Inference-Time / Zero-Footprint}
}} \\
\midrule

\makebox[0pt][l]{\hypertarget{dim:ctx}{}}
\ctx 
& An inference-time conditioning mechanism that prepends, appends, or structurally assembles task-relevant information—including natural-language instructions, retrieved documents, few-shot demonstrations, or structured tool outputs—into the model's input context window without modifying any stored parameter weights. Behavioral changes are achieved entirely through the model's existing attention and generation mechanisms acting on the enriched input; no gradient computation occurs, and the model's parameter state is identical before and after the inference session. \\
\makebox[0pt][l]{\hypertarget{dim:transAct}{}}
\transAct 
& 
A mechanism that adds learned steering vectors or direction-specifying perturbations to the intermediate activation representations of a model during the forward pass—typically targeting specific residual stream positions or attention head outputs—thereby redirecting the model's behavioral trajectory along a target direction without modifying any stored weight tensors. Because the intervention operates in activation space rather than parameter space, the model's checkpoint is entirely unaffected; the effect is confined to the forward pass and leaves no trace in the persistent parameter state.
\\

\makebox[0pt][l]{\hypertarget{dim:infTimeSearch}{}}
\infTimeSearch 
& An inference-time compute-allocation mechanism that scales response quality by executing multiple internal generation processes
without modifying model parameters, injecting external retrieval context, or steering activations. The mechanism trades additional inference compute for improved output quality by exploring a broader portion of the model's generative distribution and selecting the highest-quality candidate according to a process-internal scoring criterion. Compute is consumed at inference time rather than training time. \\
\midrule
\multicolumn{2}{c}{\cellcolor{gray!20} \textit{\scriptsize \textbf{IV. 
Pipeline-Mediated}
}} \\
\midrule
\makebox[0pt][l]{\hypertarget{dim:manip}{}}
\manip 
& An indirect adaptation pathway in which the technique modifies the composition, ordering, labeling, or synthesis of training examples supplied to a downstream learning algorithm
without directly modifying any model parameter. Because the technique operates on the data pipeline rather than on the model artifact itself, the persistence and structural scope of the resulting adaptation are entirely determined by and inherited from the downstream training method it feeds.
\\
\bottomrule
\end{tabularx}
}
\end{table}
}

\clearpage

{\scriptsize
\begin{table}[htbp]
\centering
\renewcommand{\tabularxcolumn}[1]{p{#1}}
\caption{\dimens{D}{2} (Goal of Adaptation) Categories}
\label{tab:suppD2}
\adjustbox{
max width=.9\linewidth,
max totalheight=.99\textheight
}{
\begin{tabularx}{\linewidth}{X p{.85\linewidth} }
\toprule
\multicolumn{1}{c}{\textbf{Category}} &
\multicolumn{1}{c}{\textbf{Definition}} 
\\
\midrule
\multicolumn{2}{c}{\cellcolor{gray!20} \textit{\scriptsize \textbf{I. Capability} 
}} \\
\midrule
\makebox[0pt][l]{\hypertarget{dim:build}{}}
\build 
& The objective of constructing a model with generalizable predictive or generative capability from a large, representative training corpus, establishing the baseline parameter state and intended-use scope from which all post-training adaptation techniques depart. This goal is realized during initial training and serves as the foundational reference point against which all subsequent adaptation events are characterized.\\ 

\makebox[0pt][l]{\hypertarget{dim:task}{}}
\task 
& The objective of refining a general-purpose or pre-trained model to achieve high performance on a specific downstream task by optimizing on task-labeled data within an existing distribution, trading breadth of generalization for depth of domain-specific accuracy. The resulting model exhibits a narrowed operational envelope that is more reliable within the target task but may show reduced generalization to substantially out-of-distribution inputs. \\

\makebox[0pt][l]{\hypertarget{dim:capExt}{}}
\capExt 
& The objective of broadening a model's operational scope by adding fundamentally new structural capabilities or modalities—such as new languages, reasoning domains, subject areas, or input/output modalities—through additional training on broad corpora or through architectural expansion. Distinguished from task specialization by the breadth rather than depth of the intended behavioral change: the model acquires genuinely new operational dimensions rather than deepening performance within an existing one.\\
\makebox[0pt][l]{\hypertarget{dim:reas}{}}
\reas 
& 
The objective of enhancing a model's capacity for structured multi-step inference—including mathematical problem-solving, logical deduction, code generation, and systematic planning—by training against verifiable programmatic reward signals that assess correctness independently of human preference annotation. The mechanism typically involves \gls{rl} against outcome-based oracles such as unit-test executors, mathematical verifiers, or formal logic checkers, enabling the model to develop generalizable problem-solving strategies rather than pattern-matching memorized solution templates. Verification is accomplished through programmatic oracles, simulators, or iterative self-competition rather than human raters. \\ 
\midrule
\multicolumn{2}{c}{\cellcolor{gray!20} \textit{\scriptsize \textbf{II. Lifecycle} 
}} \\
\midrule
\makebox[0pt][l]{\hypertarget{dim:DriftRemed}{}}
\DriftRemed 
& The objective of restoring or sustaining a deployed model's predictive accuracy or output consistency when the operational data distribution diverges from the training distribution through covariate shift, concept drift, or prompt distribution change. Remediation targets preservation of the model's original task objective within its established operational envelope rather than reshaping that envelope for a new task; the behavioral goal is to re-align the model with its original specification, not to extend it. \\
\makebox[0pt][l]{\hypertarget{dim:ContAdapt}{}}
\ContAdapt 
& The objective of enabling a model to sequentially accumulate knowledge from new tasks or data streams without catastrophic forgetting of previously consolidated representations, by employing stability-plasticity trade-off mechanisms that constrain updates to preserve prior knowledge while accommodating new distributional information. The commitment to accumulation rather than replacement is the defining property distinguishing this goal from drift remediation.\\

\makebox[0pt][l]{\hypertarget{dim:domAdapt}{}}
\domAdapt$^\ast$ 
& The objective of reducing the feature-space mismatch between a source domain—on which the model was originally trained—and a target domain—on which it must perform—for the same or an analogous task class. The adaptation leverages unlabeled or lightly labeled target-domain data to align feature representations or adjust decision boundaries without requiring complete retraining from scratch.  
\\
\midrule
\multicolumn{2}{c}{\cellcolor{gray!20} \textit{\scriptsize \textbf{III. Trust}
}} \\
\midrule
\makebox[0pt][l]{\hypertarget{dim:algn}{}}
\algn 
& 
The objective of steering model outputs to conform with human values, behavioral preferences, helpfulness norms, and institutional guidelines, typically by optimizing against human preference signals or rule-based reward models that encode normative expectations about appropriate responses. Alignment interventions modify the model's behavioral disposition and output style rather than its factual knowledge base or core task capability.\\
\makebox[0pt][l]{\hypertarget{dim:safety}{}}
\safety 
& 
The objective of reducing the production of harmful, dangerous, or policy-violating outputs by training the model to recognize and refuse harmful requests, applying constitutional-principle enforcement, or filtering outputs through harm-category constraints. Safety-targeted interventions explicitly shape the model's refusal and harm-avoidance behavior as a primary optimization target, distinct from general alignment with preferences. \\
\makebox[0pt][l]{\hypertarget{dim:rob}{}}
\rob 
& The objective of hardening a model against adversarial perturbations, distributional noise, or out-of-distribution inputs by exposing it to worst-case or adversarially constructed training examples. The adaptation tightens the model's minimum guaranteed performance across the input space, reducing the gap between average-case and worst-case behavior.\\
\makebox[0pt][l]{\hypertarget{dim:explain}{}}
\explain 
& The objective of adapting a model or appending interpretive modules to expose internal representations, align reasoning traces with human-interpretable logic, or generate post-hoc rationales. 
\\

\makebox[0pt][l]{\hypertarget{dim:fairness}{}}
\fairness 
& The objective of reducing systematic performance or output disparities across demographic or otherwise protected groups by intervening on the model's learned representations or decision boundaries. 
\\

\makebox[0pt][l]{\hypertarget{dim:reliability}{}}
\reliability 
& The objective of aligning the model's expressed confidence scores with empirically observed outcome frequencies, ensuring that predictions assigned a given probability are correct at correspondingly consistent rates. Reliability interventions typically apply post-hoc calibration procedures
to the model's raw output logits, without modifying its internal learned representations. \\
\midrule
\multicolumn{2}{c}{\cellcolor{gray!20} \textit{\scriptsize \textbf{IV. Knowledge Operations} 
}} \\
\midrule
\makebox[0pt][l]{\hypertarget{dim:augK}{}}
\augK 
& The objective of injecting, refreshing, or correcting factual or domain-specific knowledge encoded in the model's parameters or made accessible at inference time, either by retrieving current information from an external corpus at inference time or by performing targeted parametric edits that overwrite specific outdated or incorrect factual associations within the model's weights.\\ 

\makebox[0pt][l]{\hypertarget{dim:rem}{}}
\rem 
& The objective of selectively erasing specific learned associations, factual claims, behavioral patterns, or capability traces from a trained model without requiring full retraining and without degrading unrelated model behaviors. 
\\
\midrule
\multicolumn{2}{c}{\cellcolor{gray!20} \textit{\scriptsize \textbf{V. Operational Constraints} 
}} \\
\midrule

\makebox[0pt][l]{\hypertarget{dim:compEff}{}}
\compEff 
& The objective of reducing a model's deployment cost, memory footprint, inference latency, or training compute requirements
while preserving task accuracy within an acceptable degradation tolerance. Efficiency interventions alter the model's deployment profile without changing its core task definition or behavioral envelope. \\

\makebox[0pt][l]{\hypertarget{dim:userPerson}{}}
\userPerson 
& The objective of tailoring model outputs, tone, or domain expertise to individual user preferences, histories, or contexts through discrete per-user parameters or tracking mechanisms
without requiring broad model-wide retraining that would affect all users uniformly. \\

\makebox[0pt][l]{\hypertarget{dim:priv}{}}
\priv 
& The objective of enabling collaborative model training or adaptation across distributed data holders while guaranteeing that sensitive raw data never leaves its originating environment. This is achieved by sharing only privacy-bounded gradient updates or model deltas rather than underlying data records, enabling a global model to benefit from distributed learning signals without centralizing sensitive information. \\

\makebox[0pt][l]{\hypertarget{dim:behavControl}{}}
\behavControl 
& The objective of constraining a model's output format, stylistic register, persona, or procedural behavior through input-context conditioning
without targeting the model's underlying knowledge representation or task capability. Behavioral control operates at the inference interface and produces no persistent parametric modification.\\

\bottomrule
\end{tabularx}
}
\end{table}
}

\clearpage

{\scriptsize 
\begin{table}[htbp]
\centering
\renewcommand{\tabularxcolumn}[1]{p{#1}}
\caption{\dimens{D}{3} (Data Requirement) Categories}
\label{tab:suppD3}
\adjustbox{width=.9\linewidth
, max totalheight=0.99\textheight
}{
\begin{tabularx}{\linewidth}{p{.1\linewidth}  X}
\toprule
\multicolumn{1}{c}{\textbf{Category}} &
\multicolumn{1}{c}{\textbf{Definition}} 
\\
\midrule
\multicolumn{2}{c}{\cellcolor{gray!20} \textit{\scriptsize \textbf{I. Human-Annotated Data} 
}}\\
\midrule
\makebox[0pt][l]{\hypertarget{dim:largeD}{}}
\largeD 
& A high-volume corpus of annotated input-output pairs
providing the statistical coverage necessary to update all model parameters reliably without overfitting. 
This data regime is the defining requirement for 
methods where the unrestricted optimization landscape requires broad supervision to constrain the high-dimensional parameter search space. 
\\ 

\makebox[0pt][l]{\hypertarget{dim:smallD}{}}
\smallD 
& A limited collection of annotated task-specific instances
sufficient for parameter-efficient adapter training or partial \gls{ft} but 
insufficient for full-parameter optimization without catastrophic forgetting or overfitting. The small scale constrains the effective update surface and necessitates parameter-efficient or regularized training strategies that limit the degrees of freedom available to the optimizer. \\

\makebox[0pt][l]{\hypertarget{dim:few}{}}
\few 
& A minimal set of 
input-output demonstration examples provided directly within the model's inference-time context window to condition its behavior without any gradient-based optimization. These examples function as behavioral exemplars that the model's in-context learning mechanism generalizes from within the active session, leaving the model's parameters entirely unmodified. \\

\makebox[0pt][l]{\hypertarget{dim:pref}{}}
\pref 
& A dataset of paired model outputs annotated by human raters to indicate relative behavioral preference according to quality, helpfulness, or safety criteria. Unlike task-labeled data that specifies a correct output, preference pairs encode comparative judgments between two candidate responses, making the comparative elicitation methodology—rather than absolute correctness—the defining characteristic of this data regime. \\ 

\makebox[0pt][l]{\hypertarget{dim:taskDist}{}}
\taskDist 
& A distribution over multiple heterogeneous tasks 
to expose the model to diverse adaptation episodes. The distributional coverage over task space—rather than depth of supervision on any single task—is the defining property; the model optimizes for initial parameters or learning procedures that generalize rapidly to novel tasks encountered at deployment. \\
\midrule
\multicolumn{2}{c}{\cellcolor{gray!20} \textit{\scriptsize \textbf{II. Synthetic / Programmatic Data} 
}}\\
\midrule

\makebox[0pt][l]{\hypertarget{dim:synthTeacher}{}}
\synthTeacher 
& Data generated by a more capable or aligned model acting as an annotator or oracle (teacher).
The generating model's biases, capability limitations, and output distribution are inherited by the student through this data pathway, making provenance characterization of the teacher a prerequisite for characterizing the data. \\ 

\makebox[0pt][l]{\hypertarget{dim:synthRule}{}}
\synthRule 
& Data generated iteratively based on an explicit set of textual guidelines or constitutional principles, in which a model critiques and revises its own outputs according to the specified normative rules. The constitutional text itself functions as the primary normative artifact encoding the behavioral constraints that propagate into the adapted model through the generated training data. \\

\makebox[0pt][l]{\hypertarget{dim:synthEnv}{}}
\synthEnv 
& 
Data generated through interaction with a physics engine, game simulator, formal environment, or through the model's own autoregressive self-generation acting as an opponent or world model. The simulation parameters, reward mechanics, and environmental transition dynamics define the statistical properties of the generated data; the learning signal is derived from outcome-based environmental feedback rather than human judgment. \\

\makebox[0pt][l]{\hypertarget{dim:instrResp}{}}
\instrResp 
& 
A heterogeneous collection of prompt-response pairs spanning many tasks and domains
used to train general instruction-following capability. The defining characteristic is corpus breadth across task types rather than depth within any single task; 
\\
\midrule
\multicolumn{2}{c}{\cellcolor{gray!20} \textit{\scriptsize \textbf{III. 
Live / Production Data} 
}}\\
\midrule
\makebox[0pt][l]{\hypertarget{dim:userInteract}{}}
\userInteract 
& Logged behavioral signals and implicit telemetry generated directly by end-users during active deployment
used to drive personalization or implicit feedback-based adaptation. This regime is operationally distinct from curated labeled datasets and comparative preference pairs in that it is generated passively during production use rather than collected through an explicit annotation pipeline.\\

\makebox[0pt][l]{\hypertarget{dim:tstOnly}{}}
\tstOnly 
& 
The live, unlabeled test instances encountered during active inference, used as the sole input signal for transient self-supervised adaptation without access to any external training data or ground-truth labels. The model adapts to each test batch or individual instance in a session-local manner; adaptation does not persist beyond inference completion.\\

\makebox[0pt][l]{\hypertarget{dim:seqIncD}{}}
\seqIncD$^\ast$ 
& A temporally ordered stream of training examples arriving in batches or continuously over the model's operational lifecycle, whose distributional properties evolve across time. Adaptation mechanisms operating on this data regime must integrate new information without discarding previously consolidated knowledge; 
\\ 
\midrule
\multicolumn{2}{c}{\cellcolor{gray!20} \textit{\scriptsize \textbf{IV. 
External / Targeted Data} 
}}\\
\midrule
\makebox[0pt][l]{\hypertarget{dim:ext}{}}
\ext 
& A curated external knowledge repository
queried at inference time to retrieve contextually relevant passages that augment model outputs. The corpus is a pre-existing, structurally static reference artifact that is independent of the active user session and of the model's internal parameter state. \\
\makebox[0pt][l]{\hypertarget{dim:forget}{}}
\forget 
& A precisely delineated set of training samples, factual associations, behavioral capability traces, or user data records whose influence is to be erased from model behavior. 
\\
\midrule
\multicolumn{2}{c}{\cellcolor{gray!20} \textit{\scriptsize \textbf{V. 
Minimal / Decentralized / Delegated Data} 
}}\\
\midrule
\makebox[0pt][l]{\hypertarget{dim:zeroSh}{}}
\zeroSh 
&  Natural-language task directives provided in the model's input context without any accompanying labeled examples or demonstrations, relying entirely on the model's pre-existing instruction-following and generalization capability. The data requirement is zero labeled instances: only the task specification itself is provided, and all behavioral conditioning is achieved through the model's previously learned representations.\\ 

\makebox[0pt][l]{\hypertarget{dim:wght}{}}
\wght 
& A data-free regime in which adaptation operates exclusively on the model's pre-trained parameter vectors through algebraic composition, interpolation, or negation—requiring no training examples, labeled or unlabeled, as input to the adaptation procedure. The entire adaptation is derived from arithmetic relationships among existing weight tensors. \\

\makebox[0pt][l]{\hypertarget{dim:decent}{}}
\decent 
& Private data distributed across multiple non-collocated clients or institutions, each processed locally without sharing raw records with a central coordinator. The statistical heterogeneity across shards—arising from non-independent and non-identically distributed local distributions—constitutes the primary technical challenge for any aggregation mechanism that attempts to synthesize a coherent global model from local signals. \\

\makebox[0pt][l]{\hypertarget{dim:pipeInherit}{}}
\pipeInherit 
& A delegation regime in which the effective data requirements of the technique are fully determined by and inherited from the downstream training method that the technique augments. The pipeline technique itself imposes no independent data regime; all data characterization is subsumed by the consuming training procedure. \\

\makebox[0pt][l]{\hypertarget{dim:unlabD}{}}
\unlabD
$^\dagger$ 
& Unannotated data used for self-supervised or unsupervised training objectives, continuous pre-training, or distribution sampling for post-training calibration or activation steering. The absence of ground-truth labels defines this regime: the learning signal is derived from the statistical structure of the raw data—through masked prediction, contrastive objectives, or density estimation—rather than from explicit human annotation.\\ 
\midrule
\multicolumn{2}{c}{\cellcolor{gray!20} \textit{\scriptsize \textbf{VI. Verifier-Defined Ground Truth} 
}}\\
\midrule

\makebox[0pt][l]{\hypertarget{dim:verify}{}}
\verify 
& Tasks for which correctness can be assessed by a deterministic programmatic verifier
without requiring human preference annotation. Ground truth is established through binary or structured programmatic evaluation rather than subjective human judgment, shifting the epistemological basis of the training signal from comparative human preference to objective computational verification. \\
\midrule
\multicolumn{2}{c}{\cellcolor{gray!20} \textit{\scriptsize \textbf{VII. Multi-Modal Paired Provenance} 
}}\\
\midrule
\makebox[0pt][l]{\hypertarget{dim:multiD}{}}
\multiD
& Aligned cross-modal paired samples
used to train cross-modal alignment or grounding objectives. Each training example carries independent provenance chains for each modality, and the cross-modal correspondence itself introduces additional alignment-specific data characteristics.  
\\

\bottomrule
\\
\end{tabularx}
}
\end{table}
}

\clearpage

{\footnotesize
\begin{table}[hbt!]
\centering
\renewcommand{\tabularxcolumn}[1]{p{#1}}
\caption{\dimens{D}{4} (Persistence) Categories}
\label{tab:suppd4_persistence}
\renewcommand{\arraystretch}{1.3}
\adjustbox{width=.9\linewidth
}{
\begin{tabularx}{\linewidth}{p{.1\linewidth}  X}
\toprule
\multicolumn{1}{c}{\textbf{Category}} &
\multicolumn{1}{c}{\textbf{Definition}} 
\\
\midrule
\multicolumn{2}{c}{\cellcolor{gray!20} \textit{\scriptsize \textbf{I. Pre-Specifiable Baselines} 
}}\\
\midrule

\makebox[0pt][l]{\hypertarget{dim:ScheduledPerm}{}}
\ScheduledPerm 
& Pre-planned, periodic parameter updates executed on a fixed temporal or data-volume cadence, resulting in a permanently updated model checkpoint that fully supersedes the prior parameter state. The modification is permanent at the model-parameter level—the updated weights replace the previous checkpoint and persist across all subsequent inference sessions—and its triggering conditions and approximate timing can be declared in advance. \\

\makebox[0pt][l]{\hypertarget{dim:boundedCumul}{}}
\boundedCumul 
& Incremental parameter accumulation across a finite, pre-enumerable set of new tasks or domains, in which each update extends the model's knowledge without overwriting prior representations. The finite, pre-specified character of the task sequence means the total knowledge accumulation horizon can, in principle, be declared in advance, making this the only cumulative persistence mode whose modification trajectory admits complete pre-specification.\\

\makebox[0pt][l]{\hypertarget{dim:verPersistent}{}}
\verPersistent 
& A context-only modification at the model-parameter level—such as a maintained retrieval corpus, a persistent steering-vector store, or a versioned production system prompt—that does not alter model weights but is maintained as a version-controlled, persistent system artifact whose update constitutes a system-level change event. While the model's internal parameter state is unchanged, the system's behavioral profile is durably modified by the maintained artifact across inference sessions.\\
\midrule
\multicolumn{2}{c}{\cellcolor{gray!20} \textit{\scriptsize \textbf{II. Event-Driven Unplanned Modification} 
}}\\
\midrule
\makebox[0pt][l]{\hypertarget{dim:AdhocPerm}{}}
\AdhocPerm 
& An unplanned, event-driven parameter update triggered by sudden data drift, an emergent vulnerability, a targeted knowledge correction, or an arbitrary developer decision rather than a pre-specified schedule. Like scheduled permanent updates, the change produces a new model checkpoint that permanently supersedes the prior parameter state; unlike them, the timing, trigger, and scope of the intervention cannot be declared in advance and are determined reactively.\\
\midrule
\multicolumn{2}{c}{\cellcolor{gray!20} \textit{\scriptsize \textbf{III. Continuous Surveillance Required} 
}}\\
\midrule
\makebox[0pt][l]{\hypertarget{dim:unboundedCumul}{}}
\unboundedCumul 
& Open-ended, continuous parameter accumulation over an indefinite or unknown horizon of streaming data, in which the model's behavioral state is never stable but continuously incorporates new representational commitments. The absence of a pre-specifiable termination boundary is the defining property: unlike bounded cumulative persistence, the total scope of adaptation cannot be declared in advance, and the model's behavioral state at any future checkpoint is structurally indeterminate at deployment time. \\ 
\midrule
\multicolumn{2}{c}{\cellcolor{gray!20} \textit{\scriptsize \textbf{IV. Session-Scoped} 
}}\\
\midrule
\makebox[0pt][l]{\hypertarget{dim:eph}{}}
\eph 
& A behavioral modification achieved entirely through input-context conditioning or internal generation processes at inference time, with the model's parameter state remaining identical before and after each session. The adaptation is local to the active inference request or session and clears completely upon session termination, leaving no trace in the model's stored parameter state. While ephemeral at the parameter level, the system-level behavioral effect may be functionally persistent if the conditioning mechanism is applied uniformly across all deployed inference sessions. \\

\makebox[0pt][l]{\hypertarget{dim:trans}{}}
\trans 
& 
A modification strictly confined to a temporary inference session that does not persist once that session concludes. The transient category encompasses two mechanistically distinct implementations: parameter-modifying variants (e.g., test-time adaptation), in which gradient updates are applied during inference and then explicitly rolled back upon session completion via weight restoration; and activation-modifying variants (e.g., activation steering), in which no weight write occurs and the effect naturally clears at session end.
\\
\midrule
\multicolumn{2}{c}{\cellcolor{gray!20} \textit{\scriptsize \textbf{V. Inherited Burden} 
}}\\
\midrule
\makebox[0pt][l]{\hypertarget{dim:pipeDep}{}}
\pipeDep 
& A persistence profile in which the temporal durability of the adaptation is not intrinsic to the technique itself but is entirely delegated to and inherited from the downstream 
learning method that the technique augments. Because the technique operates on the training data pipeline rather than directly on model parameters, its effective change-control classification is entirely contingent on the downstream training method it feeds; no independent persistence characterization applies to the pipeline technique in isolation.
\\
\bottomrule
\end{tabularx}
}
\end{table}
}

\clearpage

{\scriptsize
\begin{table}[hbt!]
\centering
\renewcommand{\tabularxcolumn}[1]{p{#1}}
\caption{\dimens{D}{5} (Scope) Categories}
\label{tab:suppd5_scope}
\renewcommand{\arraystretch}{1.3}
\adjustbox{width=.9\linewidth
}{
\begin{tabularx}{\textwidth}{p{.1\linewidth} X}
\toprule
\multicolumn{1}{c}{\textbf{Category}} &
\multicolumn{1}{c}{\textbf{Definition}} 
\\
\midrule
\multicolumn{2}{c}{\cellcolor{gray!20} \textit{\scriptsize \textbf{I. Full-System Validation}
}}\\
\midrule
\makebox[0pt][l]{\hypertarget{dim:whole}{}}
\whole 
& All layers and components of the model are modified simultaneously, producing the most comprehensive structural footprint in the taxonomy. Because the entire parameter space participates in the modification, no modular isolation boundary exists; any behavioral change introduced propagates across all task dimensions simultaneously, and the model's complete representational state is overwritten or jointly updated. Rollback requires full-checkpoint reversion rather than targeted component substitution. \\

\makebox[0pt][l]{\hypertarget{dim:fusedComp}{}}
\fusedComp 
& 
A derived model artifact produced by the algebraic combination
of two or more parent model checkpoints, resulting in a merged weight state in which the individual contributing models are no longer trivially separable. The fused artifact's behavior is not directly attributable to any single parent model, making its representational lineage structurally opaque relative to the source checkpoints.
 \\
\midrule
\multicolumn{2}{c}{\cellcolor{gray!20} \textit{\scriptsize \textbf{II. Targeted Component Validation}  
}}\\
\midrule
\makebox[0pt][l]{\hypertarget{dim:partl}{}}
\partl 
& 
Only a designated component or layer subset of the model is modified
while the remainder of the parameter space is frozen. The bounded modification surface restricts behavioral change to the updated component subset, enabling targeted validation and localized regression testing of only the modified layers; unmodified layers require no re-evaluation.
 \\ 

\makebox[0pt][l]{\hypertarget{dim:surrogate}{}}
\surrogate 
& A completely new, independently deployable model artifact trained to replicate a teacher model's outputs
or to provide a learned scoring signal as an intermediate training artifact.
Because it is a structurally new model produced as a byproduct of the adaptation procedure, it constitutes a distinct artifact from the adapted model itself and carries an independent representational provenance. \\

\makebox[0pt][l]{\hypertarget{dim:compressed}{}}
\compressed
& A structurally reduced, pruned, or lower-precision derivative of the original base model.
that occupies a strictly smaller representational capacity than the original while targeting preservation of the original model's functional behavior within an acceptable accuracy tolerance. The compressed artifact is evaluated as a modified derivative of the original model rather than an independently trained entity. \\
\midrule
\multicolumn{2}{c}{\cellcolor{gray!20} \textit{\scriptsize \textbf{III. Modular Isolation} 
}}\\
\midrule
\makebox[0pt][l]{\hypertarget{dim:modSwap}{}}
\modSwap 
& The adaptation is encapsulated as a decoupled, independently stored module
that can be attached to or detached from a frozen base model without altering the base model's parameter state in any way. The base model's existing representational state is completely preserved, and the modular component constitutes an independent artifact whose versioning is entirely distinct from the base checkpoint. Rollback is achieved by removing or swapping the module without any base-model reversion.
 \\
\midrule
\multicolumn{2}{c}{\cellcolor{gray!20} \textit{\scriptsize \textbf{IV. No Parametric Footprint}  
}}\\

\midrule
\makebox[0pt][l]{\hypertarget{dim:extContxt}{}}
\extContxt 
& 
No modification is made to any model parameter or internal activation state; behavioral conditioning occurs exclusively through external input-space manipulation, output candidate selection, or inference-time search over the model's generative distribution. The model's computational graph and all stored weight tensors remain completely unaltered. The adaptation footprint is confined to the interface between the model and its deployment environment, not to any internal model state.
 \\

\makebox[0pt][l]{\hypertarget{dim:actSpace}{}}
\actSpace 
& Direct modulation of the model's intermediate forward-pass representations
with zero modification to any underlying parameter weight tensors. While the model's internal activation state is transiently altered during the forward pass, no persistent change to any stored weight tensor occurs; the model's checkpoint is identical before and after the intervention. 
 \\
\midrule
\multicolumn{2}{c}{\cellcolor{gray!20} \textit{\scriptsize \textbf{V. Distributed Aggregation}
}}\\
\midrule
\makebox[0pt][l]{\hypertarget{dim:dIVdist}{}}
\dIVdist 
& The global model state is constituted by the aggregation of locally computed parameter updates from multiple decentralized clients, such that the adaptation signal is distributed across a federated system. The persistent artifact is the aggregated global model whose parameter state incorporates learning signals from all contributing local data shards without any raw data leaving the originating device. Validation focuses on the consistency of the aggregated global state rather than on any single local update. \\

\midrule
\multicolumn{2}{c}{\cellcolor{gray!20} \textit{\scriptsize \textbf{VI. Inherited Burden}
}}\\
\midrule

\makebox[0pt][l]{\hypertarget{dim:pipelineDepScope}{}}
\pipelineDepScope 
& The structural scope of the adaptation is not intrinsic to the technique itself but is entirely delegated to and inherited from the downstream training event that the technique feeds. Because the pipeline technique modifies the data distribution rather than the model's computational graph, it imposes no independent structural footprint; the component-level regression testing burden is entirely subsumed by the downstream training method that consumes the modified data. \\

\bottomrule
\end{tabularx}
}
\end{table}
\normalsize

\newpage

\section{Technique Definitions and Relationships}
\label{sec:TechDefinitions}

\subsection{Knowledge Transfer and Task Specialization (Rows~\ref{tech:fullFT}--\ref{tech:fsl} in Table~\ref{tab:centerpiece})}
\label{sec:TechTransfer}
\paragraph{\gls{tl} as Umbrella Paradigm.}
\label{sec:TLUmbrella}
\gls{tl} is not a single technique but the paradigm governing the reuse of knowledge acquired on a source task or domain to improve performance in a new target context~\cite{pan_survey_2010, zhuang_comprehensive_2021}. It functions as the conceptual container for \gls{ft}, \gls{da}, and \gls{fsl} (Rows~\ref{tech:fullFT}--\ref{tech:fsl}), which are its principal realizations. 
Because its \dMechanism and \dPersistence coordinates inherently vary by implementation, \gls{tl} is excluded from the centerpiece table and treated as an umbrella genus whose species are individually profiled. 
The hierarchy \gls{peft}~$\subset$~\gls{ft}~$\subset$~\gls{tl} 
encodes the key taxonomic relationship: 
each level of specialization tightens the constraints on \dScope and \dData simultaneously, and it is precisely these two dimensions that govern structural validation scope and data-provenance obligations in system-engineering documentation.

\textbf{Full \gls{ft} (Row~\ref{tech:fullFT})} is the maximally expressive parametric realization of \gls{tl}: \gls{gd} (\tax{1}{paramUpd}) applies to every model layer without exception, producing a \tax{5}{whole} footprint that carries the highest structural validation burden in the taxonomy. The unrestricted update surface structurally necessitates the \tax{3}{largeD} data regime—with all parameters free to drift, the optimization landscape has sufficient degrees of freedom to overfit or catastrophically corrupt prior knowledge unless constrained by large-scale supervision~\cite{kirkpatrick_overcoming_2017}. 
The \tax{2}{task} goal draws the sharpest coordinate boundary with Retraining (Row~\ref{tech:retraining}): both are mechanistically identical across \dMechanism, \dPersistence, and \dScope, yet retraining targets \tax{2}{DriftRemed} on a fixed behavioral envelope rather than reshaping that envelope for a new task. Compared to Partial \gls{ft} (Row~\ref{tech:partFT}), Full \gls{ft} sacrifices the reduced testing surface (\dScope contracts from \tax{5}{whole} to \tax{5}{partl}) in exchange for broader update expressiveness; compared to \gls{peft} (Row~\ref{tech:peft}), it eliminates the \tax{5}{modSwap} isolation boundary that permits component-level rollback, replacing it with whole-model entanglement. Because Full \gls{ft} modifies the entire parameter space with no modular isolation boundary, rollback requires complete checkpoint restoration rather than targeted component substitution, and the validation obligation extends to full-system behavioral regression testing across all previously certified behavioral boundaries before any re-deployment.

\textbf{Partial \gls{ft} (Row~\ref{tech:partFT})} targets \tax{2}{task} via the same \tax{1}{paramUpd} mechanism as Full \gls{ft} (Row~\ref{tech:fullFT}) but constrains updates to later, task-specific layers while freezing the early layers that encode broadly transferable representations~\cite{yosinski_how_2014}. 
This architectural discipline collapses the update surface from \tax{5}{whole} to \tax{5}{partl}—a coordinate divergence from Full \gls{ft} that carries both a data consequence and a validation consequence: the data requirement contracts to \tax{3}{smallD} (the frozen layers perform feature extraction, leaving only the task-specific layers requiring supervision), and the regression testing footprint is structurally isolated to the modified layer subset, enabling targeted component-level rollback. 
Compared to \gls{peft} (Row~\ref{tech:peft}), which introduces a fully decoupled \tax{5}{modSwap} adapter that can be removed without touching any base-model layer, Partial \gls{ft} embeds updates directly into the base model's later layers—rollback requires layer-level checkpoint restoration, not a clean module swap. 
The practical distinction matters for change-control documentation: a \gls{peft} event generates a standalone adapter artifact; a Partial \gls{ft} event generates a modified base-model checkpoint whose changed layers must be explicitly versioned. 
Partial \gls{ft} necessitates versioned checkpointing of the modified layer subset and localized parametric regression testing bounded strictly to those layers; because the frozen early layers remain mathematically unchanged, they require no re-validation.

\textbf{\gls{peft} (Row~\ref{tech:peft})} 
is taxonomically the most distinctive member of the \gls{ft} family, differentiated 
by its \taxNo{5}{partlMod} scope, which no full or partial \gls{ft} variant achieves. The defining operational property is that the base model's weights remain strictly frozen throughout training; all adaptation occurs within a small set of newly introduced parameters appended to or inserted into the frozen architecture~\cite{hu_lora_2022}. This produces the \tax{5}{modSwap} coordinate: the adapter is a structurally independent artifact that can be loaded, swapped, or removed without touching the base checkpoint. 
This \tax{5}{modSwap} property creates a structural bridge to Task Arithmetic and Model Merging (Row~\ref{tech:taskArith}), where adapter task-vectors can be algebraically combined. The data requirement contracts to \tax{3}{smallD} relative to Full \gls{ft} because the adapter's parameter count is orders of magnitude smaller than the base.
At the \gls{llm} and \gls{mllm} tiers, \gls{peft} simultaneously absorbs the distributional gap goal of \gls{da} (Row~\ref{tech:da}) when applied to domain-specific corpora, and its modular architecture makes it the preferred complement to alignment techniques (\gls{rlhf}, \gls{dpo}) in production stacks. Because a change-control event in a \gls{peft} deployment affects only the adapter artifact—not the frozen base—regression testing scope is bounded to the adapter module alone; the base model's existing validation record remains unmodified, enabling clean rollback by removing the adapter checkpoint without any base-model reversion.
While categorized as a single structural decision unit in the taxonomy, the \gls{peft} paradigm encompasses a wide array of architectural variants. These include quantization-aware methods (e.g., \gls{qlora}~\cite{dettmers_qlora_2023}), continuous prompt interventions (prefix-tuning, prompt-tuning, p-tuning), sparse bias updates (\gls{bitfit}~\cite{ben_zaken_bitfit_2022}), and attention-scaling (\gls{ia3}~\cite{liu_few-shot_2022}). Furthermore, dynamic adapter composition strategies, such as adapter fusion and \gls{lora} routing (e.g., X-LoRA~\cite{buehler_x-lora_2024}, LoraHub~\cite{huang_lorahub_2024}, S-LoRA~\cite{sheng_s-lora_2024}), functionally blur the boundary between modular \gls{peft} and \gls{moe} (Row~\ref{tech:moe}) by dynamically routing inputs to specialized adapter modules.

\textbf{\gls{reft} (Row~\ref{tech:reft})} presents an alternative to standard weight-based \gls{peft} by intervening on the internal activation streams of the model~\cite{wu_reft_2024}.
While it utilizes a \tax{1}{paramUpd}, it does not alter the base weights; instead, it trains lightweight, \tax{5}{modSwap} projection matrices (interventions) that alter the model's hidden representations during the forward pass. It targets \tax{2}{task} using \tax{3}{smallD} datasets. Because it introduces permanent, swappable components, it maintains \taxNo{4}{SchedAdhocPerm} persistence. It is distinct from Activation Steering (which calculates transient vectors) because \gls{reft} permanently trains its intervention matrices via \gls{gd}.

\textbf{\gls{da} (Row~\ref{tech:da})} 
addresses \tax{2}{domAdapt}—reducing the feature-space mismatch between source and target domains when labeled target data is unavailable or scarce.
Its \taxNo{3}{unlabSmallD} profile reflects the defining operational condition: the target domain lacks sufficient labeled examples to support task-specific \gls{ft},
The \tax{1}{paramUpd} mechanism operates by training a feature extractor alongside an explicit domain discriminator or statistical divergence minimizer, yielding a \tax{5}{partl} footprint (the alignment layers and feature extractor are updated; task heads may remain frozen)~\cite{ganin_domain-adversarial_2016}. The defining taxonomic boundary is \taxNo{6}{df}: at the \gls{llm} tier, the \gls{da} goal persists but the mechanism is superseded—\gls{cpt} (Row~\ref{tech:ssl}) achieves distributional alignment by continued self-supervised training on unlabeled target-domain corpora, and domain-specific \gls{peft} (Row~\ref{tech:peft}) injects target-domain signal via lightweight adapters.
System documentation for \gls{llm}-level \gls{da} must therefore specify the concrete \gls{cpt} or \gls{peft} pathway rather than invoking classical \gls{da} terminology. 
At the \taxNo{6}{df} tier, change-control documentation should specify which alignment layers were updated and which domain-discriminator architecture was used, as these define the precise structural boundary of the modification event; regression testing should verify that source-domain performance is preserved alongside target-domain improvement—necessitating a dual-distribution evaluation protocol as part of the validation record.

\textbf{\gls{fsl} (Row~\ref{tech:fsl})}  
is defined within this taxonomy exclusively as a \tax{1}{paramUpd} intervention under extreme data scarcity. Depending on the implementation, it relies on either 1–5 examples for metric-based support sets (\tax{3}{few}) or tens of examples for gradient-based updates (\tax{3}{smallD}), producing a persistent weight modification (\taxNo{4}{SchedAdhocPerm})~\cite{vinyals_matching_2016,snell_prototypical_2017}.
This strict boundary resolves the taxonomy's most consequential terminological ambiguity: colloquial ``few-shot'' usage conflates two mechanistically and operationally distinct operations. The parametric form
produces a versioned model artifact that has undergone a parameter-modifying training event and therefore demands formal change-control tracking and baseline regression testing. 
The non-parametric, prompting-based form
shares \tax{3}{few} but operates via \tax{1}{ctx}, producing only a \tax{4}{eph} behavioral shift that leaves the underlying model artifact unmodified; this is reassigned to \gls{icl} (Row~\ref{tech:icl}). The \dPersistence divergence is the taxonomic anchor: \gls{fsl} creates a new versioned artifact requiring software change-control; \gls{icl} creates no artifact whatsoever. \gls{fsl}'s \tax{6}{df} ceiling mirrors \gls{da}'s (Row~\ref{tech:da}): at the \gls{llm} tier, \gls{icl} achieves rapid few-shot task specialization through context conditioning alone, making formal \gls{fsl} training pipelines architecturally redundant for pure few-shot tasks at that scale. \gls{fsl} necessitates version-control tracking of the resulting weight checkpoint; the extremely small \tax{3}{few} dataset constitutes the sole training provenance record and should be fully documented.
\subsection{Temporal Adaptation and Maintenance (Rows~\ref{tech:retraining}--\ref{tech:dil} in Table~\ref{tab:centerpiece})}
\label{sec:TechTemporal}
\textbf{Retraining (Row~\ref{tech:retraining})} 
updates (\tax{1}{paramUpd}) a deployed model on newer or additional data 
to 
restore performance degraded by distribution shift, 
representing the foundational lifecycle maintenance operation across 
all model types. 
Its universality across \taxNo{6}{all} reflects  
reflects the fact that distributional drift is an architecture-agnostic failure mode: any model whose training distribution diverges from its deployment distribution will degrade, regardless of architectural family~\cite{gama_survey_2014}. 
The critical \dGoal divergence from (Full/Partial) \gls{ft} (Rows~\ref{tech:fullFT}--\ref{tech:partFT}) is precise: both operate identically on \dMechanism and \dPersistence, yet retraining preserves the model's original task objective (\tax{2}{DriftRemed}) rather than reshaping its capability boundaries for a new task (\tax{2}{task}). 
This \dGoal distinction carries a direct validation consequence—a retraining event that strictly adheres to the original task specification does not require an intended-use review; an \gls{ft} event does. 
The \tax{3}{seqIncD} coordinate reflects that retraining data arrives in temporally ordered batches whose distributional properties evolve across the model's operational lifecycle—this is structurally distinct from the static, curated datasets of (Full/Partial) \gls{ft} (\taxNo{3}{SmallLargeD}) or \gls{fsl} (Row~\ref{tech:fsl}; \taxNo{3}{smallFew}).
The \taxNo{4}{SchedAdhocPerm} coordinate marks the boundary with \gls{cl} (Row~\ref{tech:cl}): unconstrained retraining overwrites prior weight states with a new fixed checkpoint, whereas \gls{cl}'s \taxNo{4}{boundedUnboundedCumul} encodes a commitment to accumulation rather than replacement. 
The \tax{5}{whole} scope means that every weight is overwritten at each retraining event; there is no modular isolation boundary, so any behavioral drift introduced by the new training data propagates across all task dimensions simultaneously—necessitating automated full-system regression testing against a frozen behavioral baseline before re-deployment, with the historical checkpoint retained as the rollback artifact.

\textbf{\gls{cl} (Row~\ref{tech:cl})} 
addresses the stability-plasticity trade-off~\cite{mccloskey_catastrophic_1989,kirkpatrick_overcoming_2017}: how to accumulate knowledge across sequential tasks or shifting distributions without overwriting previously learned representations. Its defining coordinate is \dPersistence: \taxNo{4}{boundedUnboundedCumul}—a deliberate architectural commitment to knowledge accumulation rather than replacement—which is the main axis that separates \gls{cl} from Retraining (Row~\ref{tech:retraining}, \taxNo{4}{SchedAdhocPerm}) despite their identical \dMechanism and \dData profiles. 
The bounded variant corresponds to sequential acquisition over a finite, pre-enumerable task set (\gls{til}, Row~\ref{tech:til}); the unbounded variant corresponds to open-ended streaming adaptation over an indefinite sequence of distributions (\gls{dil}, Row~\ref{tech:dil}). 
This cumulative persistence coordinate is operationally consequential: the open-ended modification trajectory of \tax{4}{unboundedCumul} fundamentally breaks the assumptions of discrete, periodic regression testing, because the model's behavioral state is never stable—it is continuously accumulating new representational commitments. 
At the \gls{llm} tier, \gls{cl} encompasses three sequential stages—continued pre-training, domain-adaptive pre-training, and continual \gls{ft}~\cite{shi_continual_2024}—each carrying distinct \dData and \dScope profiles.
The \dScope coordinate (\tax{5}{partl}) reflects that production \gls{cl} deployments at scale almost always apply constrained, parameter-isolated updates (adapters, selective layer freezing) rather than whole-model retraining, precisely to limit the forgetting surface. \gls{cl}'s cumulative persistence necessitates continuous behavioral drift monitoring rather than periodic snapshot regression testing; each task or distribution boundary must be logged as a discrete lifecycle event with its own targeted stability evaluation verifying that prior-task performance has not regressed beyond a predefined tolerance.

\textbf{\gls{til} (Row~\ref{tech:til})}
is a specific operational variant of \gls{cl} in which a model acquires a finite, pre-enumerable sequence of distinct tasks—each presented without access to prior task data—with task identity explicitly provided at inference time~\cite{van_de_ven_three_2022}. 
The \tax{4}{boundedCumul} coordinate is the main definitional anchor: the finite, pre-specified task sequence means the total knowledge accumulation horizon can, in principle, be declared in advance, making \gls{til} the only \gls{cl} variant whose change trajectory is pre-specifiable in a technical plan. Its \taxNo{5}{partlMod} scope reflects that the dominant practical implementations at the \gls{dl} and \gls{fm} tiers use parameter-isolated architectures—dedicated output heads, task-specific adapters, or masked parameter subsets—that physically separate each task's learned representation from prior ones, directly mitigating the forgetting problem through structural disentanglement rather than regularization. The \taxNo{6}{df} ceiling is imposed by architectural supersession: at the \gls{llm} tier, domain-adaptive pre-training and continual \gls{peft} subsume \gls{til}'s goals under the broader \gls{cl} (Row~\ref{tech:cl}) framework, rendering dedicated \gls{til} protocols redundant.
Compared to \gls{dil} (Row~\ref{tech:til}), \gls{til}'s inference-time task identifier is a system-level dependency that must be tracked in deployment documentation—if the task-identity signal is unavailable or corrupted, the deployed system's behavior is undefined, constituting a unique failure mode that requires explicit specification in the operational validation plan. \gls{til}'s bounded, pre-specifiable task sequence supports formal pre-deployment documentation of the complete capability accumulation trajectory; the modular scope permits per-task regression testing at each task boundary, isolating validation to the newly added task module without re-validating prior task components.

\textbf{\gls{dil} (Row~\ref{tech:dil})}
is the open-ended (\tax{4}{unboundedCumul}) variant of \gls{cl}: the model encounters a potentially indefinite sequence of domains or class distributions without access to prior training data and—critically—without a task identifier at inference time~\cite{van_de_ven_three_2022}. 
This absence of inference-time task identity is the defining coordinate divergence from \gls{til} (Row~\ref{tech:til}).
The \taxNo{5}{partWhole} scope reflects that some \gls{dil} approaches retrain the entire network 
while others selectively update domain-adaptive components. 
The compound \dGoal profile—\taxNo{2}{ContAdaptDriftRemed}—captures \gls{dil}'s dual operational goal: it simultaneously accumulates new distributional knowledge (\tax{2}{ContAdapt}) and preserves performance on prior distributions that may re-emerge (\tax{2}{DriftRemed}). 
The \taxNo{6}{df} ceiling is imposed by the same supersession mechanism as \gls{til}: at the \gls{llm} tier, continued pre-training and domain-adaptive \gls{peft} subsume \gls{dil}'s goals. \gls{dil}'s unbounded modification trajectory is the most operationally demanding persistence profile in the taxonomy: the system's behavioral state cannot be fully characterized at any fixed checkpoint, necessitating continuous runtime monitoring with per-distribution performance tracking and explicit drift-impact assessments at each domain boundary transition.


\subsection{
Alignment, Reasoning, and Trustworthiness
(Rows~\ref{tech:sft}--\ref{tech:AdvrsTrn} in Table~\ref{tab:centerpiece})}
\label{sec:TechAlignment}

\textbf{\gls{sft} (Row~\ref{tech:sft})}, also termed instruction tuning, establishes the foundational behavioral capability for an \gls{llm} to interact via natural language instructions~\cite{ouyang_training_2022}, and its taxonomic position demands precise separation from classical task-specific \gls{ft} (Rows~\ref{tech:fullFT}--\ref{tech:peft}). 
Both employ \tax{1}{paramUpd} via supervised \gls{gd}, but the \dData divergence is categorical: task \gls{ft} operates on \tax{3}{smallD} datasets of narrowly scoped, single-function examples yielding a bounded operational envelope, whereas \gls{sft} employs a massively heterogeneous \tax{3}{instrResp}—spanning question answering, summarization, code generation, reasoning chains, and multilingual instructions—to instill a general instruction-following capability that cuts across all task boundaries.
This \dData divergence propagates into the \dGoal profile: \gls{sft} achieves a multi-domain form of \tax{2}{task} alongside baseline \taxNo{2}{algnSafe}, rather than the single-function specialization of task \gls{ft}. 
While traditional \gls{sft} relies on human-curated datasets, contemporary variants frequently utilize synthetic, model-generated data~\cite{wang_self-instruct_2023} (e.g., Instruction Backtranslation~\cite{li_self-alignment_2023}, \gls{raft_rank}~\cite{dong_raft_2023}, and Self-Rewarding models~\cite{yuan_self-rewarding_2024}). This introduces a critical \dData nuance: it shifts the compliance and bias-auditing burden away from human annotator governance and onto the upstream teacher model generating the synthetic corpus.
Safety \gls{sft} is a specific instantiation that explicitly enforces \tax{2}{safety} by restricting the corpus to refusal demonstrations and harm-category filtered examples.
The \dScope coordinate—\taxNo{5}{partWhole}—is implementation-dependent: 
\gls{sft} applied to all parameters yields \tax{5}{whole} with a full-system regression burden. 
Because \gls{sft} reshapes the model's behavioral envelope across all tasks simultaneously—not just a single bounded function—it demands comprehensive cross-domain behavioral regression testing before re-deployment; a task-specific regression suite is insufficient; the validation protocol must cover the full instruction-following surface, including all task types represented in the instruction-response corpus.

\textbf{\gls{rlhf} (Row~\ref{tech:rlhf}).} 
\gls{rlhf} trains a reward model on human preference pairs and optimizes an \gls{llm} policy against it, constituting the canonical alignment technique for generative models~\cite{christiano_deep_2017}. 
The \tax{3}{pref} coordinate is the mechanism's defining input requirement and its most consequential data-governance obligation: unlike \tax{3}{smallD} datasets annotated for a fixed objective, preference pairs encode comparative behavioral judgments—which of two model outputs a human annotator prefers—making annotator demographics, selection criteria, and inter-rater reliability first-class artifacts in the training provenance record.  
The \taxNo{2}{algnSafe} goal and the strict \taxNo{6}{lm} bound reflect that preference-based alignment is architecturally contingent on the model's capacity for open-ended natural-language generation; the preference signal is structurally meaningless for discriminative or shallow architectures. 
The \dScope coordinate is implementation-dependent in a way that carries direct validation consequences: applying \gls{rl} policy optimization across all network parameters yields \tax{5}{whole}, demanding full-system regression testing; applying it via \gls{peft} adapters restricts the scope to \tax{5}{partl}, isolating the change-control footprint to the adapter; 
this structural distinction is operationally critical, as it determines whether an alignment modification event demands full-system regression testing or a targeted, component-level validation assessment.
Training the reward model introduces a third \tax{5}{surrogate} component—a fully independent model artifact that triggers its own conformity validation chain.
This last point is the taxonomically critical divergence from \gls{dpo} (Row~\ref{tech:dpo}): both share \dMechanism, \dGoal, \dData, and \dPersistence, but \gls{rlhf}'s reward model produces a \tax{5}{surrogate} footprint that \gls{dpo} eliminates entirely by recasting alignment as a direct classification objective. \gls{rlhf} deployments require independent validation of the reward model artifact as a separate system component; any update to the reward model constitutes a distinct change-control event from the policy update, as the reward model defines the optimization target for all subsequent policy training.
  
\textbf{\gls{dpo} (Row~\ref{tech:dpo})} 
eliminates the explicit reward model required by \gls{rlhf}, 
recasting alignment as a supervised classification objective directly on preference pairs~\cite{rafailov_direct_2023}.
\gls{dpo} shares \dMechanism, \dGoal, \dData, and \dPersistence with \gls{rlhf} (Row~\ref{tech:rlhf})—making the \dScope divergence the sole but operationally decisive difference between them.
\gls{dpo}'s elimination of the reward model collapses \dScope to \taxNo{5}{partWhole} alone, removing the \tax{5}{surrogate} component entirely. This \dScope reduction is the most system-engineering-consequential property of \gls{dpo}: it eliminates the need for a separate reward model validation pipeline, 
and the three-stage orchestration infrastructure.
Compared to \gls{rlaif}/\gls{cai} (Row~\ref{tech:rlaif}), \gls{dpo} shares the same \dScope simplification but retains the \tax{3}{pref} human-annotation requirement; \gls{rlaif} substitutes AI-generated constitutional preferences (\tax{3}{synthRule}), shifting the validation burden from annotator oversight to constitutional text auditing. \gls{dpo}'s structural simplicity narrows the validation surface to a single policy artifact; however, the preference pair dataset remains the sole provenance record determining the alignment boundary, and its annotator-selection protocol, demographic coverage, and inter-rater reliability metrics must be fully documented as the system's behavioral specification.
Recent advancements have expanded this family into broader preference-optimization variants, including \gls{ipo}~\cite{azar_general_2024}, \gls{kto}~\cite{ethayarajh_model_2024}, and \gls{simpo}~\cite{meng_simpo_2024}. 
Notably, \gls{orpo}~\cite{hong_orpo_2024} collapses the traditional multi-stage pipeline by integrating baseline instruction tuning and alignment into a single gradient update. 
  
\textbf{\gls{rlaif}/\gls{cai} (Row~\ref{tech:rlaif})} 
share the optimization infrastructure of \gls{rlhf} but fundamentally substitutes the \dData profile~\cite{lee_rlaif_2024}: human-annotated preference pairs are replaced by synthetic preferences generated by an AI model operating under a set of constitutional principles (\tax{3}{synthRule})~\cite{bai_constitutional_2022}. 
This \dData substitution is the technique's defining taxonomic property and its 
consequential system-engineering decision: 
it transfers the entire source of the alignment signal from human annotators—whose demographics, biases, and consistency are tracked through inter-rater reliability metrics—to a text-based constitutional document and an AI feedback model whose behavioral properties, latent biases, and edge-case failure modes must instead be audited. 
Compared to \gls{rlhf} (Row~\ref{tech:rlhf}) and \gls{dpo} (Row~\ref{tech:dpo}), \gls{rlaif}/\gls{cai} shares \dMechanism, \dGoal, \dPersistence, \dScope, and \dModel but diverges exclusively on {\dData}—making the data-source substitution the main structural change. 
The scalability advantage is architectural: AI annotation parallelizes without human bottlenecks, enabling preference dataset generation at orders-of-magnitude larger scale than human annotation pipelines permit. 
Because the constitutional text and the AI feedback model jointly constitute the alignment specification, both must be treated as version-controlled engineering artifacts; any modification to either triggers a new change-control event, and the validation protocol must include semantic-stability testing of the constitutional principles across edge-case behavioral scenarios and regression testing of the AI critic pipeline's preference consistency.

\textbf{\gls{rlvr}/\gls{grpo} (Row~\ref{tech:rlvr})}  
targets \tax{2}{reas}—multi-step problem-solving, planning, and chain-of-thought verification—a goal that diverges fundamentally from both the alignment goal of \gls{rlhf}/\gls{dpo}/\gls{rlaif} (\taxNo{2}{algnSafe}) and the \tax{2}{task} goal of \gls{ft}~\cite{shao_deepseekmath_2024,deepseek_r1_2025}.
The \dData coordinate is also a differentiator: \tax{3}{verify} relies on programmatic reward signals (e.g., mathematical correctness verified by symbolic solvers, code execution verified by test suite pass/fail, logical entailment verified by formal proof checkers) rather than subjective human preference annotation. 
This substitution does not merely simplify the data pipeline; it categorically changes the nature of the validation obligation. The preference-pair regime (\gls{rlhf}/\gls{dpo}) requires mitigating annotator bias and tracking demographic representation; the verifiable-task regime requires instead certifying the correctness and coverage of the programmatic oracle—a shift from human-factors verification to formal functional-specification testing.
The \dData-based differentiation between \gls{rlvr} and \gls{rlhf} is the clearest demonstration of the multi-dimensional taxonomy's analytical value: a mechanism-only single-axis framework collapses both into a generic ``\gls{rl}-based tuning'' category, completely obscuring that they target distinct \dGoal goals and require radically different \dData provenance documentation.

\textbf{\gls{lp} (Row~\ref{tech:lp})}
is framed as an interpretability and explainability intervention—determining its \dGoal assignment as \tax{2}{explain}—rather than merely as a downstream evaluation metric, task performance, or behavioral alignment~\cite{yosinski_how_2014}. 
By freezing all base model parameters and training only a lightweight linear classifier appended to a designated intermediate layer, \gls{lp} surfaces latent representational structure as interpretable task-specific outputs without modifying the base model's core feature extraction function~\cite{alain_understanding_2018}. 
The \tax{1}{paramUpd} classification applies exclusively to the probe head's weight matrix; the base remains mathematically unchanged.
Its \tax{5}{modSwap} scope indicates that \gls{lp} probes constitute structurally decoupled artifacts whose installation and removal leave the base model's parametric state and validated behavior unaffected.
Compared to \gls{peft} (Row~\ref{tech:peft}), which also introduces a modular artifact onto a frozen base, \gls{lp} diverges critically on \dGoal: \gls{peft} targets task performance (\taxNo{2}{taskCompEff}), while \gls{lp} targets representational transparency. Compared to \gls{leace} (Row~\ref{tech:leace}), which operates via \tax{1}{WeightSpaceComp} to algebraically rewrite the base model's weight matrices, \gls{lp} never touches the base weights at all.
For production deployments requiring verifiable transparency, the probe's classification performance across sensitive concept dimensions can be documented as engineering evidence of representational alignment or bias.
Because \gls{lp} does not modify the base model's parameters, it does not trigger a model-level change-control event; the decoupled probe artifact (\tax{5}{modSwap}) requires its own version-controlled documentation specifying the target layer, concept labels, and labeled probe dataset—but the base model's existing validation record remains entirely unaffected.

\textbf{\gls{leace} (Row~\ref{tech:leace})} 
targets \tax{2}{fairness} through a mechanistically novel \dMechanism pathway: rather than employing iterative \gls{gd}, it operates via \tax{1}{WeightSpaceComp}—computing a covariance-based orthogonal projection from a \tax{3}{smallD} and algebraically projecting the model's weight matrices into the null space of a target concept's linear representation~\cite{belrose_leace_2023}, a parameter composition operation analogous to Task Arithmetic (Row~\ref{tech:taskArith}) but operating on concept subspaces rather than task capability directions.
This closed-form linear algebra operation bypasses the computational footprint of traditional parameter-updating methods entirely, producing a \tax{4}{AdhocPerm}, \tax{5}{partl} modification at negligible training cost. 
Compared to \gls{lp} (Row~\ref{tech:lp}), which trains a probe head on top of a frozen base to expose linear concept encodings, \gls{leace} inverts the operation: it takes the same linear concept subspace identified by probing and algebraically projects it out of the base model's weight matrices—making \gls{lp} and \gls{leace} complementary, with \gls{lp} as the diagnostic step and \gls{leace} as the corrective intervention targeting the same identified representation. 
The \tax{4}{AdhocPerm} Permanent coordinate reflects that \gls{leace} produces an irreversible, event-driven modification at negligible compute cost—unlike gradient-based fairness interventions that require training runs, \gls{leace} can be applied as a post-hoc weight surgery.
The \tax{5}{partl} scope is bounded by the targeted layers: only the weight matrices encoding the concept's linear subspace are projected; all other parameters remain unchanged. 
Its operational and validation significance lies in the deterministic auditability of the intervention: 
because the modification is derived analytically from a strict concept specification, it is fully reproducible and mathematically characterizable; the concept subspace, projection matrix, and labeled probe dataset constitute a complete audit trail for the fairness intervention—validation requires verifying that the modified representation is linearly indecipherable for the target concept while baseline task performance is preserved, both of which are statistically testable properties.

\textbf{Adversarial training (Row~\ref{tech:AdvrsTrn})} 
improves model robustness by incorporating adversarially perturbed examples directly into the optimization objective through a min-max formulation~\cite{madry_towards_2018}. 
The \tax{1}{paramUpd} coordinate reflects the defining operational property: adversarial perturbations are generated dynamically within the optimization loop, meaning the gradient signal itself is adversarially shaped at each training step—this is the mechanistic distinction from Data Augmentation (Row~\ref{tech:dataAug}), which modifies training inputs as a static pre-processing step independent of the gradient loop.
This distinction is taxonomically consequential: adversarial training is a gradient-coupled technique whose computational cost scales with the inner maximization procedure at each step, while Data Augmentation's cost is absorbed in the data-loading stage. Both share \tax{2}{rob}, but only adversarial training carries \tax{1}{paramUpd} with an adversarially shaped gradient signal, rather than the \tax{1}{manip} mechanism of augmentation. 
The \taxNo{6}{dfl} ceiling reflects an active research frontier: cross-modal adversarial robustness for vision-language \gls{mllm} architectures lacks standardized post-training protocols and established benchmark coverage, making the \gls{mllm} tier currently out of scope.
Adversarial training universally trades off standard operational performance against robustness~\cite{tsipras_robustness_2018}; this Pareto trade-off must be explicitly characterized in the validation record, with the robustness-accuracy frontier documented as a system performance boundary—any deployment threshold revision following adversarial training constitutes a system behavior change requiring separate regression validation.

\subsection{Training Strategies (Rows~\ref{tech:metaLrn}--~\ref{tech:activeLrn} in Table~\ref{tab:centerpiece})}
\label{sec:TechStrategies}
\textbf{Meta-learning (Row~\ref{tech:metaLrn})}  
frames adaptation itself as an optimization objective: rather than learning to perform a single task, the model learns an initialization, metric space, or learning procedure that enables rapid generalization to new tasks from minimal data (\tax{3}{taskDist})~\cite{finn_model-agnostic_2017,hospedales_meta-learning_2021}. 
Its \dData coordinate is the defining characteristic: instead of a labeled dataset for one task, meta-learning requires a diverse distribution of training tasks—a data-provenance structure qualitatively different from the task-specific corpora of Full \gls{ft} (\tax{3}{largeD}) or \gls{peft} (\tax{3}{smallD}).
Its \dModel ceiling at \taxNo{6}{df} encodes a significant structural observation within the taxonomy: at the \gls{llm} tier, \gls{icl} (Row~\ref{tech:icl}) has architecturally supplanted gradient-based meta-learning as the primary mechanism for rapid few-shot adaptation.
The supersession is mechanistically grounded: \glspl{llm} contain sufficiently rich pre-trained representations that few-shot task generalization can occur through in-context conditioning (\tax{1}{ctx}, \tax{4}{eph}) rather than through expensive parametric optimization (\tax{1}{paramUpd}, \taxNo{4}{SchedAdhocPerm}).
This supersession relationship dictates that system engineers deploying \glspl{llm} should route rapid adaptation requirements through prompt-level \gls{icl} architectures, reserving formal meta-learning pipelines exclusively for \gls{dl} and early \gls{fm} deployments. 
The \dPersistence coordinate is counterintuitive for a technique framed around adaptability: meta-learning produces a permanent parametric modification (\taxNo{4}{SchedAdhocPerm})—not a transient or session-ephemeral one. This means meta-learning demands the same rigorous full-system regression testing and baseline validation protocols as any standard training event, structurally distinct from \gls{icl}'s zero-artifact execution.
Meta-learning requires that the task distribution used for outer-loop training be fully documented as the primary provenance record; the diversity and coverage of this task distribution directly determines the generalization bounds of the resulting meta-initialization, and any modification to the task distribution constitutes a change event requiring re-validation of the adaptation baseline.
 
\textbf{\gls{mtl} (Row~\ref{tech:mtl})} 
jointly optimizes shared representations across a fixed task set, exploiting inter-task relatedness to improve generalization on each individual task~\cite{caruana_multitask_1997}. 
The defining coordinate is \tax{3}{taskDist}: the training corpus spans a predetermined collection of tasks simultaneously available in memory, rather than a single-task labeled dataset (Full \gls{ft}, Row~\ref{tech:fullFT}, \tax{3}{largeD}).
Its structural distinction from Meta-Learning (Row~\ref{tech:metaLrn}) is that, while both share \tax{3}{taskDist}, \gls{mtl}'s task set is statically fixed at training time rather than dynamically drawn from a distribution.
The \tax{5}{whole} scope is the critical consequence of the joint optimization: because all task-specific output heads share the same backbone parameters, there is no modular isolation boundary between tasks—any change to the shared encoder propagates across all tasks simultaneously. 
This means modifying the task set, task-weighting schedule, or encoder architecture requires full-system regression testing across all tasks to verify that positive transfer to new tasks has not induced negative interference on existing ones. 
At the \gls{mllm} tier, \gls{mtl} is the foundational architecture underlying multimodal instruction tuning (Row~\ref{tech:multimodInstr})—a unified backbone must serve heterogeneous vision, audio, and text task heads simultaneously.
\gls{mtl} requires comprehensive cross-task regression testing at every task-set modification boundary; the inter-task interference structure must be explicitly characterized and documented, as negative transfer to any task constitutes a behavioral regression on that task's validated performance envelope.

\textbf{Self-play (Row~\ref{tech:selfplay})}
generates training data by having a model compete against or evaluate itself, producing an auto-curriculum of increasing difficulty without requiring externally provided labeled data.
The \tax{3}{synthEnv} coordinate is the defining structural property: the training signal is generated internally through the model's own interactions with copies or historical checkpoints of itself—no human annotation, no static corpus, and no external oracle are involved in data generation. 
This self-referential data source produces the compound \taxNo{2}{taskAlgnReas} profile—the three operationally distinct goals served by self-play at different \dModel levels: domain mastery through competitive co-evolution in \gls{dl} game agents (\tax{2}{task})~\cite{silver_mastering_2017}, instruction-following improvement through self-generated preference pairs in \gls{llm} alignment (\tax{2}{algn})~\cite{chen_spin_2024}, and chain-of-thought reasoning cultivation through iterative self-verification in reasoning models (\tax{2}{reas})~\cite{zelikman_star_2022}.
This multi-goal profile would be irresolvable in a single-axis taxonomy; the taxonomy's value is precisely in making explicit that the same \dMechanism and \dData regime serve fundamentally different \dGoal depending on the deployment context—and that each dictates entirely distinct evaluation metrics, regression test suites, and operational baselines.
Compared to Curriculum Learning (Row~\ref{tech:curriLrn}), which also generates structured training sequences, self-play mainly differs on \dMechanism (\tax{1}{paramUpd} vs. \tax{1}{manip}) and \dData (\tax{3}{synthEnv} vs. \tax{3}{pipeInherit})—Curriculum Learning sequences existing data, while self-play generates new data through model-environment interaction.
The \taxNo{6}{dl} scope (skipping \gls{fm}) reflects that standard \gls{fm} \gls{ft} pipelines do not typically employ self-play, which re-emerges only as a specialized adaptation mechanism for alignment and reasoning at the \gls{llm} tier. 
Because the specific \dGoal being pursued dictates the validation protocol, any self-play deployment must explicitly document which goal dimension applies, providing a clear audit trail for the intended-purpose and risk-rationale in compliance documentation.

\textbf{Curriculum learning (Row~\ref{tech:curriLrn})} 
structures the ordering of training examples rather than their content, presenting data in an easy-to-hard sequence to improve optimizer trajectory and final convergence~\cite{bengio_curriculum_2009}.
Its \tax{1}{manip} classification is the decisive taxonomic property: curriculum learning never executes a gradient update on the model—it operates strictly as a data-sequencing function upstream of the optimizer. The parametric modification that results from training on a curricularized dataset is performed entirely by the downstream training method; curriculum learning contributes only the temporal ordering of the input stream. 
This indirect mechanism is the structural reason for \tax{4}{pipeDep} and \tax{5}{pipelineDepScope}: the \dPersistence and \dScope profile of the resulting model are fully inherited from whichever downstream training method consumes the sequenced data.
Compared to Data Augmentation (Row~\ref{tech:dataAug}),
Data Augmentation synthetically extends the training set (via transformation), while curriculum learning operates on the existing dataset without modification—it changes the sampling order, not the examples.
Compared to Active Learning (Row~\ref{tech:activeLrn}), both share \tax{3}{pipeInherit}, but Active Learning introduces a \tax{2}{domAdapt} component by querying strategically chosen unlabeled samples, while Curriculum Learning's \tax{2}{task} reflects that it optimizes the convergence trajectory rather than the distributional coverage.
Because curriculum learning does not independently modify model parameters, it must be documented as a named algorithmic dependency of the primary training event it serves—specifying the difficulty-scoring methodology, sequencing protocol, and any pacing functions—rather than as a standalone change-control entry; the downstream training event inherits the full parametric and structural documentation burden.

\textbf{Active learning (Row~\ref{tech:activeLrn})} 
strategically selects the most informative unlabeled samples for oracle annotation, maximizing model improvement per labeled example by targeting the data points the model is most uncertain about~\cite{settles_active_2009}.  
Like Curriculum Learning (Row~\ref{tech:curriLrn}), it carries \tax{1}{manip}: the model's parameters are updated by a downstream training method, not by the active learning mechanism itself.
However, Active Learning diverges from Curriculum Learning on \dGoal: where curriculum learning sequences existing labeled data for convergence optimization (\tax{2}{task} only), active learning's query strategy inherently introduces a \tax{2}{domAdapt} component—by selectively targeting uncertainty-rich, informationally dense regions of the unlabeled distribution, it constructs a labeled dataset whose coverage is deliberately biased toward the model's current capability boundaries.
This \dGoal distinction is operationally significant: curriculum learning does not change the statistical composition of the labeled dataset; active learning does, and the algorithmic selection strategy is the direct determinant of the distributional coverage of the resulting training corpus.
The \tax{4}{pipeDep} and \tax{5}{pipelineDepScope} coordinates confirm that persistence and structural scope are fully inherited from the downstream training method—but the labeled dataset itself, as an artifact of the active selection process, constitutes an independent data-provenance record that must be documented and audited separately.
The active learning acquisition function must be explicitly documented and version-controlled as a system configuration artifact; the selection strategy directly defines the boundaries of the model's validated operational envelope, and any change to the acquisition function constitutes a data-pipeline modification requiring re-evaluation of the distributional coverage of the resulting labeled corpus before downstream training commences.


\subsection{Data-Centric and Privacy-Preserving Methods (Rows~\ref{tech:dataAug}--\ref{tech:fl} in Table~\ref{tab:centerpiece})}
\label{sec:TechDataCentric}
\textbf{Data Augmentation (Row~\ref{tech:dataAug})} 
increases training data diversity through synthetic transformations of existing examples, without acquiring new labeled data~\cite{shorten_survey_2019,feng_survey_2021}. 
Its \tax{1}{manip} classification reflects its structural position: augmentation operates strictly upstream of the gradient loop, modifying training inputs 
before they reach the optimizer—it executes no forward pass on the model and computes no gradients.
This upstream positioning means the technique shares \tax{4}{pipeDep} and \tax{5}{pipelineDepScope} coordinates with all other boundary-extension methods (Curriculum Learning, Active Learning, \gls{semiSL}), reflecting that the \dPersistence and \dScope of any resulting model change are entirely determined by the downstream training method that consumes the augmented data.
The compound \taxNo{2}{taskRob} profile reveals a structurally important internal distinction within augmentation: standard transformations (flips, crops, synonym substitution) diversify the input distribution to improve generalization (\tax{2}{task}), while adversarially constructed perturbations specifically target distributional robustness (\tax{2}{rob})~\cite{tsipras_robustness_2018}.
The latter variant overlaps in goal with Adversarial Training (Row~\ref{tech:AdvrsTrn}), but differs critically on \dMechanism: adversarial training generates perturbations dynamically within the gradient loop (d\tax{1}{paramUpd} with adversarially shaped gradients), while adversarial augmentation applies fixed perturbation transforms as a static pre-processing step (\tax{1}{manip})—an operationally meaningful distinction because the dynamic coupling in adversarial training produces adaptive perturbations that scale with the model's current vulnerability, while static augmentation perturbations cannot.
The augmentation policy—comprising the exact set of transformation functions, their mathematical parameterizations, hyperparameter bounds, and random seed configurations—constitutes a component of the training data specification and must be version-controlled and documented as an immutable dependency of the training event; a change to the augmentation policy redefines the statistical distribution the model optimizes against and requires a new data-provenance record.

\textbf{\gls{semiSL} (Row~\ref{tech:semiSl})} 
leverages abundant unlabeled data alongside a small labeled set, making it highly valuable in domains where annotation is prohibitively expensive. 
Its \tax{1}{manip} classification groups it with the other data-pipeline strategies, but \gls{semiSL} occupies a structurally distinct position within this group: unlike Curriculum Learning (which sequences existing labeled data) and Data Augmentation (which transforms existing labeled data), \gls{semiSL} introduces unlabeled data—whose labels were never verified by a human or certified oracle—as a training signal source~\cite{berthelot_mixmatch_2019,sohn_fixmatch_2020}.
\gls{semiSL} is uniquely complex because it integrates two distinct data sources (\taxNo{3}{unlabSmallD}). Because the unlabeled pool lacks ground-truth verification, its inherent sampling bias and distributional coverage directly shape the model’s learned behavior, creating a structural transparency gap that distinguishes \gls{semiSL} from purely supervised pipelines.
Compared to \gls{ssl}/\gls{cpt} (Row~\ref{tech:ssl}), which shares the \tax{3}{unlabD} dependency and uses self-supervised objectives directly, \gls{semiSL} is \tax{1}{manip}—it does not itself prescribe an optimization objective or update parameters; it delivers a mixed labeled-and-pseudo-labeled dataset to a downstream training method. \gls{ssl}/\gls{cpt} is \tax{1}{paramUpd}, defining its own objective and directly executing weight modifications.
Because the unlabeled data pool's distributional properties directly determine the scope and boundaries of the downstream model's validated capabilities, the unlabeled corpus must be subjected to the same provenance documentation standards as the labeled set—including source identification, collection methodology, distributional characterization, and bias auditing.
 
\textbf{\gls{ssl}/\gls{cpt} (Row~\ref{tech:ssl})} 
occupies a structurally dual role in the taxonomy (marked $\dagger$ in Table~\ref{tab:centerpiece}): foundational pre-training from scratch is explicitly out of scope, but applying \gls{ssl} as an adaptation technique to an already-deployed architecture is within scope.  
The critical coordinate that distinguishes \gls{ssl}-as-adaptation from all other pipeline-mediated data strategies 
is \tax{1}{paramUpd}~\cite{gururangan_dont_2020}: unlike Data Augmentation (Row~\ref{tech:dataAug}) or \gls{semiSL} (Row~\ref{tech:semiSl}), which modify the data pipeline and delegate parameter updates to a downstream method, \gls{cpt} directly defines its own optimization objective and formally executes \gls{gd} on model weights.
The \tax{3}{unlabD} coordinate is shared with \gls{semiSL} (Row~\ref{tech:semiSl}), but the mechanism is structurally distinct: \gls{semiSL} uses unlabeled data to construct pseudo-labeled training examples for a downstream training method; \gls{cpt} uses unlabeled data as the direct training signal for its own self-supervised objective, making it a non-supervised parametric adaptation technique in the taxonomy. 
The \dPersistence coordinate bifurcates based on execution mode: \tax{4}{ScheduledPerm} when \gls{cpt} is applied as a discrete offline batch on a fixed domain corpus; \tax{4}{unboundedCumul} when applied as continuous domain-specific streaming—the latter creating an open-ended modification trajectory that breaks the assumptions of static regression testing and demands dynamic behavioral monitoring.
At the \gls{llm} and \gls{fm} tiers, the primary adaptation strategy leveraging \gls{ssl} objectives is \gls{cpt}: applying self-supervised objectives (e.g., next-token prediction) to unlabeled, domain-specific corpora to internalize novel vocabulary, syntax, and factual structures without requiring supervised labels. \gls{cpt} functions as an initialization phase preceding \gls{sft} (Row~\ref{tech:sft}) or \gls{peft} (Row~\ref{tech:peft}), providing a domain-enriched representational baseline that ensures the model possesses foundational domain competency before task-specific supervision is applied~\cite{gururangan_dont_2020}.

\textbf{\gls{dpft} (Row~\ref{tech:dpft})},
using \gls{dpsgd}, incorporates gradient clipping and noise injection during training to provide mathematical guarantees against membership inference and data extraction attacks.
\gls{dpft} is a \tax{1}{paramUpd} dedicated exclusively to the \tax{2}{priv} goal. 
Unlike \gls{fl}, which achieves privacy via \tax{5}{dIVdist} data sovereignty, \gls{dpft} operates on centralized data (\taxNo{3}{SmallLargeD}) but mathematically protects the records. 
Like standard parametric updates, \gls{dpft} produces a permanently modified weight checkpoint that can be deployed on a planned cadence or as an event-driven patch, assigning it \taxNo{4}{SchedAdhocPerm}.
It can be applied to the \tax{5}{whole} or as \gls{dppeft} (\tax{5}{partl}). Because the \gls{dpsgd} algorithm is foundational, its applicability spans deeply into the hierarchy, yielding \taxNo{6}{dflm}.
 
\textbf{\gls{fl} (Row~\ref{tech:fl})} 
trains a global model by aggregating gradient updates from local models on decentralized datasets, ensuring raw training data never leaves local devices~\cite{kairouz_advances_2021,mcmahan_communication-efficient_2023}. 
Its \tax{1}{paramUpd} classification reflects that \gls{fl} does execute \gls{gd} and produce formal weight modifications—distinguishing it from pipeline-mediated data strategies—but the critical structural differentiator is \tax{3}{decent}: no centralized training corpus exists.
Each client computes gradients locally on its private data shard and transmits only model updates (weight deltas or gradients) to the server for aggregation; raw data never leaves the originating device. This decentralized data geometry is what justifies \tax{2}{priv} as a \dGoal: \gls{fl} does not merely apply privacy as a constraint layered onto a task-training objective—privacy preservation is an intrinsic structural property of the data access pattern, not an add-on mechanism.
The absence of centralized training data means that data-quality auditing, distributional characterization, and bias assessment must be delegated to client-local verification processes or statistical reporting protocols; the server cannot directly inspect the training corpus, making third-party data-provenance attestation mechanisms a system-engineering requirement rather than an optional audit supplement.


\subsection{Efficiency and Composition (Rows~\ref{tech:kd}--\ref{tech:moe} in Table~\ref{tab:centerpiece})}
\label{sec:TechEfficiency}
\textbf{\gls{kd} (Row~\ref{tech:kd})} 
transfers 
a trained teacher model's learned behavioral distribution into
a smaller student model, most commonly by training the student to match the teacher's soft probability output distribution, enabling compression of capability without commensurate compression of performance~\cite{hinton_distilling_2015,gou_knowledge_2021}. 
Its \tax{1}{CrossModelTransf} classification captures the defining structural property: a complete, separately trained model is the proximate information source for the student's training—not a labeled dataset, not a teacher's parameters directly, but the teacher's output distribution.
This cross-model information flow distinguishes \gls{kd} from all other compression-adjacent techniques: Model Compression (Row~\ref{tech:compress}) operates on a single model's own parameters 
while \gls{kd} requires a functioning teacher model to generate training signal, making \tax{3}{synthTeacher}. 
The \tax{2}{compEff} goal drives the student's architectural reduction (smaller, faster inference).
From a validation standpoint, a \gls{kd} event establishes a strict behavioral dependency chain: verifying the student model requires comprehensive technical documentation of both the teacher model's original training data provenance and the exact algorithmic distillation protocol used to generate the soft targets, ensuring full auditability of the behavioral alignment downstream.
The student artifact inherits the teacher's behavioral envelope but introduces a compression-induced approximation gap; this gap must be characterized quantitatively—via behavioral regression testing across the teacher's validated operational envelope.

\textbf{Model compression (Row~\ref{tech:compress})} 
reduces the computational footprint of an already-trained model through structural modification of its parameters, producing a deployable compressed artifact with lower memory and inference cost~\cite{han_deep_2016}.
Its \tax{1}{StructReduct} classification explicitly encodes reduction as the defining operation: unlike all \tax{1}{paramUpd} techniques that modify parameters to improve capability, \tax{1}{StructReduct} modifies parameters to reduce resource consumption while attempting to preserve capability. 
Its \tax{2}{compEff} coordinate designates an objective of footprint reduction that inherently lacks a target for intrinsic capability development.
Compared to \gls{kd} (Row~\ref{tech:kd}), which achieves efficiency through architectural reduction of a student trained from teacher outputs (\tax{1}{CrossModelTransf}, \tax{3}{synthTeacher}), Model Compression achieves efficiency by directly restructuring a single model's existing parameters, 
i.e., no second model is required.
Its \dData profile diverges fundamentally based on the chosen algorithmic implementation:
Data-free compression methods 
operate strictly as \tax{3}{wght} interventions; 
because they consume zero training or calibration data, they completely bypass data-provenance pipelines, making them optimal for strict zero-data-access operational deployments. 
Conversely, calibration-dependent compression methods 
require a \tax{3}{unlabD} calibration dataset to calculate activation statistics and minimize quantization error~\cite{frantar_gptq_2023}. 
The \tax{5}{compressed} is a structurally altered model artifact—not a parameter overlay applied to the original—and must be registered as a distinct model version requiring its own validation suite: the compression operation introduces approximation errors whose distribution across the input space cannot be inferred from the original model's validation record; full behavioral regression testing on the compressed artifact against the original model's validated operational envelope is required.
From a regulatory and change-control perspective, this category encompasses two dimensionally distinct approaches. Post-Training Quantization (\gls{ptq}, e.g., \gls{gptq}~\cite{frantar_gptq_2023}) operates as a training-data-free (\tax{3}{wght}) algebraic derivation applied after the model is fully trained. In contrast, \gls{qat} simulates lower precision during active \gls{gd}, classifying it as a distinct \tax{1}{paramUpd} training event with differing validation and data-provenance requirements.
 
\textbf{Task Arithmetic and Model Merging (Row~\ref{tech:taskArith})} 
combine pre-trained and fine-tuned checkpoints into a single multi-capable model through arithmetic operations on parameter space, without any additional training data~\cite{ilharco_editing_2023,wortsman_model_2022,yadav_ties-merging_2024}. 
Its \tax{1}{WeightSpaceComp} and \tax{3}{wght} classification suggest no gradient computation, no data consumption at merge time, no teacher model inference—only algebraic operations on weight tensors.
Machine Unlearning (Row~\ref{tech:mUnLrn}) shares \tax{1}{WeightSpaceComp} via task-vector negation (subtracting a forget-task vector to remove a learned capability), making Machine Unlearning a targeted polarity reversal of the exact same algebraic operation that Task Arithmetic uses to add capabilities.
Compared to \gls{mtl} (Row~\ref{tech:mtl}), which achieves multi-task capability through joint optimization over a shared task-distribution corpus (\tax{1}{paramUpd}, \tax{3}{taskDist}), Task Arithmetic achieves the same compound capability profile through post-hoc parameter composition requiring no additional training and no access to the original task datasets.
To mitigate parameter interference and preserve specialized capabilities during the fusion process, practitioners employ advanced mathematical merge operators, including \gls{ties}~\cite{yadav_ties-merging_2024}, \gls{dare}~\cite{yu_language_2024}, \gls{slerp}~\cite{shoemake_animating_1985, lu_fine-tuning_2025}, and Fisher merging~\cite{matena_merging_2022}.
Consequently, the regression validation suites used to verify targeted machine unlearning—specifically, verifying the erasure of a specific capability while ensuring the preservation of adjacent foundational knowledge—are directly applicable to Task Arithmetic (task-vector operations). 
Because Model Merging bypasses traditional training telemetry, system validation must rely on downstream functional testing and post-merge automated regression benchmarks to verify that parameter-space interference has not degraded core model capabilities.

\textbf{\longContext (Row~\ref{tech:lce})} methods (e.g., \gls{pi}~\cite{chen_extending_2023}, \gls{yarn}~\cite{peng_yarn_2023},
and \gls{pepe}~\cite{hu_pepe_2025}) extend a language model's operational context window by adapting its positional encoding (e.g., through 
\gls{rope}~\cite{su_roformer_2024} scaling) and, when required, fine-tuning on long-sequence data, enabling the model to process inputs far exceeding its original context length.
This operates via a composite \dMechanism: it requires an \tax{1}{arch} to alter the frequency scaling of the \gls{rope}, typically followed by a brief \tax{1}{paramUpd} on long documents. The goal is strictly \tax{2}{capExt}, as it structural expands the model's capacity rather than teaching a new task. It targets the \tax{5}{partl} scope 
and utilizes \taxNo{3}{unlabSmallD} long-sequence corpora. 
Because it permanently alters the positional embedding architecture and corresponding parameters, it produces a versioned checkpoint exhibiting \taxNo{4}{SchedAdhocPerm} persistence.
Unlike techniques that require generation mechanics, context scaling can be applied to base \glspl{genfm}, establishing its \dModel footprint at \taxNo{6}{flm}.

\textbf{\gls{moe} Adaptation (Row~\ref{tech:moe})} 
introduces sparse gating over a population of expert sub-networks or integrating novel expert sub-networks into an existing \gls{moe}, enabling a model to route each input to the subset of parameters most relevant to that input's characteristics—achieving scalable multi-task specialization without proportionally scaling inference cost~\cite{shazeer_outrageously_2017,dou_loramoe_2024,pfeiffer_adapterfusion_2021}.
This structural distinction necessitates a compound \dMechanism profile: the topological addition of a new expert constitutes \tax{1}{arch}—physically restructuring the computational graph—while the subsequent tuning of expert weights or the central routing mechanism constitutes \tax{1}{paramUpd}.
This graph-level intervention is what mandates \tax{5}{modSwap}: the model is 
a collection of individually addressable expert modules coordinated by a router—each expert constituting an independently auditable behavioral unit whose activation patterns can be tracked per input.
Compared to \gls{peft} (Row~\ref{tech:peft}), which adds lightweight parameter overlays (adapters, \gls{lora}) to a frozen base with \taxNo{5}{partlMod}, \gls{moe} Adaptation adds structural routing logic that fundamentally changes the model's computational topology—adapter modules are additive overlays; expert modules are gated substitutes for dense computation.
The compound \dGoal profile—\taxNo{2}{capExtCompEff}—reflects a unique operational property: \gls{moe} adaptation is the only technique in the taxonomy that simultaneously scales systemic capacity while explicitly preserving fixed inference latency bounds. 
The \tax{1}{arch} classification has stringent system-engineering implications in the taxonomy: a new model topology is being registered, not a parametric override of an existing one—requiring a complete re-specification of the model's computational graph and expert routing logic
audit trail as components of the system configuration record, since any change to router architecture or gating threshold alters which computation paths are activated for which inputs.


\subsection{Inference-Time Adaptation (Rows~\ref{tech:pe}--\ref{tech:ttCompute} in Table~\ref{tab:centerpiece})}
\label{sec:TechInference}
\textbf{\gls{pe} (Row~\ref{tech:pe})} 
designs input text to guide 
a deployed model's output distribution entirely through natural language instructions placed in the context window,
without requiring any parameter modification, representing the lowest-overhead adaptation method in the taxonomy~\cite{brown_language_2020,liu_pre-train_2023}. 
Its \tax{1}{ctx} classification suggests that it produces no persistent model artifact whatsoever—no checkpoint, no parameter delta, no weight version, no trained embedding. 
The behavioral effect exists only while the prompt occupies the active context window (\tax{4}{eph}) and operates exclusively at the model's input/output interface (\tax{5}{extContxtAbbr}). 
Compared to \gls{icl} (Row~\ref{tech:icl}), which shares \tax{1}{ctx} and \tax{4}{eph} but adds \tax{3}{few} to steer task-specific output patterns through example-based conditioning, \gls{pe} operates with \tax{3}{zeroSh} through instruction specification alone—relying on generalization from the model's pre-training rather than in-context exemplar matching.
This zero-data, zero-gradient profile positions \gls{pe} as the \tax{2}{behavControl} technique in the taxonomy that imposes the lowest modification overhead but provides the weakest behavioral guarantee: steering is non-parametric and entirely contingent on the model's pre-existing instruction-following capability.
Compared to Prompt Learning (Row~\ref{tech:pl}), which replaces discrete natural-language tokens with continuously differentiable embedding vectors optimized via \gls{gd} (\tax{1}{paramUpd}, \tax{3}{smallD}), \gls{pe} operates entirely in human-legible natural language—making it interpretable but non-optimizable.
The \taxNo{6}{lm} coordinate reflects that effective \gls{pe} requires a model with emergent instruction-following capability; below \gls{llm} scale, models lack the generalization from instruction composition.
Because \gls{pe} produces no auditable architectural or parametric artifact, every deployed system whose behavior is governed by a prompt template must specify that template as a versioned configuration artifact; any change to the prompt—including rewording, structural reordering, or instruction addition—must be treated as a system configuration change requiring re-evaluation of behavioral coverage and output boundary characterization. 
The system-level variant—where core system prompts or metaprompts are hardcoded into production deployment infrastructure—exhibits \tax{4}{verPersistent} persistence.
This boundary requires rigorous tracking within software configuration management and release engineering protocols, as changes to the global system prompt dictate the functional behavioral envelope of the entire application despite the underlying static model weights.

\textbf{Prompt Learning (Row~\ref{tech:pl})} 
converts the discrete, human-authored natural-language prompt into a set of continuously differentiable embedding vectors—soft tokens—prepended to the model's input representation and optimized via \gls{gd} on a task dataset~\cite{lester_power_2021,li_prefix-tuning_2021}. 
It resembles \gls{pe} in its input-space operation—prepending sequence tokens to the input—but functions mechanically like \gls{ft} in its optimization: learnable continuous embeddings (soft tokens) are explicitly trained via backpropagation (\tax{1}{paramUpd}).
Therefore, the parametric modification is confined entirely to the prepended embedding vectors (\tax{5}{partl}); all base model parameters remain frozen.
The \taxNo{4}{SchedAdhocPerm} coordinate means trained soft prompt tensors must be registered as distinct parametric artifacts in the system configuration record: deploying a different soft prompt tensor with the same frozen base model constitutes a parametric modification to the deployed system and requires version control and behavioral regression testing against the task coverage specification defined during original validation.

\textbf{\gls{apo} (Row~\ref{tech:apo})} frameworks, such as \gls{dspy}~\cite{khattab_dspy_2024} or \gls{orpo_prompt}~\cite{yang_large_2024}, replace manual prompt engineering with algorithmic search, treating language models as programmable engines.
\gls{apo} utilizes \tax{1}{ctx} within the \tax{5}{extContxt}, meaning the underlying model parameters remain completely frozen. However, because the optimization compiles a tested, metric-optimized prompt pipeline, the resulting artifact is classified as \tax{4}{verPersistent} rather than \tax{4}{eph}. 
Data requirements are flexible: \gls{apo} can utilize \tax{3}{few} to bootstrap the optimization, but it can also operate strictly \tax{3}{zeroSh} by relying entirely on programmatic evaluation metrics or an "LLM-as-a-judge"~\cite{zheng_judging_2023,li_generation_2025} to score unannotated inputs. Ultimately, this algorithmic search aims for both \tax{2}{task} (optimizing accuracy on a target task) and \tax{2}{behavControl} (systematically enforcing strict output formats, stylistic guidelines, or reasoning templates without altering weights).
 
\textbf{\gls{rag} (Row~\ref{tech:rag})} 
augments \gls{llm} generation by retrieving relevant documents from an external knowledge base and injecting them into the input context at inference time~\cite{lewis_retrieval-augmented_2020}. Its non-parametric nature (\tax{1}{ctx}) and \tax{4}{verPersistent} coordinate confirm that \gls{rag} strictly does not modify underlying model parameters: the injected knowledge alters the model's transient output without touching its foundational weights. 
This fundamentally distinguishes \gls{rag} from Knowledge Editing (Row~\ref{tech:ke}), which permanently rewrites factual associations within the weight space itself. 
The information source is a separately maintained, indexable knowledge repository whose content is decoupled from the model's weights and can be updated independently (\tax{3}{ext})—documents can be added, revised, or removed from the corpus.
This decoupling is the sharpest structural contrast with \gls{cpt} (Row~\ref{tech:ssl}), which internalizes domain knowledge directly into model weights (\tax{1}{paramUpd}, \taxNo{4}{unboundedCumulSchedPerm}): \gls{cpt} makes knowledge intrinsic and persistent at the cost of a full retraining cycle for every knowledge update; \gls{rag} makes knowledge external and updateable at inference time at the cost of retrieval latency and context-window occupation.
The compound \taxNo{2}{taskAugK} profile reflects the dual function: the retrieval mechanism enables the model to answer with temporally current information (\tax{2}{augK}) while the retrieved context shapes the generation to the specific domain and task at hand (\tax{2}{task}).
\gls{rag} functions as a complement to \gls{ft} rather than a direct substitute: \gls{ft} encodes structural behaviors and stylistic patterns directly into parameters, while \gls{rag} retrieves localized factual context from an external corpus. 
From a system engineering and validation perspective, the retrieval corpus constitutes a critical external data dependency. Consequently, the quality, currency, and provenance of the vector database must be rigorously version-controlled; updating the retrieval corpus constitutes a significant system-level configuration change that dictates the model's factual operational envelope and demands end-to-end integration testing, even when the core model parameters remain entirely unmodified.

\textbf{\gls{icl} (Row~\ref{tech:icl})} 
leverages a deployed \gls{llm}'s pattern-matching capability to perform task adaptation from a small number of input-output demonstrations placed in the context window, with no gradient computation and no parameter modification~\cite{brown_language_2020}.
It shares a common structural mechanism—\dMechanism, \dPersistence, \dScope, {\dModel}—with \gls{pe}, but fundamentally differs on {\dData}—the inclusion of \tax{3}{few}—which shifts its primary \dGoal focus toward \tax{2}{task}.

From a system validation perspective, because  
\gls{icl} leaves the foundational model parameters entirely unchanged, 
it
triggers no artifact-level regression testing. Instead, the specific demonstration sets act as transient input dependencies and must be documented.
 
\textbf{\gls{ce} (Row~\ref{tech:ce})} 
is the broader architectural discipline of dynamically assembling an \gls{llm}'s inference context from multiple retrieved, historical, or real-time data sources—encompassing external document retrieval~\cite{gao_retrieval-augmented_2023}, agent memory~\cite{park_generative_2023}, user profiles, tool outputs, and conversational history~\cite{park_generative_2023} (\taxNo{3}{extUser}). 
\gls{ce} differs from \gls{pe} in the structural locus of design: \gls{pe} optimizes the content of a static input prompt, while \gls{ce} designs the automated retrieval pipelines and orchestration systems that dynamically populate the model's context window at inference time. 
\gls{ce}'s \taxNo{3}{extUser} coordinate strictly distinguishes it from \gls{pe} (\tax{3}{zeroSh}) by explicitly demanding complex, version-controlled retrieval and state-management infrastructure.
Crucially, the live inference query serves solely as an execution trigger and data constraint, never as an on-the-fly parametric gradient signal (\taxNot{3}{tstOnly})—a strict architectural boundary that fundamentally separates \gls{ce} from \gls{tta} (Row~\ref{tech:tta}). From a system engineering perspective, \gls{ce} shifts the validation burden away from artifact-level model testing and toward complex state-machine verification. Because the model's operational envelope is dictated by dynamically retrieved external state, regression testing must rigorously validate the orchestration logic, context truncation algorithms, and data pipeline integrity rather than the static underlying model weights.

\textbf{\gls{tta} (Row~\ref{tech:tta})} 
updates model parameters (\tax{1}{paramUpd}) transiently during inference on individual test examples or execution batches (\tax{3}{tstOnly}), remediating distribution shift (\tax{2}{DriftRemed}), systematically reverting those gradient updates before processing the next prediction (\tax{4}{trans})~\cite{wang_tent_2021}. 
\gls{tta} is strictly classified at a \taxNo{6}{df} ceiling because executing transient parameter updates via backpropagation is computationally prohibitive at the \gls{llm} scale; at the \gls{llm} tier, prompt-based adaptation, e.g., \gls{pe} (Row~\ref{tech:pe}), \gls{icl} (Row~\ref{tech:icl}), or Activation Steering (Row~\ref{tech:actSteer}) serves the functionally equivalent role without executing parameter modification. 
To maintain terminological precision, \gls{tta} must be distinguished from \gls{ttt}~\cite{sun_test-time_2020}. While both adapt parameters at inference, \gls{tta} typically relies on unsupervised entropy minimization to counter distribution shift, whereas \gls{ttt} executes transient gradient updates against a distinct self-supervised objective formulated directly on the test instance.
From a system validation perspective, \gls{tta}'s transient modification profile technically bypasses the need for persistent, artifact-level regression testing, as the base artifact theoretically remains unaltered. However, the engineering and quality-assurance burden shifts heavily to execution-state management: the inference pipeline must mathematically guarantee and continuously verify the perfect restoration of the base model weights between inference episodes to prevent insidious cross-inference parameter leakage or progressive state corruption.
 
\textbf{Test-Time Compute Scaling (Row~\ref{tech:ttCompute})} 
dynamically allocates increased inference-time computation budget (\tax{1}{infTimeSearch}) to systematically improve reasoning quality (\tax{2}{reas}) without modifying any underlying model parameters (\tax{5}{extContxt})~\cite{snell_scaling_2024}. 
Unlike \gls{tta} (Row~\ref{tech:tta}), it executes absolutely no gradient updates; unlike \gls{pe} (Row~\ref{tech:pe}), its performance gain derives mechanically from internal search algorithms and multi-step generation topologies (\tax{3}{zeroSh}) rather than explicit input phrasing. 
Its non-parametric \dMechanism classification and \tax{4}{eph} coordinate confirm that it leaves the model architecture mathematically unchanged. Test-time compute scaling serves as the foundational mechanism driving modern reasoning models' characteristic ``extended thinking'' behavior: by generating multiple candidate solutions, applying reward-model reranking, or algorithmically exploring a tree of intermediate reasoning steps, the system intentionally trades inference-time compute latency for output quality~\cite{yao_tree_2023}. 
From a system engineering and validation perspective, test-time compute scaling is strictly an execution strategy rather than an artifact modification. Because the base model weights remain untouched, it bypasses parameter-level regression testing. However, because dynamic search algorithms fundamentally alter the determinism, upper latency bounds, and operational behavioral envelope of the system, this strategy demands rigorous validation through execution-state stress testing.

\subsection{Calibration, Personalization, and Multimodal Adaptation (Rows~\ref{tech:calib}--\ref{tech:multimodInstr} in Table~\ref{tab:centerpiece})}
\label{sec:TechCalibration}
\textbf{Calibration (Row~\ref{tech:calib})} 
post-hoc aligns a model's output confidence scores with its empirical accuracy, correcting the systematic overconfidence or underconfidence produced by standard training objectives that optimize for discriminative accuracy rather than probabilistic faithfulness~\cite{platt_probabilistic_1999,guo_calibration_2017}.
Calibration does not improve the model's task accuracy, 
it improves the correspondence between the model's stated confidence and its actual accuracy (\tax{2}{reliability}).
From a system engineering and safety validation perspective, calibration operates not merely as a performance optimization, but as a critical operational safety constraint. Systems integrated into automated pipelines must mathematically bound their uncertainty; uncalibrated confidence scores inherently invalidate downstream threshold-based decision logic and fail-safe routing, mandating rigorous statistical auditing of calibration metrics.

\textbf{Cross-Modal Alignment (Row~\ref{tech:crossModel})} 
trains a model to map representations from divergent data modalities—most commonly vision and language—into a unified, shared embedding space using contrastive optimization objectives~\cite{radford_learning_2021}.
Its \tax{1}{paramUpd} classification applies to the alignment bridge—the projection network~\cite{liu_visual_2023}, cross-attention module~\cite{li_blip-2_2023}, or contrastive alignment layer~\cite{radford_learning_2021}—while the modality-specific encoders may be frozen.
Cross-Modal Alignment requires aligned pairs (\tax{3}{multiD}) from two modalities—image-caption pairs, video-transcript pairs, audio-text pairs—where the pairing reflects semantic correspondence across modalities, not merely co-occurrence.
The \tax{2}{capExt} coordinate reflects that Cross-Modal Alignment enables capabilities that are structurally impossible in unimodal models, capabilities that require the bridge, not merely improved unimodal encoders.
Its \dModel classification at \tax{6}{fm} reflects the immense architectural scale required: contrastive training across massive, noisy paired corpora is computationally feasible only at the \gls{fm} and \gls{mllm} scale.  
Since the modification is concentrated in the alignment bridge (\tax{5}{partl}), any change to the bridge
must be treated as a capability-level modification requiring full re-characterization of the system's multimodal behavioral envelope.

\textbf{Modality-Specific Adapters (Row~\ref{tech:modelAdapter})}  
extend a frozen multi-modal model's capabilities to new modalities or improve its performance on existing ones by inserting lightweight \gls{peft}-style (\tax{1}{paramUpd}) modules (\tax{5}{modSwap}) within each modality's dedicated processing pathway, without modifying the shared model backbone or cross-modal bridge~\cite{sung_vl-adapter_2022}.
In contrast to Cross-Modal Alignment (Row~\ref{tech:crossModel}), which optimizes the interface between representations to enable semantic integration, Modality-Specific Adapters refine within-modality processing, making them complementary adaptation layers rather than alternatives.
Compared to standard \gls{peft} (Row~\ref{tech:peft}), which applies within a single modality's processing pathway, Modality-Specific Adapters are architecturally instantiated per modality, creating a collection of independently versioned parametric modules whose composition defines the system's active multi-modal capability profile. 
Their classification strictly at \tax{6}{mllm} reflects the architectural prerequisite: independent per-modality adapter injection inherently requires a multi-stream architecture featuring distinct, unimodal computational pathways. 
The \tax{5}{modSwap} 
ensures that modality-specific adaptation artifacts can be independently versioned, tested, and rolled back. 
From a system engineering and validation standpoint, this structural isolation profoundly simplifies configuration management: it mathematically guarantees that high-frequency updates to one modality's adapter (e.g., integrating a new visual encoding standard) can be executed and validated completely independent of the other (e.g., text generation), systematically bypassing the computationally expensive need for end-to-end, multi-stream regression testing.

\textbf{Multimodal Instruction Tuning (Row~\ref{tech:multimodInstr})} 
fine-tunes (\tax{1}{paramUpd}) a multimodal model on multimodal instruction-response pairs
(\tax{3}{multiD})
teaching the model to follow natural-language instructions~\cite{liu_visual_2023}.
Its \dScope varies by implementation (\taxNo{5}{partWhole}). 
The \dData coordinate of \tax{3}{multiD} establishes a highly distinctive data-provenance dependency: unlike text-only instruction tuning, multimodal datasets require perfectly synchronized, temporally aligned instances across modalities. This severely raises the bar for annotation quality pipelines and introduces strict cross-modal consistency as a mandatory validation criterion. From an engineering and system integration perspective, this technique constitutes a high-impact, \taxNo{5}{partWhole} modification event that fundamentally alters the model's core behavioral envelope across all supported modalities simultaneously. Consequently, it mandates exhaustive, cross-modal regression testing to mathematically guarantee that optimization and capability alignment in one specific modality do not destructively interfere with or degrade baseline performance in the others.

\subsection{Knowledge Modification and Activation-Based Adaptation (Rows~\ref{tech:mUnLrn}--\ref{tech:actSteer} in Table~\ref{tab:centerpiece})}
\label{sec:TechKnowledgeMod}
\textbf{Machine Unlearning (Row~\ref{tech:mUnLrn})} 
selectively erases 
learned knowledge, behavioral patterns, or training-data influence (\tax{2}{rem})
from a deployed model without retraining from scratch, 
addressing privacy deletion requirements, copyright compliance, and safety-critical knowledge removal (\tax{2}{priv})~\cite{cao_towards_2015}.
Compared to \gls{leace} (Row~\ref{tech:leace}), which removes a concept's linear influence from activations by modifying the weight matrix's projection geometry (\tax{1}{wght}, applied to a concept direction), Machine Unlearning targets a broader knowledge construct—facts, training data points, behavioral patterns.
Its \dMechanism coordinate spans two distinct mechanistic profiles that must be rigorously distinguished within system configuration management and regression tracking. 
Gradient-based approaches apply gradient ascent on the target forget set (\tax{3}{forget}) to mathematically reverse the historical learning signal, operating as \tax{1}{paramUpd} with a \tax{5}{partl} scope. Conversely, task-vector negation approaches algebraically subtract a localized task vector from the base model's weights to structurally strip out a target capability, operating as \tax{1}{WeightSpaceComp} with \tax{5}{fusedComp}—mechanistically identical to Task Arithmetic (Row~\ref{tech:taskArith}) executed with a negation operator. 
Consequently, they must be governed by the exact same change-control validation logic and interference testing suites. 
Adversarial membership inference attacks, targeted behavioral probes, and jailbreak-style elicitation of the supposedly forgotten knowledge constitute 
some available verification tools, and none provides a formal completeness guarantee; any system deploying Machine Unlearning as a compliance mechanism must document this verification gap explicitly and specify the probe suite used to assess forgetting fidelity~\cite{hu_unlearning_2024}.
  
\textbf{Knowledge Editing (Row~\ref{tech:ke})} 
modifies a specific factual association encoded in a trained model's parameters
without retraining and ideally without disrupting the model's knowledge of logically unrelated facts~\cite{meng_locating_2022}.
Its \dMechanism coordinate is highly implementation-dependent and must be explicitly codified in system configuration and artifact versioning pipelines. Closed-form methods treat internal \gls{mlp} layers as key-value stores and analytically compute rank-one weight updates, operating strictly as \tax{1}{WeightSpaceComp}. Conversely, meta-learning and optimization-based methods apply iterative gradient updates focused on the target fact, operating as \tax{1}{paramUpd}. 
Knowledge Editing structurally differs from Machine Unlearning (Row~\ref{tech:mUnLrn}) primarily in operational directionality: editing corrects or injects data (\tax{2}{augK}), while unlearning ablates it (\tax{2}{rem}). While both share a \tax{4}{AdhocPerm} persistence profile, their \dData differ fundamentally (\tax{3}{smallD} vs.\ \tax{3}{forget}). Crucially, strict sequential scalability bounds remain a limiting operational boundary that must be explicitly quantified in system capability documentation~\cite{gupta_model_2024}.
 
\textbf{Activation Steering (Row~\ref{tech:actSteer})} 
controls \gls{llm} behavior by algebraically injecting computed or learned steering vectors directly into intermediate hidden activations during inference execution, strictly without modifying the underlying stored parameter weights~\cite{zou_representation_2023,rimsky_steering_2024,turner_activation_2023}. 
A notable advancement in this space leverages \glspl{sae} to identify and manipulate interpretable latent features, providing a highly precise, semantically grounded source for generating these transient steering vectors~\cite{cunningham_sparse_2023}.
Its \tax{1}{transAct} classification occupies a mathematically unique structural position: it is mechanistically distinct from both input-layer \gls{pe} (Row~\ref{tech:pe}, which operates exclusively on the discrete input token space) and traditional \gls{ft} (Rows~\ref{tech:fullFT}--\ref{tech:peft}, which alters permanent weight matrices). The \tax{4}{trans}, \tax{5}{actSpace} coordinate profile dictates that activation steering imposes zero permanent structural modification on the base artifact, yet enforces an immediate, highly non-linear behavioral shift during the active inference session. A profound structural insight within this framework emerges when contrasting activation steering with \gls{tta} (Row~\ref{tech:tta}): while both represent transient execution-time adaptations, they physically bifurcate at the architectural scale boundary. 
\gls{tta} executes via transient gradient-based parameter updates at \taxNo{6}{df}, whereas activation steering operates via direct latent-space vector injection at \taxNo{6}{lm}. 
\subsection{Classification Tensions}
\label{sec:ClassTensions}
Several techniques resist clean single-category classification. These tensions are inherent to the adaptation landscape and motivate the multi-dimensional approach.

\textbf{\gls{fsl} (Row~\ref{tech:fsl}).} In this framework, the \gls{fsl} categorization is strictly reserved for parametric approaches—such as gradient-based meta-learning or metric-based networks—where the model's weights undergo a \taxNo{4}{SchedAdhocPerm} update based on limited labeled data. While the term \gls{fsl} is colloquially used to describe providing examples in an \gls{llm} prompt, this framework classifies that non-parametric, \tax{4}{eph} mechanism distinctly as \gls{icl} (Row~\ref{tech:icl}). \gls{icl} functions primarily to induce a \taxNo{2}{taskAugK} mapping during inference, bypassing parameter updates entirely.

\textbf{\gls{da} (Row~\ref{tech:da}).} Stops at \taxNo{6}{df}: the goal of bridging domain gaps persists at the \gls{llm} level, but the mechanism is superseded by \gls{cpt} and domain-specific \gls{peft}.

\textbf{\gls{pe} vs.\ \gls{icl} (Rows~\ref{tech:pe} vs.~\ref{tech:icl}).} 
While structurally identical on their underlying mechanism (\dimens{D}{1}), scope (\dimens{D}{5}), and model tier (\dimens{D}{6}), and partially overlapping in their goals (\dimens{D}{2}) and persistence (\dimens{D}{4}), the shift in data requirements (\tax{3}{zeroSh} vs. \tax{3}{few}) causes their profiles to diverge. \gls{icl} uniquely introduces \tax{2}{task}, while \gls{pe} distinctly extends to \tax{2}{behavControl} and \tax{4}{verPersistent} implementations.
 
\textbf{\gls{moe} adaptation (Row~\ref{tech:moe}).} Straddles architecture and adaptation; the row captures the adaptation act (training routers, adding experts) rather than the architectural pattern.

\textbf{\gls{rlvr} vs.\ \gls{rlhf}/\gls{dpo} (Row~\ref{tech:rlvr} vs.\ Rows~\ref{tech:rlhf}--\ref{tech:dpo}).} 
All share \tax{1}{paramUpd}
but differ on \dimens{D}{2} (\tax{2}{reas} vs. \taxNo{2}{algnSafety}) and \dimens{D}{3} (\tax{3}{verify} vs. \tax{3}{pref}).
\gls{rlvr} and \gls{dpo} share \taxNo{5}{partWhole}, while \gls{rlhf} adds \tax{5}{surrogate}.
The boundary blurs when \gls{rlvr} induces alignment-like properties or when \gls{rlhf} pipelines incorporate verifiable components; in production, both are often applied sequentially to the same base model.

\textbf{\gls{ssl}/\gls{cpt} (Row~\ref{tech:ssl}).} Has a dual role: pre-training paradigm (out of scope) and adaptation technique (in scope); unlike other data-manipulation techniques, \gls{ssl}-as-adaptation prescribes its own training objective and receives \taxNo{4}{unboundedCumulSchedPerm}.

\textbf{Machine Unlearning vs.\ Knowledge Editing (Rows~\ref{tech:mUnLrn} vs.~\ref{tech:ke}).} share \dimens{D}{1} and \dimens{D}{4} but differ on \dimens{D}{2}: unlearning targets \tax{2}{rem} while editing targets \tax{2}{augK}. The boundary blurs when correction effectively requires erasing the old association.

\textbf{Activation Steering vs.\ \gls{tta} (Rows~\ref{tech:actSteer} vs.~\ref{tech:tta}).} 
Both produce \tax{4}{trans} adaptations but differ on 
model type (\taxNo{6}{lm} vs.\ \taxNo{6}{df}), suggesting transient adaptation bifurcates at the \gls{fm}/\gls{llm} boundary.

\textbf{Machine Unlearning and Task Arithmetic (Rows~\ref{tech:mUnLrn} vs.~\ref{tech:taskArith}).} 
Share \tax{1}{WeightSpaceComp}
(task-vector manipulation): unlearning via negation is structurally identical to Task Arithmetic with a minus sign, raising the question of whether unlearning is a distinct technique or a specific application of Task Arithmetic. 

\textbf{\gls{lp} (Row~\ref{tech:lp}).} A strictly frozen linear probe does not modify the base model's internal representations or predictive behavior; it merely surfaces intermediate states. It is included in the adaptation framework strictly because appending a probe fundamentally alters the deployed system's explainability footprint and output characteristics (\tax{2}{explain}), triggering regulatory transparency mandates (e.g., EU AI Act Article 13) even when the core parameters remain untouched.

\textbf{Knowledge editing (Row~\ref{tech:ke}).} Straddles two fundamentally different structural mechanisms depending on the algorithm deployed. Closed-form editors (\gls{rome}~\cite{meng_locating_2022}, \gls{memit}~\cite{meng_mass-editing_2023}) algebraically rewrite weight matrices (\tax{1}{WeightSpaceComp}), whereas gradient-based editors (\gls{mend}~\cite{mitchell_fast_2022}) rely on compute-intensive backpropagation (\tax{1}{paramUpd}).
 
\section{Exploratory Structural Consistency Analysis}
\label{appendix:quantitative}

To rigorously verify the structural consistency of the characterizations defined in Table~\ref{tab:centerpiece}, we performed distance-based clustering analysis to quantify the internal diversity and topological distribution of our taxonomy.

\subsection{Methodology}
We calculated a pairwise Gower-style dissimilarity matrix for all \NbTech techniques across \dimens{D}{1}--\dimens{D}{6}. 

\textbf{1. Jaccard Distance:} 
The dimension values
were parsed into native mathematical sets. The distance was calculated using the Jaccard distance to faithfully preserve set-subset relationships (e.g., preventing techniques applicable to \tax{6}{llm} and \taxNo{6}{lm} from being treated as strictly orthogonal):
$$s_6(i,j) = 1 - \frac{|A \cap B|}{|A \cup B|}$$

\textbf{2. Final Unweighted Gower Distance:} 
The fundamental formulation of Gower distance allows for variable weighting across dimensions. In this analysis, we applied equal weighting ($w_k = 1$) across all six dimensions. The overall pairwise distance $D(i,j)$ between two techniques was computed as the arithmetic mean of their six independent dimensional distances:
$$D(i,j) = \frac{\sum_{k=1}^{K} w_k \ s_k(i,j)}{\sum_{k=1}^{K} w_k},$$
where $i,j \in \{1,…,\NbTech\}$ and $K = 6$.
This mean operation serves a critical normalizing function. By averaging the dimensional penalties, we ensure that the global distance strictly bounds to $[0, 1]$, and more importantly, we guarantee that no single dimension dominates the topological projection. A divergence in a technique's Mechanism ($D_1$) exerts the exact same geometric repulsive force in the latent space as a divergence in its Goal ($D_2$) or Persistence ($D_4$), ensuring the UMAP projection faithfully reflects the holistic multi-dimensional profile.

We subsequently projected the Gower distance matrix into a 2D latent space using UMAP (using a fixed random seed, \texttt{n\_neighbors=7}, and \texttt{metric='precomputed'}) to preserve both local neighborhoods and broader topological gradients.

\subsection{Topological Overlap of Functional Families}

\begin{figure}[htbp]
    \centering
    \includegraphics[width=\linewidth]{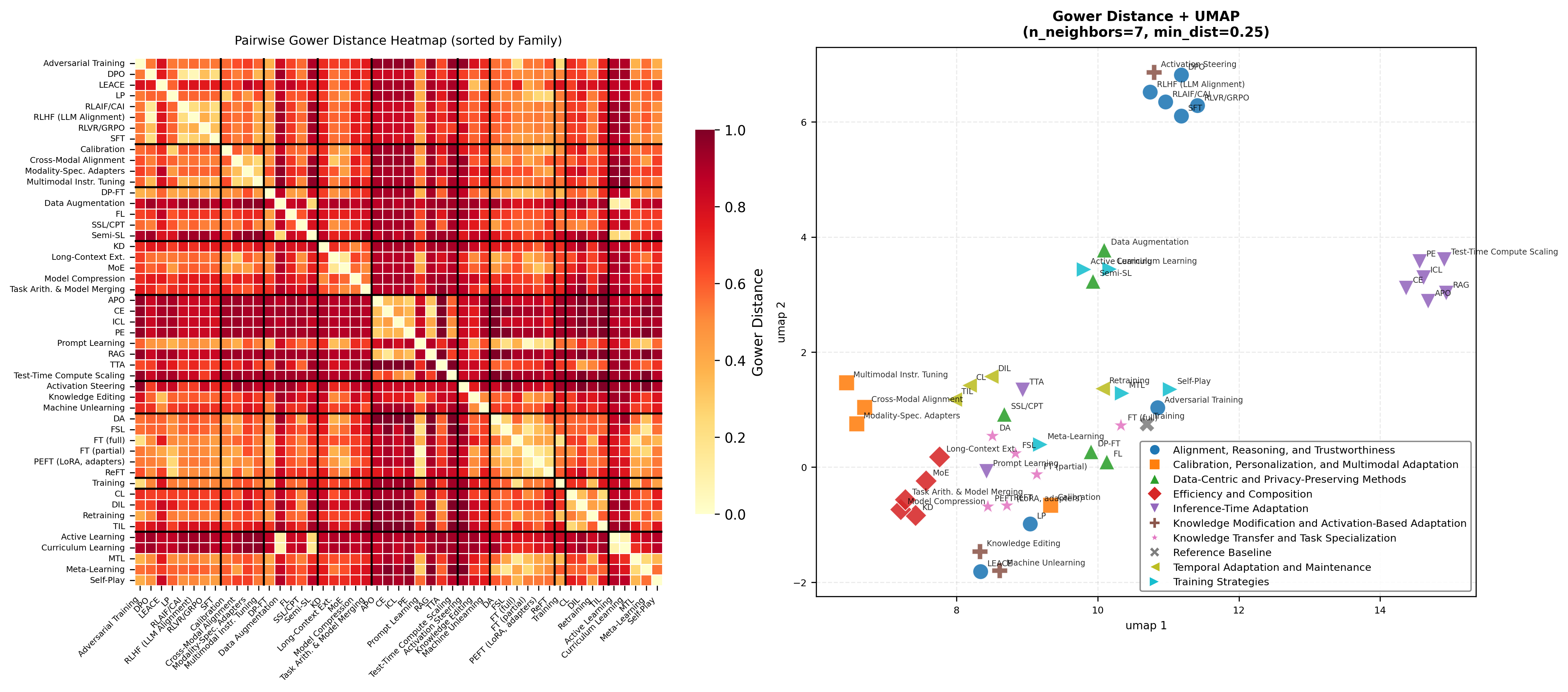}
    \caption{\textbf{Structural consistency analysis of the adaptation taxonomy.} (Left) Pairwise Gower distance heatmap of the \NbTech techniques, grouped by functional family.
    Light blocks indicate low Gower distance (high structural similarity); dark blocks indicate high distance (structural orthogonality). 
    (Right) 2D UMAP projection colored by the \nbFamilies predefined functional families.
    }
    \label{fig:GowerUmap}
\end{figure}

The Gower heatmap (Figure \ref{fig:GowerUmap}; left) provides a raw mathematical view of the structural overlaps. The presence of lighter-colored blocks across different family boundaries confirms that these categories are not isolated silos but share significant mechanical traits. This high-dimensional proximity is geometrically realized in the UMAP projection (Figure \ref{fig:GowerUmap}; right).
This visual distribution confirms that the functional families defined in Section 
\ref{sec:TechDefinitions} 
occupy complex, transitional regions of the categorical profile space rather than mutually exclusive mathematical clusters. 

\subsection{Quantitative Findings}
Silhouette analysis was used to measure the mathematical cohesion of the \nbFamilies Functional Families against the raw 6D coordinates. The metrics confirm that functional intent (Goal) does not strictly dictate mechanical structure:

\begin{itemize}
    \item \textbf{Raw Gower Silhouette ($+0.0173$):} Indicates baseline structural overlap across families in the high-dimensional space.
    \item \textbf{UMAP Silhouette ($-0.0346$):} Reflects compactness pressure 
    on highly diverse families 
    during lower-dimensional embedding.
\end{itemize}

Furthermore, intra-family distance variance highlights exactly where technical diversity occurs. 
The 
\textit{
Knowledge Modification and Activation-Based
} 
family exhibited the highest internal diversity (mean intra-family Gower distance = $0.7222$).
These metrics quantitatively support the visual topology presented in Figure \ref{fig:GowerUmap}, demonstrating how single-axis classification forces artificial boundaries on fundamentally multi-dimensional techniques.

\subsection{Dimensional Drivers of Topology}
\begin{figure}[htbp]
    \centering
    \includegraphics[width=\linewidth]{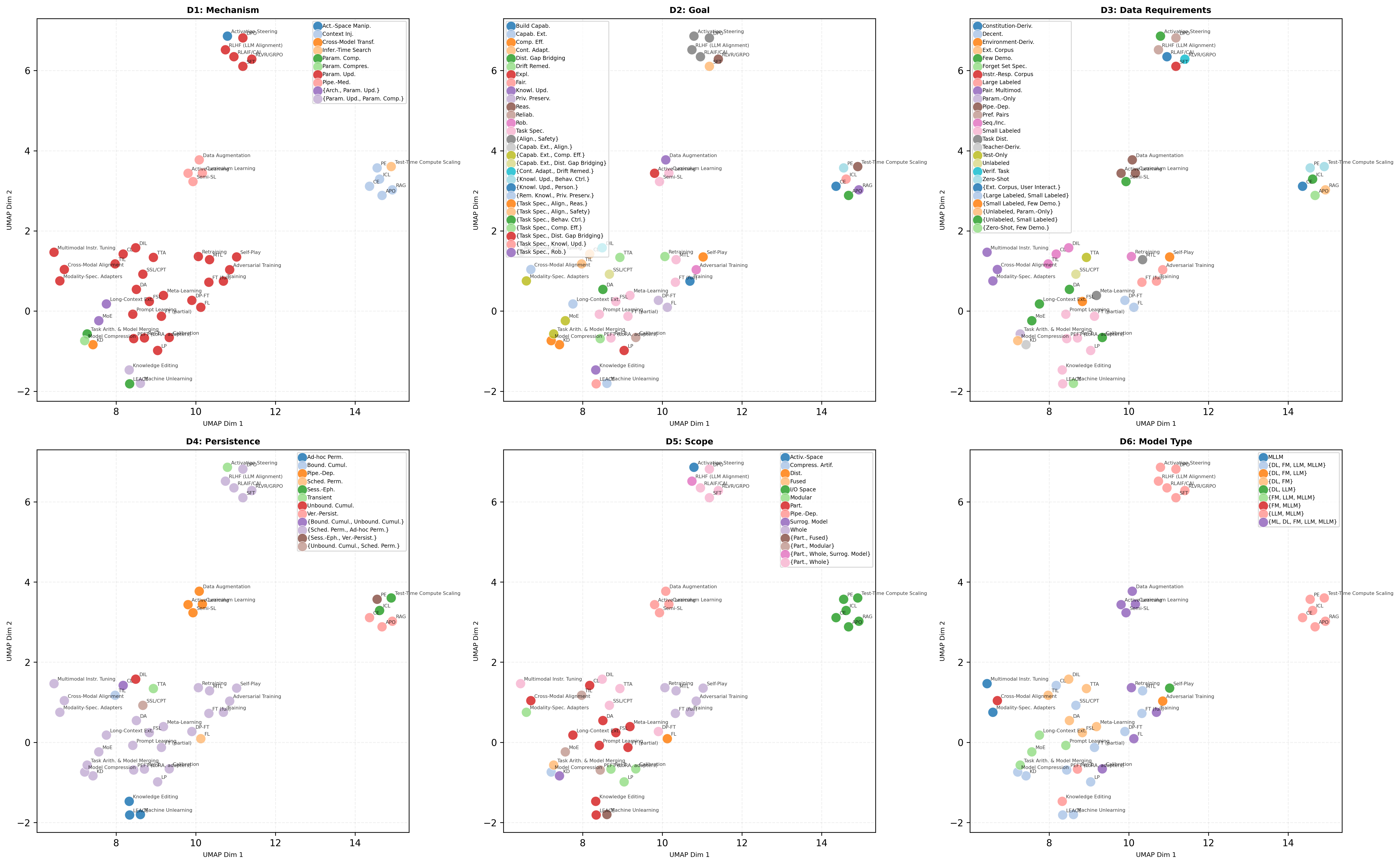}
    \caption{UMAP projection of the latent space, with each panel colored by an individual dimension \dimens{D}{1}--\dimens{D}{6}.}
    \label{fig:taxonomy_per_dimension}
\end{figure}

To isolate how individual axes of our framework influence the global geometric space, we additionally generated a per-dimension topological breakdown (Figure \ref{fig:taxonomy_per_dimension}). By recoloring the UMAP projection according to each independent dimension, this visualization reveals the precise structural drivers of the latent space. For instance, the clustering of the \textit{Inference-Time Adaptation} family is 
mainly governed by a convergence on Mechanism (\tax{1}{ctx}) and Scope (\tax{5}{extContxt}).
This dimensional breakdown empirically demonstrates that while techniques share primary structural anchors, no single dimension exerts monopolistic control over the adaptation landscape.

\section{Governance and Compliance Detail}
\label{appendix:governance}

This appendix provides extended treatment of three governance topics that the main text covers in summary form: (i)~the lifecycle-stage-by-stage mapping between framework dimensions and the \gls{nist} AI \gls{rmf}; (ii)~the three-way breakdown of how \dimens{D}{1}--\dimens{D}{6} support \gls{mdr}/\gls{ivdr} compliance documentation; and (iii)~the broader trustworthiness and responsible-adaptation agenda for \gls{fm} deployments.

\subsection{Lifecycle-Stage Mapping to the \gls{nist} AI \gls{rmf}}
The U.S.\ \gls{nist} AI \gls{rmf} (AI \gls{rmf} 1.0) outlines a seven-stage lifecycle for trustworthy AI systems, namely Plan and Design, Collect and Process Data, Build and Use Model, Verify and Validate, Deploy and Use, Operate and Monitor, and Use or Impacted By, across which organizations apply four core functions: Govern, Map, Measure, and Manage~\cite{noauthor_nist_2023}. Complementary management-system standards, notably ISO/IEC~42001 (AI management systems) and ISO/IEC~23894 (AI risk management guidance), provide organizational scaffolding for implementing these controls.
In 2025 and 2026, this foundational framework was expanded via the ``Generative AI Profile'' (\gls{nist}-AI-600-1)~\cite{autio_artificial_2024} and the April 2026 ``Trustworthy AI Profile'' for critical infrastructure concept note~\cite{stanley_concept_2026}, emphasizing model-specific threat taxonomies that are highly relevant to post-training modifications.

\paragraph{Operate and Monitor.} This stage mandates continuous performance tracking and model updating, directly invoking \taxNo{2}{ContAdaptDriftRemed}
(Retraining, \gls{cl}) and the 
\taxNo{4}{SchedAdhocPerm} or \taxNo{4}{boundedUnboundedCumul}
persistence categories that these techniques produce. The framework clarifies what ``model update'' means in compliance terms: not a single operation, but a choice among \dMechanism with different \dPersistence and \dData, each carrying distinct risk profiles.

\paragraph{Verify and Validate.} This stage requires assessments of model validity, reliability, safety, and fairness. 
\taxNo{2}{algnSafety}
(\gls{rlhf}, \gls{dpo}), 
\tax{2}{reas}
(\gls{rlvr}), and 
\tax{2}{rob}
(Adversarial Training) all map directly to these requirements. The \gls{nist} Trustworthy and Responsible AI guidelines (AI 100-2 E2025) further formalize the requirement to evaluate adversarial threats throughout the model lifecycle~\cite{vassilev_adversarial_2025}, which the framework captures through Adversarial Training (Row~\ref{tech:AdvrsTrn}) and its 
\tax{2}{rob}
classification.

\paragraph{Collect and Process Data.} This stage addresses data governance, which connects to \dData. Techniques requiring preference pairs (\gls{rlhf}, \gls{dpo}) introduce human annotator governance requirements absent from techniques using only existing labeled data. Techniques requiring verifiable task data (\gls{rlvr}) introduce different governance considerations: the provenance and correctness of verification oracles must be documented, but human annotator bias is less of a concern. Techniques requiring decentralized data (\gls{fl}) invoke data sovereignty and cross-jurisdictional compliance requirements. The \dData dimension makes these data-governance obligations explicit and technique-specific.

\subsection{EU \gls{mdr}/\gls{ivdr} Three-Way Breakdown}
Under the EU \gls{mdr}, software qualifying as a medical device, including AI/ML-driven \gls{mdsw}, is subject to lifecycle obligations extending beyond initial conformity assessment. Article~83 requires manufacturers to plan, establish, document, implement, maintain, and update a \gls{pms} system proportionate to the risk class and appropriate for the device type~\cite{european_parliament_and_of_the_council_regulation_2017}. Article~84 further requires that this system be based on a \gls{pms} plan, which forms part of the technical documentation for non-custom-made devices~\cite{european_parliament_and_of_the_council_regulation_2017}. The \gls{ivdr} establishes parallel post-market obligations for in vitro diagnostic devices~\cite{european_parliament_and_of_the_council_regulation_2017_746}. Although the \gls{mdr} does not explicitly distinguish adaptive AI/ML systems, its requirement for continuous post-market data collection, evaluation, and documentation implies ongoing lifecycle documentation obligations, which are particularly demanding for systems that evolve over time.
The \nbDim-dimensional framework supports \gls{mdr}/\gls{ivdr} compliance documentation in three specific ways.

\paragraph{First, \dMechanism and \dPersistence support technical documentation.} The technical-documentation requirement is that the manufacturer describe the design and characteristics of the device, including planned modifications. A statement that a deployed model ``was updated'' is insufficient; \gls{mdr} Annex II/III documentation expectations are better met by stating that the model received a 
\tax{1}{paramUpd}
producing a 
\taxNo{4}{SchedAdhocPerm}, \taxNo{5}{partlMod}
change via a specified \gls{peft} method.

\paragraph{Second, \dGoal and \dData support benefit-risk and clinical-evaluation documentation.} A change driven by 
\tax{2}{DriftRemed}
using 
\tax{3}{smallD}
has a different benefit-risk profile from a change driven by 
\tax{2}{task}
using the same data category, even if \dMechanism and \dPersistence are identical.

\paragraph{Third, \dPersistence and \dModel inform change-control framing.} These dimensions support decisions about whether a planned modification is best handled within an approved change-control framework, treated as a substantial modification requiring renewed conformity assessment, or addressed through \gls{pms}-driven updates to technical documentation. The framework does not make these determinations, which remain context-dependent, but it provides the vocabulary needed to describe the intervention precisely enough that the determination can be made consistently.

The \gls{mdr}'s vigilance reporting obligations (Articles~87--92) and the \gls{ivdr}'s corresponding obligations (Articles~82--87) are not adaptation events themselves but may trigger them: a reported incident may motivate a 
\tax{2}{DriftRemed}
retraining, a 
\tax{2}{rem}
unlearning step, or a 
\tax{2}{algn}
update. The framework's \dGoal axis makes the link between vigilance obligations and the appropriate adaptation response explicit.

\subsection{Trustworthiness and Responsible Adaptation}
Beyond specific regulatory frameworks, the broader trustworthiness agenda for \gls{fm} deployments intersects with adaptation at multiple dimensions.

\paragraph{Safety and alignment.} The \tax{2}{algn} category (\gls{rlhf}, \gls{dpo}) and 
\tax{2}{reas}
category (\gls{rlvr}) represent the primary technical mechanisms for embedding safety constraints and reasoning capability in \glspl{llm} and \glspl{mllm}. The distinction among alignment, reasoning, and specialization (all \dimens{D}{2} categories) is critical: a model well-specialized for a domain but poorly aligned may produce harmful outputs; a model well-aligned but lacking reasoning depth may fail at complex tasks; a model that reasons well but is unspecialized may produce unhelpful outputs. Governance frameworks must address all three goals, and the \dGoal axis makes this requirement explicit.

\paragraph{Data provenance and attribution.} The production of synthetic content by \glspl{fm} introduces requirements for training-data verification and attribution~\cite{hausenloy_towards_2025}. The \dData dimension identifies which techniques consume which data types, enabling traceability from adaptation output back to training-data provenance, a requirement explicitly mandated by emerging clinical AI reporting standards such as TRIPOD+AI for adaptive prediction models~\cite{collins_tripodai_2024}. Techniques using preference pairs require documentation of annotator selection, guidelines, and potential biases. Techniques using external corpora (\gls{rag}, 
\tax{3}{ext}
require documentation of retrieval-source quality and currency.

\paragraph{Dynamic benchmarking and continual monitoring.} 
Static evaluation is insufficient for adapted models,
a limitation heavily emphasized by the clinical-AI-governance community~\cite{lekadir_future-ai_2025,gilbert_algorithm_2021}. Because adaptation may improve performance on a target task while simultaneously degrading safety benchmarks or exacerbating bias, experts emphasize the need for continual, localized monitoring of clinical AI~\cite{muehlematter_approval_2021}.
Dynamic benchmarking platforms that evaluate models across multiple \dimens{D}{2} goals simultaneously~\cite{huang_trustworthiness_2025} align with the framework's multi-dimensional design: a well-adapted model should score well on its target \dGoal without regression on safety-related 
\dGoal.

\paragraph{Privacy.} The 
\tax{5}{dIVdist}
category (\gls{fl}) and \gls{peft}-based personalization address the tension between adaptation and privacy. The framework identifies this tension structurally: personalization 
(\tax{2}{userPerson})
requires user data 
(\tax{3}{userInteract}),
which conflicts with the 
\tax{2}{priv}
goal. \gls{fl} and differential-privacy mechanisms resolve this conflict by operating on 
\tax{5}{dIVdist},
ensuring that raw data never leaves the user's device.

\paragraph{Right to be forgotten and knowledge removal.} The \gls{gdpr}'s right to erasure and similar regulations create a direct mandate for the 
\tax{2}{rem}
goal category~\cite{noauthor_art_nodate}. Machine Unlearning (Row~\ref{tech:mUnLrn}) is the primary technical mechanism for compliance, but its current limitations (approximate rather than exact erasure, vulnerability to relearning attacks) mean that governance frameworks may need to specify acceptable thresholds for knowledge suppression rather than requiring provable deletion. Knowledge editing (Row~\ref{tech:ke}) addresses a related compliance need: correcting factual errors or outdated information without full model replacement. Both techniques intersect with the EU AI Act's post-market monitoring requirements, since factual errors discovered post-deployment may require targeted correction rather than wholesale retraining.

\end{document}